\documentclass[dvipsnames]{article}

\PassOptionsToPackage{sort,numbers}{natbib}

\usepackage[final]{neurips_data_2023}

\usepackage[utf8]{inputenc} \usepackage[T1]{fontenc}    \usepackage{hyperref}       \usepackage{url}            \usepackage{booktabs}       \usepackage{amsfonts}       \usepackage{nicefrac}       \usepackage{microtype}      \usepackage{xcolor}         \usepackage{caption}
\usepackage{subcaption}
\usepackage{amssymb}\usepackage{pifont}\usepackage{bbding}
\usepackage{realboxes}

\usepackage{lipsum}
\usepackage{xspace}
\usepackage{bm}
\usepackage{bbm}
\usepackage{multirow}
\usepackage{paralist}
\usepackage{rotating}
\usepackage{nicefrac}
\usepackage{enumitem}
\usepackage{listings}
\usepackage{xcolor}

\usepackage{microtype}
\usepackage{balance}
\usepackage{booktabs}
\usepackage{tabularx}
\usepackage{ragged2e}
\usepackage{multirow}
\usepackage{microtype}
\usepackage{balance}
\usepackage{setspace}

\graphicspath{{./}{./graphics/}}
\newcolumntype{H}{>{\setbox0=\hbox\bgroup}c<{\egroup}@{}}

\usepackage{amsmath,amssymb,amsthm}

\newcommand{\eat}[1]{\ignorespaces}

\usepackage{tikz}
\usepackage{verbatim}
\usetikzlibrary{arrows}
\usetikzlibrary{shapes,snakes}
\usetikzlibrary{decorations.pathmorphing} \usetikzlibrary{fit}					\usetikzlibrary{backgrounds}	

\usepackage{ragged2e}
\usepackage{multirow}
\usepackage{microtype}
\usepackage{balance}
\usepackage{setspace}

\graphicspath{{./}{./graphics/}}
\newcolumntype{R}[1]{>{\RaggedLeft\arraybackslash}} \newcolumntype{L}[1]{>{\RaggedRight\arraybackslash}}

\newcommand{\abs}[1]{\left|#1\right|}

\newcommand{\eg}{\emph{e.g.}}
\newcommand{\ie}{\emph{i.e.}}

\renewcommand{\subsubsection}[1]{\noindent\textbf{#1:}}

\providecommand{\mat}[1]{\boldsymbol{\mathrm{#1}}}\renewcommand{\vec}[1]{\boldsymbol{\mathrm{#1}}}

\DeclareMathOperator{\hugeE}{\mbox{\huge\raise-0.3ex\hbox{E}}}
\DeclareMathOperator{\p}{\mathbb{P}}
\DeclareMathOperator{\hugep}{\mbox{\huge\raise-0.3ex\hbox{$\p$}}}

\newcommand{\RR}{\mathbb{R}}

\newcommand{\NN}{\mathbb{N}}

\providecommand{\mM}{\ensuremath{\mat{M}}}

\providecommand{\mP}{\ensuremath{\mat{P}}}

\providecommand{\ve}{\ensuremath{\vec{e}}}

\providecommand{\vm}{\ensuremath{\vec{m}}}

\providecommand{\vx}{\ensuremath{\vec{x}}}

\newcommand{\xmark}{\ding{55}}

\newcommand{\fig}[3][\relax]{\begin{figure}[htp]\centering
\includegraphics[#2]{#3}\vspace{-0.1in}
		\ifx\relax#1\else\caption{{#1}}\fi
\end{figure}}

\newcommand{\dashrule}[1][black]{\color{#1}\rule[\dimexpr1.0ex-.2pt]{4pt}{.4pt}\xleaders\hbox{\rule{2pt}{0pt}\rule[\dimexpr1.0ex-.2pt]{4pt}{.4pt}}\hfill\kern-3pt}

\newcommand{\benchmark}{{GLEMOS}\xspace}
\newcommand{\benchmarkurl}{\url{https://github.com/facebookresearch/glemos}}

\newcommand{\rs}{RandSel\xspace}
\newcommand{\gbperf}{GB-Perf\xspace}
\newcommand{\gbperffull}{Global Best-AvgPerf\xspace}
\newcommand{\gbrank}{GB-Rank\xspace}
\newcommand{\gbrankfull}{Global Best-AvgRank\xspace}
\newcommand{\as}{AS\xspace}
\newcommand{\asfull}{ARGOSMART\xspace}
\newcommand{\isac}{ISAC\xspace}

\let\ss\undefined
\newcommand{\ss}{S2\xspace}

\newcommand{\alors}{ALORS\xspace}
\newcommand{\ncf}{NCF\xspace}
\newcommand{\metaod}{MetaOD\xspace}
\newcommand{\metagl}{MetaGL\xspace}

\newcommand{\cbit}{\begin{compactitem}}
	\newcommand{\ceit}{\end{compactitem}}
\newcommand{\cben}{\begin{compactenum}}
	\newcommand{\ceen}{\end{compactenum}}

\newcommand{\bal}{\begin{align}}
\newcommand{\ean}{\end{align}}
\newcommand{\bit}{\begin{itemize}}
\newcommand{\eit}{\end{itemize}}
\newcommand{\ben}{\begin{enumerate}}
\newcommand{\een}{\end{enumerate}}
\newcommand{\beq}{\begin{equation}}
\newcommand{\eeq}{\end{equation}}

\DeclareMathAlphabet{\mathbcal}{OMS}{cmsy}{b}{n}

\newcommand{\modelSet}{\mathbcal{M}}

\newcommand{\testG}{G_{\rm test}}

\makeatletter
\renewcommand*\env@matrix[1][*\c@MaxMatrixCols c]{\hskip -\arraycolsep
	\let\@ifnextchar\new@ifnextchar
	\array{#1}}
\makeatother

\makeatother

\newcommand{\numModels}{366\xspace}
\newcommand{\numLinkPredModels}{350\xspace}
\newcommand{\numNodeClassModels}{327\xspace}

\newcommand{\numGraphs}{457\xspace}
\newcommand{\numNodeClassGraphs}{128\xspace}
\newcommand{\numGraphDiffs}{329\xspace}  

\newcommand{\numGraphDomains}{37\xspace}
\newcommand{\numNodeClassGraphDomains}{25\xspace}

\newcommand{\numBaselinesInWords}{ten\xspace}

\newcommand{\Mregular}{\textit{\textbf{M}\textsubscript{regular}}\xspace}
\newcommand{\Mgraphlets}{\textit{\textbf{M}\textsubscript{graphlets}}\xspace}
\newcommand{\Mcompact}{\textit{\textbf{M}\textsubscript{compact}}\xspace}
\newcommand{\Mreggraph}{\textit{\textbf{M}\textsubscript{reg+graphlets}}\xspace}

\newcommand*\sq{\mathbin{\vcenter{\hbox{\rule{.5ex}{.5ex}}}}}

\usepackage{framed}
\newtheorem{protoproblem}{Problem}

\newenvironment{cproblem}
{
\colorlet{shadecolor}{white}
\begin{shaded}\setlength{\topsep}{0pt}\begin{protoproblem}}{\end{protoproblem}\end{shaded}}

\usepackage{cleveref}
\crefname{protoproblem}{problem}{problems}
\usepackage{soul,ulem}
\usepackage{makecell}
\usepackage{lipsum}

\usepackage{mdframed}
\newenvironment{probwrapper}
{\par\noindent\normalfont\vspace{-0.0em}
\begin{mdframed}[
	linewidth=2pt,
	linecolor=black,
	bottomline=false,topline=false,rightline=false,
	innerrightmargin=0pt,innertopmargin=0pt,innerbottommargin=0pt,
	innerleftmargin=0.75em,skipabove=-1.0\baselineskip
	]}
{\end{mdframed}}

\usepackage{tabularx}
\usepackage{wrapfig}
\usepackage{longtable}
\usepackage{diagbox}

\title{\benchmark: Benchmark for Instantaneous\\Graph Learning Model Selection}

\author{\textbf{Namyong Park}$^1$\thanks{Correspondence: namyongp@meta.com}\:, \textbf{Ryan Rossi}$^2$, \textbf{Xing Wang}$^1$, \textbf{Antoine Simoulin}$^1$,\\ \textbf{Nesreen Ahmed$^3$,} \textbf{Christos Faloutsos}$^4$\\
$^1$Meta AI  $^2$Adobe Research  $^3$Intel Labs $^4$Carnegie Mellon University\\
}

\begin{document}

\maketitle

\begin{abstract}
The choice of a graph learning (GL) model (\ie, a GL algorithm and its hyperparameter settings) has
a significant impact on the performance of downstream~tasks.
However, selecting the right GL model becomes increasingly difficult and time consuming as more and more GL models are developed.
Accordingly, it is of great significance and practical value to equip users of GL with the ability~to~perform~a
\textit{near-instantaneous} selection of an effective GL model without manual intervention.
Despite the recent attempts to tackle this important problem, 
there has been no comprehensive benchmark environment to evaluate the performance of GL model selection methods.
To bridge this gap, we present \benchmark in this work, 
a comprehensive benchmark for instantaneous GL model selection that makes the following contributions.
(i) \benchmark provides extensive benchmark data for fundamental GL tasks, 
\ie, link prediction and node classification, including the performances of 
\numModels models on \numGraphs graphs on these tasks. (ii) \benchmark designs multiple evaluation settings, and 
assesses how effectively representative model selection techniques 
perform in these different settings.
(iii) \benchmark is designed to be easily extended with new models, new graphs, and 
new performance records. 
(iv) Based on the experimental results, we discuss the limitations of existing approaches and 
highlight future research directions.
To promote research on this significant problem, we make the benchmark data and code publicly available
at \benchmarkurl.
\end{abstract}

\vspace{-0.75em}
\section{Introduction}\label{sec:intro}
\vspace{-0.75em}

Graph learning (GL) methods~\cite{DBLP:journals/tai/00010YAWP021,DBLP:journals/tkde/ZhangCZ22} 
have achieved great success across multiple domains and 
applications that involve graph-structured data~\cite{DBLP:conf/www/Fan0LHZTY19,DBLP:conf/kdd/ParkKDZF20,DBLP:journals/corr/abs-2101-11174,DBLP:journals/bib/SuTZCW20,DBLP:journals/corr/abs-1709-03741,DBLP:conf/cikm/CaiCLGN0C21,DBLP:conf/wsdm/ParkLMCFD22,DBLP:conf/www/ParkRKBKDAF22}. At the same time, previous
studies~\cite{DBLP:journals/tec/DolpertM97,DBLP:conf/nips/YouYL20,park2023metagl} have shown that 
there is no universally good GL model that performs best across
\begin{wrapfigure}{r}{0.5\textwidth}
	\vspace{-1.85em}
	\begin{center}
		\includegraphics[width=0.5\textwidth]{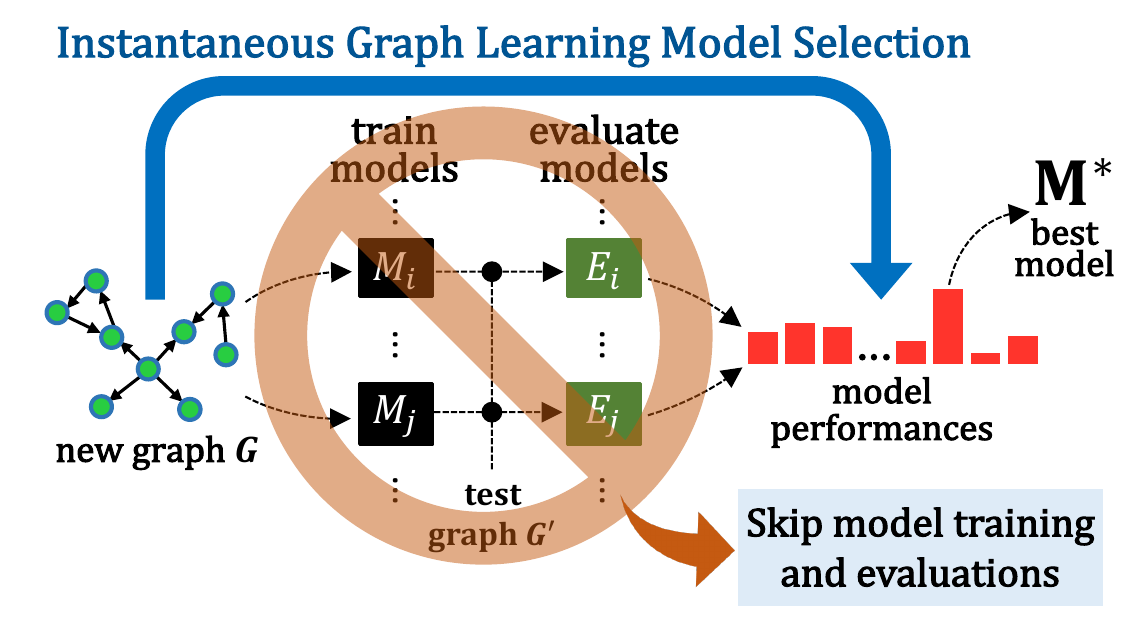}
	\end{center}
	\setlength{\abovecaptionskip}{-0.6em}
	\caption{Via instantaneous graph learning model selection, the best model can be found without performing computationally expensive model training and evaluations.}
	\label{fig:problem}
	\vspace{-3.5em}
\end{wrapfigure}
\textit{all} graphs and graph learning tasks.
Therefore, to effectively employ GL given a wide array of available
models, it is important to select the right GL model (\ie, a GL algorithm and
its hyperparameter settings) that will perform well for the given graph data and GL~task.

Ideally, we would want to be able to select the best GL model for the given graph
\textit{near-instantaneously}, that is, without having to train or evaluate different models multiple times on the new graph, 
since even a few such training and evaluations might take a considerable amount of time and resources (\Cref{fig:problem}). 
Enabling an instantaneous model selection for a completely new graph
involves addressing several technical challenges, which includes
modeling how well different GL methods perform on various graphs, and
establishing a connection between the new graph and observed graphs,
such that the best model for the new graph can be estimated 
in light of observed model performances on similar graphs.

\textbf{Problem Formulation.} 
With these considerations, we formally define this important problem, which we call \textit{Instantaneous Graph Learning Model Selection}, and the related terms as follows.

{\textit{Graph:}}
Let $G=(V, E, X, Y)$ be a graph where $V \subseteq \NN $ and
$E=\{ (i, j)  \mid i, j \in V \}$ denote the sets of nodes and edges, respectively;
$X$ denotes the input features, which can be 
node features ($X \in \RR^{|V| \times F_N}$), edge features ($X \in \RR^{|E| \times F_E}$),
or a set of both, where $F_N$ and $F_E$ denote the~dimension of corresponding input features; and
$Y$ denotes node labels ($ Y \in \NN^{|V|} $) or edge labels ($ Y \in \NN^{|E|} $).
Note that input features $ X $ and labels $ Y $ are considered optional since not all graphs have this information.

{\textit{Model:}} A model $ M $ refers to a GL method for the given GL task, such as link prediction, with specific hyperparameter settings.
In general, a GL model consists of two components, namely, 
(graph embedding method, hyperparameters) and (predictor, hyperparameters),
where the former produces a vector representation of the graph (\eg, node embeddings) and
the latter makes task-specific predictions (\eg, link prediction) given the embeddings.
The set $ \mathbcal{M} $ of models, from which the model selection is made, is normally heterogeneous,
where the configuration of each model is unique in the choice of its two components and their hyperparameter settings.

{\textit{Performance Matrix:}}
Let $\mP \in \RR^{n \times m}$ be a matrix containing observed model performances,~where 
$P_{ij}$ is the performance (\eg, accuracy) of~model~$j$ on graph~$i$.
$\mP$ can be sparse with missing~entries.

\vspace{-0.65em}
\begin{probwrapper}
\begin{cproblem}[{Instantaneous Graph Learning Model Selection}]\normalfont
\label{prob}~

\textbf{Given}
\begin{itemize}[leftmargin=1.05em,nosep,topsep=-0.25em]
		\item a training meta-corpus of $n$ graphs $\mathbcal{G} = \{G_i\}_{i=1}^{n}$ and $m$ models $\mathbcal{M} = \{M_j\}_{j=1}^{m}$ for a GL task (\eg, link prediction and node classification):
		
		\begin{enumerate}[label={(\arabic*)},leftmargin=1.6em]
\item performance matrices $\{\mP_k\}_{k=1}^{\ell}$, \ie, $\ell$ records of $m$ models' performance on $n$ graphs

			\item input features of the graphs in $\mathbcal{G}$ (if available) \item configurations (\ie, a GL method and its hyperparameter settings) of $m$ models in $\mathbcal{M}$
		\end{enumerate}
	
		\item an unseen test graph $\testG \notin \mathbcal{G}$
		
	\end{itemize}
\textbf{Select}
	\begin{itemize}[leftmargin=1.05em,nosep,topsep=-0.5em]
		\item the best  model $M^{*} \in \mathbcal{M}$ for $G_{\rm test}$
		\textbf{\textit{without}} training or evaluating any model in $\modelSet$~on~$G_{\rm test}$.
\end{itemize}
\end{cproblem}
\end{probwrapper}
\vspace{-0.9em}

\textbf{Status Quo and Our Contributions.}
In recent years, several methods have been developed for an efficient selection of GL models.
However, most of them cannot tackle Prob.~\ref{prob} as they require multiple rounds of model training and evaluations;
we review these methods in~Sec.~\ref{sec:relatedwork}.
Most recently, a subset of Prob.~\ref{prob} was studied by MetaGL~\cite{park2023metagl},
which proposed a GL model selection technique
that assumes plain graphs without input features, and operates without utilizing model configurations.
A few recent works~\cite{DBLP:conf/nips/YouYL20,DBLP:conf/iclr/CaoYLL23,DBLP:conf/nips/QinZWZZ22}
also provide performances of graph neural networks~(GNNs), although they cannot address Prob.~\ref{prob}.
While the datasets used in \cite{park2023metagl,DBLP:conf/nips/YouYL20,DBLP:conf/iclr/CaoYLL23,DBLP:conf/nips/QinZWZZ22} are  available,
they fall short of being a comprehensive benchmark environment to study this significant problem due to the following~reasons.

\begin{itemize}[leftmargin=0.90em,nosep]
	\item \textbf{Limited GL Task and Data.} 
	Focusing on link prediction, MetaGL~\cite{park2023metagl} only provides link~prediction performances, and 
	does not support other widely-used tasks, such as node classification,
	which limits follow-up studies and use of the benchmark for different GL tasks.
	Also, other related works~\cite{DBLP:conf/nips/YouYL20,DBLP:conf/iclr/CaoYLL23,DBLP:conf/nips/QinZWZZ22} 
	are limited in terms of the number and diversity of graphs they cover~(\Cref{tab:benchmark:comparison}). 

	\item \textbf{Limited Evaluation Settings.} Some important evaluation settings were not considered in MetaGL's benchmark, 
	such as out-of-domain and small-to-large settings as we later describe, which can be useful
	in evaluating the performance of model selection techniques in different practical settings.
	
	\item \textbf{Limited Extensibility.} The sets of models and graphs are assumed to be fixed, and 
	it is not easy to extend the benchmark with new graphs and models in a consistent and reproducible manner.

\end{itemize}

In this work, we address these limitations by developing a comprehensive benchmark for instantaneous graph learning model selection.
Overall, the contributions of this work are as follows.
\begin{itemize}[leftmargin=0.90em,nosep]
	\item \textbf{Extensive Benchmark Data with Multiple GL Tasks.} 
	We construct a benchmark dataset~that includes the performances of \numModels models on \numGraphs graphs over fundamental GL tasks,
	\ie, link prediction and node classification, 
	which is by far the largest benchmark for Prob.~\ref{prob} to our~knowledge.
	The benchmark also provides meta-graph features to capture the structural characteristics of graphs.

	\item \textbf{Comprehensive Evaluation Testbeds.}
	We evaluate \numBaselinesInWords representative methods for \Cref{prob}, 
	including both classical methods and deep learning-based ones, 
	using multiple evaluation~settings designed to assess the quality of model selection techniques from practical perspectives.
	
	\item \textbf{Extensible Open Source Benchmark Environment.}
	Our benchmark is designed to be easily extended with new models, new graphs, and new performance records.
	To promote further research on this significant problem, 
	we make the benchmark environment publicly available.

	\item \textbf{Future Research Directions.}
	We discuss the limitations of existing model selection~\mbox{methods},~and 
	highlight future research directions towards an instantaneous selection of graph learning models.
\end{itemize}

After reviewing related work in \Cref{sec:relatedwork},
we present the proposed benchmark data and testbeds in \Cref{sec:data,sec:meta-graph-features,sec:testbeds}.
Then we provide experimental results in \Cref{sec:exp}, and conclude~in~\Cref{sec:conclusion}.

\vspace{-0.5em}
\section{Related Work}\label{sec:relatedwork}
\vspace{-0.5em}
\subsection{Model Selection}\label{sec:relatedwork:modelselection}
\vspace{-0.5em}

Model selection refers to the process of selecting a learning algorithm and its hyperparameter settings.
In this section, we review existing model selection approaches, which we divide into two groups
depending on whether they require model evaluations
(\ie, performance queries for the new dataset).

\textbf{Evaluation-Based Model Selection:}
Most existing approaches to select machine learning models belong to this group,
ranging from simple solutions, 
such as random search~\cite{DBLP:journals/jmlr/BergstraB12} and grid search~\cite{DBLP:journals/corr/abs-1912-06059},
to more advanced and efficient ones that employ techniques such as
adaptive resource allocation~\cite{DBLP:journals/jmlr/LiJDRT17}, early stopping~\cite{DBLP:conf/kdd/GolovinSMKKS17}, and 
Bayesian optimization~\cite{DBLP:conf/nips/SnoekLA12,wu2019hyperparameter,DBLP:conf/icml/FalknerKH18}.
Inspired by these advancements, several model selection methods were recently developed for graph learning (GL) models.
To tackle challenges involved with GL model selection, 
these methods adapt existing ideas to GL models, such as
reinforcement learning~\cite{DBLP:conf/ijcai/GaoYZ0H20,DBLP:journals/corr/abs-1909-03184,DBLP:conf/kdd/LaiZZH20},
evolutionary algorithm~\cite{DBLP:conf/pricai/BuLL21},
Bayesian optimization~\cite{DBLP:conf/kdd/TuM0P019}, and
hypernets~\cite{DBLP:conf/sigir/ZhuTLL21}, 
as well as developing techniques specific to graph data, \eg,
subgraph sampling~\cite{DBLP:conf/kdd/TuM0P019} and
graph coarsening~\cite{DBLP:journals/corr/abs-2101-06427}.
Note that all of the above approaches cannot tackle the instantaneous GL model selection problem~(\Cref{prob})
as they rely on multiple model evaluations for performance queries of different combinations of GL methods and hyperparameter settings on the new dataset.

\begin{figure}[t!]
	\par\vspace{-1.5em}\par
	\centering
	\includegraphics[width=0.99\linewidth]{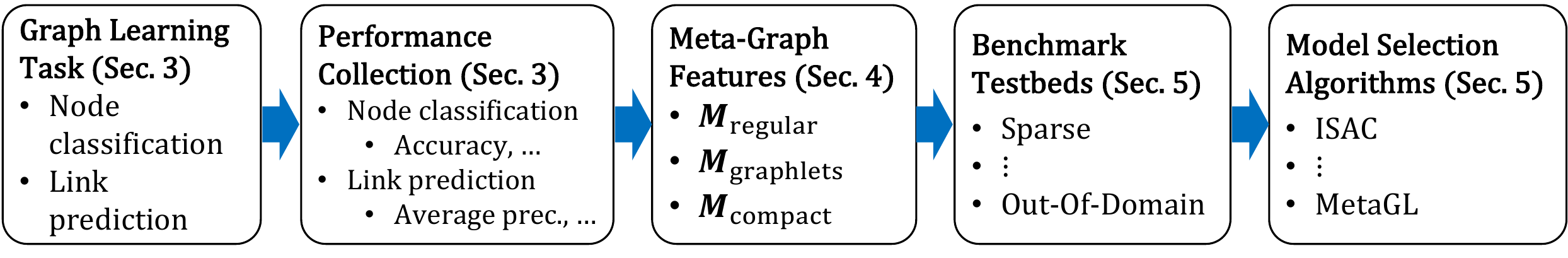}
\setlength{\abovecaptionskip}{0.1em}
	\caption{
		\benchmark provides a comprehensive benchmark environment, covering the steps required to achieve effective instantaneous GL model selection,
		with multiple options for major building~blocks.
	} 
	\label{fig:overview}
	\vspace{-1.3em}
\end{figure}

\textbf{Instantaneous Model Selection:} 
To select the best model without querying model performances on the new dataset, 
methods in this category typically utilize prior model performances or characteristic features of a dataset (\ie, meta-features).
A simple approach~\cite{DBLP:journals/ml/AbdulrahmanBRV18}
finds the globally best model (\ie, the one with the overall best performance over all observed datasets), and 
thus its model selection is independent of query datasets.
This can be refined by narrowing the search scope to similar datasets, 
where dataset similarities are modeled in the meta-feature space, \eg, using 
$ k $-nearest neighbors~\citep{nikolic2013simple} or clustering~\citep{DBLP:conf/ecai/KadiogluMST10}.
Another line of methods~\cite{xu2012satzilla2012,DBLP:conf/nips/ZhaoRA21,park2023metagl} 
take a different approach, which aims to predict the model performance on the given dataset
by learning a function that maps meta-features into estimated model performances.
Due to their ability to learn such a function in a data-driven manner,
this second group of methods generally outperformed the first group in previous studies~\cite{DBLP:conf/nips/ZhaoRA21,park2023metagl}.
While the above methods are one of the first efforts to achieve~instantaneous GL model selection,
several open challenges remain to be solved, as we discuss in~\Cref{sec:exp:discussion}.

\begin{table}[!htp]\centering
\par\vspace{-1.0em}\par
\setlength{\abovecaptionskip}{0.5em}
\caption{Comparison of \benchmark with previous works providing performances of GNN models.
}\label{tab:benchmark:comparison}
\setlength{\tabcolsep}{2pt}
\makebox[0.4\textwidth][c]{
\resizebox{1.0\textwidth}{!}{\begin{tabular}{lccccrrr}\toprule
	&\makecell[c]{\textbf{Benchmark}\\\textbf{Testbeds}} &\makecell[c]{\textbf{Instantaneous}\\\textbf{Selection Methods}} &\makecell[c]{\textbf{Meta-Graph}\\\textbf{Features}} &\makecell[c]{\textbf{Graph Learning}\\\textbf{Models}} &\makecell[c]{\textbf{\# Graph}\\\textbf{Datasets}} &\makecell[c]{\textbf{Graph Size}\\\textbf{(max \# nodes)}} &\makecell[c]{\textbf{\# Data}\\\textbf{Domains}} \\\midrule
	GNN-Bank-101~\cite{DBLP:conf/iclr/CaoYLL23} &\xmark &\xmark &\xmark &GNNs &12 &34k &5 \\
	NAS-Bench-Graph~\cite{DBLP:conf/nips/QinZWZZ22} &\xmark &\xmark &\xmark &GNNs &9 &170k &4 \\
	GraphGym~\cite{DBLP:conf/nips/YouYL20} &\xmark &\xmark &\xmark &GNNs &32 &34k &7 \\ 
	\textbf{GLEMOS (Ours)} &\Checkmark &\Checkmark &\Checkmark &\makecell[c]{\textbf{GNNs \& Non-GNNs}\\(\eg, node2vec, label prop.)} &\textbf{\numGraphs} &\textbf{496k} &\textbf{\numGraphDomains} \\
	\bottomrule
\end{tabular}
}}
\end{table}

\vspace{-0.7em}
\subsection{Benchmarks for Instantaneous Graph Learning Model Selection}\label{sec:relatedwork:benchmark}
\vspace{-0.5em}
A few recent works~\cite{DBLP:conf/nips/YouYL20,DBLP:conf/iclr/CaoYLL23,DBLP:conf/nips/QinZWZZ22}
address problems related to GL model selection, and provide~performances of GNNs on different datasets.
However, all of them perform evaluation-based model selection discussed above, 
which requires multiple rounds of model evaluations given a new dataset.

As the first benchmark for instantaneous GL model selection (Prob.~\ref{prob}),
\benchmark provides more~than just a collection of performance records, \ie, 
(1) benchmark testbeds and (2) existing algorithms (\Cref{sec:testbeds}) for instantaneous model selection, 
as well as different sets of (3) meta-graph features (\Cref{sec:meta-graph-features}). 
These features (1)-(3) are not provided by these previous works~\cite{DBLP:conf/nips/YouYL20,DBLP:conf/iclr/CaoYLL23,DBLP:conf/nips/QinZWZZ22}. 
Furthermore, \benchmark provides more comprehensive and diverse performance records than these works in several aspects
(\eg, in terms of included GL models and graph data distributions), 
as summarized in~\Cref{tab:benchmark:comparison}.

The two major components for instantaneous GL model selection  are 
\textit{historical model performances} of the GL task of interest (\eg, accuracy for node classification), and 
\textit{meta-graph features} to quantify graph similarities.
For each component, \benchmark provides several options to choose from.
Once~these components are chosen, users select a \textit{model selection algorithm}, as well as a \textit{benchmark testbed} to perform evaluation, 
out of several options available in \benchmark.
Fig.~\ref{fig:overview} summarizes these steps to use \benchmark. 
In the next sections, we describe what \benchmark provides for these~steps.

\begin{table}[!t]\centering
	\par\vspace{-1.0em}\par
	\setlength{\abovecaptionskip}{0.5em}
	\caption{Summary statistics of the \benchmark benchmark.}\label{tab:benchmark_summary}
	\small
\setlength{\tabcolsep}{6pt}
	\makebox[0.4\textwidth][c]{
		\resizebox{0.85\textwidth}{!}{\begin{tabular}{lrr}\toprule
				& \textbf{Node Classification Task} & \textbf{Link Prediction Task} \\ \midrule
				\makecell[l]{\textbf{Total performance evaluations}~(\Cref{sec:data})\\(\# model performances on benchmark graphs) } & 41,856 & 152,070 \\ \midrule
				\textbf{Total graphs}~(\Cref{sec:data:nodeclass:graphs,sec:data:linkpred:graphs}) & 128 & 457 \\
				$~\sq$ \textbf{Num nodes} & 34--421,961 & 34--495,957 \\
				$~\sq$ \textbf{Num edges} & 156--7,045,181 & 156--7,045,181\\
				$~\sq$ \textbf{Num node feats} & 2--61,278 & 2--61,278 \\
				$~\sq$ \textbf{Num node classes} & 2--195 & \textsc{n/a} \\
				$~\sq$ \textbf{Num graph domains} & 25 & 37 \\ \midrule
				\textbf{Total GL models}~(\Cref{sec:data:nodeclass:models,sec:data:linkpred:models}) & 327 & 350 \\ \midrule
				\textbf{Total meta-graph features}~(\Cref{sec:meta-graph-features}) & 58--1,074 & 58--1,074 \\ \midrule
				\textbf{Total model selection methods}~(\Cref{sec:testbeds}) & 10 & 10 \\ \midrule
				\textbf{Total benchmark testbeds}~(\Cref{sec:testbeds}) & 5 & 5 \\
\bottomrule
			\end{tabular}
	}}
	\vspace{-0.8em}
\end{table}

\vspace{-0.5em}
\section{Graph Learning Tasks and Performance Collection}\label{sec:data}
\vspace{-0.5em}

Prior model performances play an essential role in instantaneous GL model selection algorithms,
as they can estimate a candidate GL model's performance on the new graph based on its observed performances on similar graphs.
\benchmark provides performance collections for two fundamental graph learning tasks, 
\ie, node classification and link prediction. 
Below we discuss how the graphs and models are selected, and describe how model performances are evaluated for each GL task.

\vspace{-0.7em}
\subsection{Graphs and Models}
\vspace{-0.5em}

\textbf{Graphs.}
Our principle of selecting the graphs in \benchmark is to include diverse graph datasets, 
in terms of both the size and domain of the graph.
The size of selected graphs ranges from a few hundred edges to millions of edges, and
the graph set covers various domains, \eg, co-purchase networks, protein networks, citation graphs, and road networks.
As listed in \Cref{tab:benchmark:comparison},
the resulting graph set outperforms existing data banks in terms of the number and size of graphs, as well as the diversity of data domain.
\Cref{tab:benchmark_summary} shows the summary statistics of graphs, and the graph list is given in \Cref{tab:graphs}.

\textbf{Models.}
Our principle for selecting the models to include in \benchmark is to cover representative and widely-used GL methods.
We include graph neural network methods 
(\eg, GCN~\cite{DBLP:conf/iclr/KipfW17}, GAT~\cite{DBLP:conf/iclr/VelickovicCCRLB18}, and SGC~\cite{DBLP:conf/icml/WuSZFYW19}),
random walk-based node embeddings (\eg, node2vec~\cite{DBLP:conf/kdd/GroverL16}),
self-supervised graph representation learning methods (\eg, DGI~\cite{DBLP:conf/iclr/VelickovicFHLBH19}), and 
classical methods (\eg, spectral embedding~\cite{DBLP:journals/pr/LuoWH03}).\\
The resulting model set is more diverse than previous works, which considered GNNs alone (\Cref{tab:benchmark:comparison}).

\subsection{Node Classification}
\vspace{-0.3em}

\textbf{Graph Set.}\label{sec:data:nodeclass:graphs}
A subset of the graphs have node labels. Excluding the graphs without node labels, 
the node classification~graph~set is comprised of \numNodeClassGraphs graphs from \numNodeClassGraphDomains domains.

\textbf{Model Set.}\label{sec:data:nodeclass:models}
Most methods in \benchmark are applicable for both node classification and link prediction. 
In addition to these common methods, we also include label propagation~\cite{Zhu2002LearningFL}, which can be used for node classification.
The GL models evaluated for node classification and their hyperparameter settings are listed in \Cref{tab:modelset}.
In total, \numNodeClassModels models comprise our model set for node classification.

\textbf{Performance Collection.}\label{sec:data:nodeclass:perf}
For node classification, supervised models are optimized to produce the class distribution.
For unsupervised models, we first train them to produce latent node embeddings based on their own objective, 
and apply a trainable linear transform to transform embeddings into the class distribution.
More details on the experimental settings are given in \Cref{app:expsettings}.
To evaluate performance, we calculate multiple classification metrics, 
including accuracy, F1 score, average precision, and ROC AUC score.
Given the graph set $ \mathbcal{G} $ and model set $ \mathbcal{M} $ described above, 
we construct the performance matrix $ \mP $ 
by evaluating every model $M_j \in \mathbcal{M}$ on every graph $G_i \in \mathbcal{G}$,~\ie, 
\begin{align}
	P_{ij} = \text{Performance (\eg, accuracy, and ROC AUC) of model } M_j \in \mathbcal{M} \text{ on graph } G_i \in \mathbcal{G}.
\end{align}
\textit{Splitting:} 
We generate the train-validation-test node splits with a ratio of 64\%-16\%-20\%, respectively,
and train each model applying validation-based early stopping.
For reproducibility, we release all data splits, such that future model evaluations can be done using the same node splits.

\vspace{-0.5em}
\subsection{Link Prediction}
\vspace{-0.5em}

\textbf{Graph Set.}\label{sec:data:linkpred:graphs}
As link prediction task does not require node labels for evaluation, 
we greatly expand the graph set used for node classification by adding \numGraphDiffs more graphs.
With these graphs, the link prediction graph set consists of \numGraphs graphs from \numGraphDomains domains.
\Cref{tab:graphs} in Appendix gives the full list.

\textbf{Model Set.}\label{sec:data:linkpred:models}
All models used for node classification are used for link prediction, 
except for label propagation, which requires node labels.
We also add models designed for link prediction, \eg, 
SEAL~\cite{DBLP:conf/nips/ZhangC18} and Adamic/Adar~\cite{DBLP:journals/jasis/Liben-NowellK07}.
In total, \numLinkPredModels models comprise the link prediction model set~(\Cref{tab:modelset}).

\textbf{Performance Collection.}\label{sec:data:linkpred:perf}
For link prediction, GL models are optimized to produce latent node embeddings, and 
we apply a dot product scoring between the two node embeddings, followed by a sigmoid function, 
to obtain the link probability between the corresponding nodes.
We calculate multiple evaluation metrics to measure the link prediction performance, 
including average precision, ROC AUC score, and NDCG (normalized discounted cumulative gain)~\cite{DBLP:conf/colt/WangWLHL13}.

\textit{Splitting:} 
We randomly split edges into train-validation-test sets, with a ratio of 64\%-16\%-20\%, which form positive edge sets.
For positive edges, we randomly select the same amount of negative edges (\ie, nonexistent edges), which form negative edge sets.
Again, we release all edge splits.

\begin{table}[!t]\centering
	\par\vspace{-2.0em}\par
	\setlength{\abovecaptionskip}{0.25em}
	\setlength{\tabcolsep}{1.0pt}
	\caption{Graph learning methods and their hyperparameter settings that comprise the model set~$ \modelSet $. NC: Applicable for node classification. LP: Applicable for link prediction.}
	\label{tab:modelset}
	\fontsize{7.25}{8.25}\selectfont

	\makebox[0.4\textwidth][c]{
		\resizebox{0.95\textwidth}{!}{\renewcommand{\arraystretch}{0.95}
			\begin{tabularx}{1.0\linewidth}{lccXr}\toprule
				\textbf{Method} &\textbf{NC} &\textbf{LP} &\textbf{Hyperparameter Settings} &\textbf{Count} \\\midrule
				GCN~\cite{DBLP:conf/iclr/KipfW17} & \Checkmark & \Checkmark & act~$a \in \{\text{relu},\text{tanh},\text{elu}\}$, dropout~$d \in \{0.0,0.5\}$, hidden channels~$h \in \{16,64\}$, num layers~$\ell \in \{1,2,3\}$ & 30 \\
				GraphSAGE~\cite{DBLP:conf/nips/HamiltonYL17} & \Checkmark & \Checkmark & act~$a \in \{\text{relu},\text{tanh}\}$, aggr~$g \in \{\text{mean},\text{max}\}$, hidden channels~$h \in \{16,64\}$, jumping knowledge~$j \in \{\text{none},\text{last}\}$, num layers~$\ell \in \{1,2\}$ & 24 \\
				GAT~\cite{DBLP:conf/iclr/VelickovicCCRLB18} & \Checkmark & \Checkmark & concat~$c \in \{\text{true},\text{false}\}$, dropout~$d \in \{0.0,0.5\}$, heads~$n \in \{1,4\}$, hidden channels~$h \in \{16,64\}$, num layers~$\ell \in \{1,2,3\}$ & 40 \\
				GIN~\cite{DBLP:conf/iclr/XuHLJ19} & \Checkmark & \Checkmark & eps~$e \in \{0.0\}$, hidden channels~$h \in \{16,64\}$, num layers~$\ell \in \{1,2,3\}$, train eps~$t \in \{\text{true},\text{false}\}$ & 10 \\
				EGC~\cite{DBLP:conf/iclr/TailorOLL22} & \Checkmark & \Checkmark & aggregators~$a \in \{\text{[sum]},\text{[mean]},\text{[symnorm]},\text{[min]},\text{[max]},\text{[var]},\text{[std]}\}$, hidden channels~$h \in \{16,64\}$, num bases~$b \in \{4,8\}$, num layers~$\ell \in \{2\}$ & 28 \\
				SGC~\cite{DBLP:conf/icml/WuSZFYW19} & \Checkmark & \Checkmark & bias~$b \in \{\text{true},\text{false}\}$, num hops~$k \in \{1,2,3,4,5\}$ & 10 \\
				ChebNet~\cite{DBLP:conf/nips/DefferrardBV16} & \Checkmark & \Checkmark & Chebyshev filter size~$k \in \{1,2,3\}$, hidden channels~$h \in \{16,64\}$, normalization~$r \in \{\text{none},\text{sym},\text{rw}\}$, num layers~$\ell \in \{1,2\}$ & 27 \\
				PNA~\cite{DBLP:conf/nips/CorsoCBLV20} & \Checkmark & \Checkmark & aggregators~$a \in \{\text{[sum]},\text{[mean]},\text{[max]},\text{[var]}\}$, hidden channels~$h \in \{16\}$, num layers~$\ell \in \{1,2\}$, scalers~$s \in \{\text{[identity]},\text{[amplification]},\text{[attenuation]},\text{[linear]}\}$, towers~$t \in \{1\}$ & 32 \\
				Spectral Emb.~\cite{DBLP:journals/pr/LuoWH03} & \Checkmark & \Checkmark & num components~$h \in \{16,64\}$, tolerance~$t \in \{0.1,0.01,0.001,0.0001\}$ & 8 \\
				GraRep~\cite{DBLP:conf/cikm/CaoLX15} & \Checkmark & \Checkmark & num components~$h \in \{16,32,64\}$, power~$p \in \{1,2\}$ & 6 \\
				DGI~\cite{DBLP:conf/iclr/VelickovicFHLBH19} & \Checkmark & \Checkmark & encoder act~$a \in \{\text{prelu},\text{relu},\text{tanh}\}$, hidden channels~$h \in \{16,64\}$, summary~$s \in \{\text{mean},\text{max},\text{min},\text{var}\}$ & 24 \\
				node2vec~\cite{DBLP:conf/kdd/GroverL16} & \Checkmark & \Checkmark & context size~$w \in \{5,10\}$, hidden channels~$h \in \{16,64\}$, $p \in \{1,2,4\}$, $q \in \{1,2,4\}$, walk length~$l \in \{10,20\}$ & 72 \\
				Label Prop.~\cite{Zhu2002LearningFL} & \Checkmark &  & alpha~$\alpha \in \{0.99,0.9,0.8,0.7\}$, num layers~$\ell \in \{1,2,3,4\}$ & 16 \\
				Jaccard's Coeff.~\cite{DBLP:journals/jasis/Liben-NowellK07} &  & \Checkmark & - & 1 \\
				Resource Alloc.~\cite{zhou2009predicting} &  & \Checkmark & - & 1 \\
				Adamic/Adar~\cite{DBLP:journals/jasis/Liben-NowellK07} &  & \Checkmark & - & 1 \\
				SEAL~\cite{DBLP:conf/nips/ZhangC18} &  & \Checkmark & GNN conv~$c \in \{\text{GCN},\text{SAGE},\text{GAT}\}$, GNN hidden channels~$g \in \{16,64,128\}$, $k \in \{0.6,0.1\}$, MLP hidden channels~$m \in \{32,128\}$, num hops~$n \in \{1\}$ & 36 \\ 
				\bottomrule
				& & & \makecell[r]{\textbf{{Total Count}}} &\textbf{366} \\
			\end{tabularx}
	}}
	\vspace{-3.0em}
\end{table}

\section{Collection of Meta-Graph Features}\label{sec:meta-graph-features}
\vspace{-2.5em}

\begin{wrapfigure}{r}{0.35\textwidth}
	\par\vspace{-2.7em}\par
	\begin{center}
		\includegraphics[width=1.1\linewidth,trim={1.0cm 0 0 0},clip]{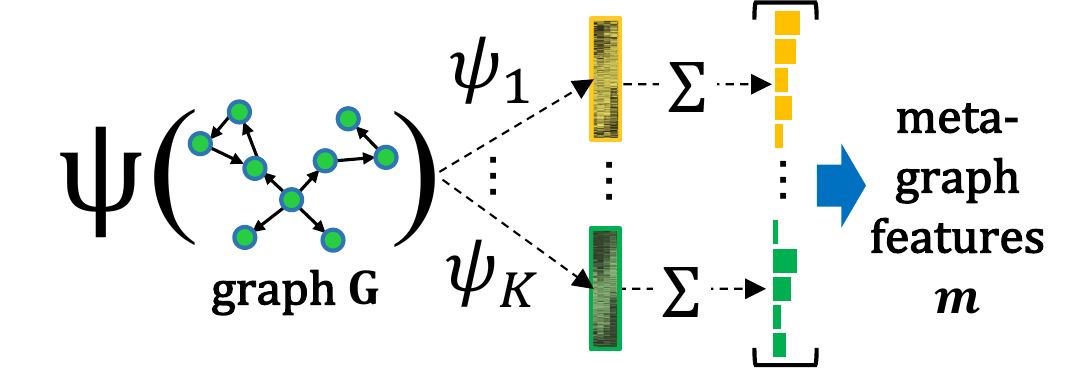}
\end{center}
\setlength{\abovecaptionskip}{0.0em}
\caption{
		Meta-graph features summarize structural graph characteristics into a fixed-size feature vector.
} 
	\label{fig:meta-graph-feature}
	\vspace{-1.2em}
\end{wrapfigure}

Instantaneous model selection algorithms carry over historical model performances on various graphs 
to estimate how GL models would perform on a new graph.
In that process, performance transfer can be done more effectively when we consider graph similarities,
such that the performance transfer would be done adaptively based on the similarities between graphs.
Structural meta-graph features provide an effective way to that end by summarizing a graph into a fixed-size feature vector in terms of its structural characteristics.
\benchmark provides various meta-graph features, which can capture important graph structural properties.
Below we first discuss how \benchmark generates fixed-length meta-graph features, as depicted in \Cref{fig:meta-graph-feature}, 
and then describe the structural features included in \benchmark, which are organized into three sets for convenience.

\textbf{Feature Generation.} \benchmark produces meta-features in two steps, following an earlier work~\cite{park2023metagl}.

$ \circ $\:\textit{Step 1—Structural Feature Extraction:} 
A structural meta-feature extractor $ \psi_k $ is a function that transforms an input graph $ G $ 
into a vector, which represents a distribution of structural features over nodes or edges.
For example, degree and PageRank scores of all nodes correspond to node-level feature distributions, 
and triangle frequency for each edge corresponds to an edge-level feature distribution.
In general, we apply a set of such extractors $\Psi = \{ \psi_1, \ldots, \psi_K \}$ to the graph $ G $,
obtaining a set $ \Psi(G) = \{\psi_1(G), \ldots, \psi_K(G) \}$ of multiple feature distributions.

$ \circ $\:\textit{Step 2—Statistical Feature Summarization:}
Since the number of nodes or edges in each graph determines the size of the output from the meta-feature extractors $ \psi_k(G) $, 
those structural feature distributions cannot be directly used to compare graphs with different number of nodes or edges.
Step 2 addresses this issue via statistical feature summarization, which applies a set $ \Sigma $ of statistical functions
(\eg, mean, entropy, skewness, etc)
that summarize feature distributions $ \psi_k(G) $ of varying size into fixed-length feature vectors;
\ie, $ \dim(\Sigma(\psi_k(G_i))) = \dim (\Sigma(\psi_k(G_j))) $ for two graphs $ G_i $ and $ G_j $.
By combining all $ K $ summaries, 
graph $ G $'s meta-graph feature $ \vm $ is obtained to~be 
$\vm = [\Sigma(\psi_1(G)); \cdots; \Sigma(\psi_K(G)) ]$.
\Cref{table-meta-features} lists the statistical functions $ \Sigma $ used in \benchmark.

\textbf{Collection of Meta-Graph Features.}
Different graph features may capture different structural~properties.
Thus, \benchmark aims to provide representative and diverse graph features, which have been proven effective in earlier studies, 
while making it easy to work with any set of features.
For the convenience of the users, we group the currently supported features into the following three sets:
\textit{\textbf{M\textsubscript{regular}}} includes widely used features that capture structural characteristics of a graph at both node and graph levels;
\textit{\textbf{M\textsubscript{graphlets}}} considers features based on the frequency of graphlets, as they can provide additional information;
\textit{\textbf{M\textsubscript{compact}}} is intended to use the least space, 
while providing several important features that capture node-, edge-, and graph-level characteristics.
The details of each~set~are~as~follows.

\textit{\textbf{M\textsubscript{regular}}:} 
This set includes 318 meta-graph features.
We derive the distribution of node degrees, k-core numbers, PageRank scores, 
along with the distribution of 3-node paths (wedges) and 3-node cliques (triangles).
Given these five distributions, we summarize each using the set of 63 statistical functions $\Sigma$, giving us a total of 315 features.
We include three additional features based on the density of graph $G$ and the density of the symmetrized graph, along with the assortativity coefficient.

\textit{\textbf{M\textsubscript{graphlets}}:}
This set includes 756 meta-graph features.
First, we derive the frequency of all 3 and 4-node graphlet orbits per edge in $G$.
Next, we summarize each of the 12 graphlet orbit frequency distributions using the set of 63 statistical functions $\Sigma$, 
giving us a total of 756 meta-graph features.

\textit{\textbf{M\textsubscript{compact}}:} 
This set consists of 58 total meta-graph features, including 9 simple statistics such as 
number of nodes and edges, density of the graph $ G $, 
max vertex degree, average degree, assortativity coefficient, 
and the maximum k-core number of $G$, along with the mean and median k-core number of a vertex in $G$.
We also include the global clustering coefficient, total triangles, as well as the mean and median number of triangles centered at an edge.
We further include the total 4-cliques as well as the mean and median number of 4-cliques centered at an edge.
Besides the above 16 features, we also compute the frequency of all 3 and 4-node graphlet orbits per edge, and 
from these 12 frequency distributions, we derive the mean, median, and max.
We also derive the graphlet frequency distribution from the counts of all six 4-node graphlets and include those values directly as features.

Note that the framework is flexible, and users can choose to use any set of features, 
either a subset of the current features (\eg, to further improve efficiency and use less space),
or their superset (\eg, to capture distinct structural characteristics using different features in addressing new tasks).

\vspace{-1.0em}
\section{Benchmark Testbeds and Algorithms}\label{sec:testbeds}
\vspace{-0.8em}

\subsection{Benchmark Testbeds}
\vspace{-0.5em}

\benchmark provides multiple benchmark testbeds (\ie, evaluation settings and tasks) 
designed to assess model selection performance in different usage scenarios. We describe them in detail below.

\textbf{Fully-Observed Testbed.}
In this setup, model selection algorithms are provided with a full performance matrix~$ \mP $ for the given graph learning task, 
\ie, without any missing entry in $ \mP $.
Accordingly, this testbed measures model selection performance in the most information-rich setting,
where all models in the model set $ \modelSet $ have been evaluated on all observed graphs.

\textit{Splitting:} We apply a stratified 5-fold cross validation, \ie, 
graphs are split into five folds, which are (approximately) of the same size, and balanced in terms of graph domains, 
and then as each fold (20\%) is held out to be used for testing, the other folds (80\%) are used for model training.
Note that graph splits are used to split the performance matrix $ \mP $ and meta-graph features $ \mM $.

\textbf{Sparse Testbed.} The performance matrix $ \mP $ in this setting is sparse and partially observed, 
\ie, we may only have a few observations for each graph.
This setting is important since it can be costly to add a new model to the benchmark, 
which requires training and evaluating the model multiple times on the graphs in the benchmark.
By dealing with model selection using a sparse $ \mP $,
this testbed addresses significant practical considerations, 
\eg, making it more cost-effective to be able to add new models to the benchmark.
Using this testbed, researchers can develop and test specialized algorithms capable of learning from such partially-observed performances.
To construct a sparse performance matrix $ \mP^{\prime} $, we sample uniformly at random $ pm $ values from each row of $ \mP $,
where $ p $ is the fraction of values to sample and $ m $ is the total number of models.
This graph-wise sampling strategy ensures the same number of observations for each graph,
which matches the practical motivation that we have a limited budget per graph.
For this benchmark, we use different sparsity levels $p \in \{0.1,0.3,0.5,0.7,0.9\}$.

\textit{Splitting:} We use the same stratified 5-fold cross validation as in Fully-Observed testbed.
Algorithms are trained using a sparse split of $ \mP $, and evaluated with fully-observed performances of the test split.

\textbf{Out-Of-Domain Testbed.}
Graphs from a particular domain (\eg, road graphs, social networks, and brain networks) often have similar characteristics to each other.
In other words, graphs from a certain domain can be considered as having its own distribution,
which makes model selection for graphs from a new domain a challenging task.
This testbed evaluates the effectiveness of model selection methods for such an out-of-distribution setting
by holding out graphs from a specific network domain, and trying to predict for the held-out domain by learning from graphs from all the other domains.

\textit{Splitting:} We use a group-based 5-fold cross validation for this testbed such that each domain appears once in the test set across all folds.

\textbf{Small-To-Large Testbed.}
Training a GL model can take a lot of time and resources, especially for large-scale graphs.
While model selection methods may benefit from having more prior performances, 
having to obtain performance records for large graphs presents a significant computational bottleneck.
The meta-training process can be made significantly faster by enabling model selection algorithms 
to learn from relatively small graphs to be able to predict for larger graphs.
This testbed focuses on this challenging yet practical setting, which evaluates the ability to generalize from small to large graphs.

\textit{Splitting:} Graphs with less than $ \epsilon $ nodes form a small-graph set used for training.
The other graphs with at least $ \epsilon $ nodes form a large-graph set, which is used for evaluation.
We evaluate using a threshold value $ \epsilon $ of 10000 for this testbed.

\textbf{Cross-Task Testbed.}
The above testbeds operate on the model performances measured for one specific type of GL task.
By contrast, in this testbed, 
model selection methods learn from performances of one GL task (\eg, node classification), and 
are evaluated by predicting performances of a different GL task (\eg, link prediction).
This task present an additional challenge to model the relation between two different, yet related GL tasks, and 
utilize the learned knowledge for transferable model~selection.

\textit{Splitting:} 
We first choose the source and target tasks, and split the graphs into the two sets, \ie, the source task set and the target task set.
Then the graphs in the source set are used for training, and 
the graphs in the target set are used for testing.

\vspace{-1.0em}
\subsection{Model Selection Algorithms}\label{sec:baselines}
\vspace{-0.5em}

\benchmark provides state-of-the art algorithms for instantaneous model selection, which are listed in~\Cref{tab:baselines}.
These algorithms are selected such that the benchmark covers representative techniques for model selection,
in terms of whether they use meta-graph features (C1, \Cref{sec:meta-graph-features}) 
and prior model performances (C2, \Cref{sec:data}), and 
whether they are optimizable with trainable parameters~(C3).
\textit{Random Selection (\rs)} is used as a baseline to see how well model selection algorithms perform in comparison to random scoring.
\textit{Global Best (GB)-AvgPerf} and \textit{GB-AvgRank} select a model~that performed globally well on average.
In contrast, \textit{\isac}~\cite{DBLP:conf/ecai/KadiogluMST10} and \textit{\asfull (\as)}~\cite{nikolic2013simple} 
perform model selection more locally with respect to the given graph, using meta-features.
As GB methods rely only on prior performance, 
comparisons against them can help with investigating the effectiveness of meta-graph features.
\textit{Supervised Surrogates (\ss)}~\cite{xu2012satzilla2012}, \textit{\alors}~\cite{DBLP:journals/ai/MisirS17}, \textit{\ncf}~\cite{DBLP:conf/www/HeLZNHC17}, \textit{\metaod}~\cite{DBLP:conf/nips/ZhaoRA21}, and \textit{\metagl}~\cite{park2023metagl}
are optimizable algorithms,
which learn to estimate model performance by capturing the relation between meta-features and observed performances.
In comparison to the simpler, non-optimizable algorithms above, 
we can investigate the advantages of different optimization components for instantaneous model selection.
\Cref{app:algorithms} provides a more detailed description of each algorithm.

\begin{table}[!t]\centering
	\par\vspace{-1.0em}\par
	\setlength{\abovecaptionskip}{0.5em}
	\caption{\benchmark provides representative algorithms for instantaneous model selection. Algorithm characteristics denote whether they utilize meta-graph features (C1) and observed model performances (C2) for model selection, and whether they are optimizable (\ie, have trainable parameters)~(C3).}\label{tab:baselines}
	\setlength{\tabcolsep}{2pt}
\fontsize{7.5}{8.5}\selectfont
	\makebox[0.4\textwidth][c]{
		\resizebox{1.00\textwidth}{!}{\begin{tabular}{lcccccccccc}\toprule
				\diagbox{\textbf{Characteristics}}{\textbf{Algorithm}} &\makecell[c]{\textbf{Random}\\\textbf{Selection}} &\makecell[c]{\textbf{GB-Avg}\\\textbf{Perf}~\cite{DBLP:conf/nips/ZhaoRA21}} &\makecell[c]{\textbf{GB-Avg}\\\textbf{Rank}~\cite{park2023metagl}} &\makecell[c]{\textbf{\isac}\\\cite{DBLP:conf/ecai/KadiogluMST10}} &\makecell[c]{\textbf{\as}\\\cite{nikolic2013simple}} &\makecell[c]{\textbf{Spv. Surro.}\\\textbf{(S2)}~\cite{xu2012satzilla2012}} &\makecell[c]{\textbf{\alors}\\\cite{DBLP:journals/ai/MisirS17}} &\makecell[c]{\textbf{\ncf}\\\cite{DBLP:conf/www/HeLZNHC17}} &\makecell[c]{\textbf{\metaod}\\\cite{DBLP:conf/nips/ZhaoRA21}} &\makecell[c]{\textbf{\metagl}\\\cite{park2023metagl}} \\\midrule
				\textbf{C1. Use meta-features} 		& \xmark & \xmark & \xmark & \Checkmark & \Checkmark &\Checkmark &\Checkmark &\Checkmark &\Checkmark &\Checkmark \\
				\textbf{C2. Use prior performances}	& \xmark & \Checkmark & \Checkmark & \Checkmark & \Checkmark &\Checkmark &\Checkmark &\Checkmark &\Checkmark &\Checkmark \\
				\textbf{C3. Optimizable}				& \xmark & \xmark & \xmark & \xmark & \xmark &\Checkmark &\Checkmark &\Checkmark &\Checkmark &\Checkmark \\
				\bottomrule
			\end{tabular}
		
}}
	\vspace{1.0em}
\end{table}

\vspace{-1.2em}
\section{Experiments}\label{sec:exp}
\vspace{-0.8em}

In this section, we report how model selection methods perform in different testbeds. 
Based on those observations, we discuss the limitations of existing methods and future research directions.

\vspace{-1.0em}
\subsection{Model Selection Performance}\label{sec:exp:results}
\vspace{-0.6em}

\textbf{Evaluation Protocol.}
To measure how well model selection methods perform on the testbeds presented in~\Cref{sec:testbeds},
we evaluate their top-1 prediction results (\ie, the model predicted to be the best for the query graph)
as model selection aims to find the best performing model as accurately as possible.
Specifically, top-1 prediction performance is measured
in terms of AUC, MAP (mean average precision), and NDCG (normalized discounted cumulative gain), 
all of which range from zero to one, with larger values indicating a better performance.
We apply AUC and MAP by treating the task as a binary classification problem,
in which the top-1 model is labeled as one, and all other models are labeled as zero.
For NDCG, we report NDCG@1, which evaluates the ranking quality of the top-1 model.\linebreak
We evaluate these metrics multiple times for the data splits each testbed provides, and report the averaged performance.
For reproducibility, \benchmark provides the data splits of all testbeds.

\begin{table}[!t]\centering
	\par\vspace{-1.5em}\par
	\setlength{\abovecaptionskip}{-0.2em}
	\caption{Fully-Observed testbed results for link prediction (top) and node classification (bottom)~tasks.
		Higher~($ \uparrow $) scores are better. The \textbf{best} result is in bold, and the \underline{second best} result is underlined.}
	\label{tab:results:fully-observed}
	\setlength{\tabcolsep}{3pt}
\fontsize{7.5}{8.5}\selectfont
	
\begin{subtable}[h]{1.0\textwidth}
	\setlength{\abovecaptionskip}{-0.5em}
	\subcaption{Link prediction}
	\label{tab:results:fully-observed:linkpred}
	\begin{center}
		
		\begin{tabular}{lcccccccccc}\toprule
			\textbf{Perf. Metric}&\makecell[c]{\textbf{\rs}} &\makecell[c]{\textbf{\gbperf}} &\makecell[c]{\textbf{\gbrank}} &\makecell[c]{\textbf{\isac}} &\makecell[c]{\textbf{\as}} &\makecell[c]{\textbf{\ss}} &\makecell[c]{\textbf{\alors}} &\makecell[c]{\textbf{\ncf}} &\makecell[c]{\textbf{\metaod}} &\makecell[c]{\textbf{\metagl}} \\\midrule
			\textbf{AUC} ($ \uparrow $)	& 0.524 & 0.735 & 0.730 & 0.807 & \underline{0.864} & 0.809 & 0.843 & 0.728 & 0.764 & \textbf{0.875} \\ 
			\textbf{MRR} ($ \uparrow $) & 0.016 & 0.087 & 0.064 & 0.134 & \textbf{0.371} & 0.198 & 0.201 & 0.073 & 0.096 & \underline{0.295} \\ 
			\textbf{NDCG@1} ($ \uparrow $) & 0.813 & 0.942 & 0.934 & 0.944 & 0.957 & 0.950 & \underline{0.961} & 0.943 & 0.937 & \textbf{0.969} \\
			\bottomrule
		\end{tabular}
		
		\vspace{-0.5em}
	\end{center}
\end{subtable}
\begin{subtable}[h]{1.0\textwidth}
	\setlength{\abovecaptionskip}{-0.5em}
	\subcaption{Node classification}
	\label{tab:results:fully-observed:nodeclass}
	\begin{center}
		
		\begin{tabular}{lcccccccccc}\toprule
			\textbf{Perf. Metric}&\makecell[c]{\textbf{\rs}} &\makecell[c]{\textbf{\gbperf}} &\makecell[c]{\textbf{\gbrank}} &\makecell[c]{\textbf{\isac}} &\makecell[c]{\textbf{\as}} &\makecell[c]{\textbf{\ss}} &\makecell[c]{\textbf{\alors}} &\makecell[c]{\textbf{\ncf}} &\makecell[c]{\textbf{\metaod}} &\makecell[c]{\textbf{\metagl}} \\\midrule
			\textbf{AUC} ($ \uparrow $)	& 0.518 & 0.747 & 0.744 & 0.746 & \underline{0.762} & \textbf{0.772} & 0.734 & 0.745 & 0.581 & 0.740 \\ 
			\textbf{MRR} ($ \uparrow $) & 0.029 & 0.102 & 0.124 & 0.118 & \textbf{0.181} & 0.110 & 0.103 & 0.124 & 0.041 & \underline{0.129} \\ 
			\textbf{NDCG@1} ($ \uparrow $) & 0.747 & 0.865 & 0.860 & 0.885 & \underline{0.892} & \textbf{0.916} & 0.886 & 0.883 & 0.839 & 0.863 \\
			\bottomrule
		\end{tabular}
		
	\end{center}
\end{subtable}
\vspace{-0.5em}
\end{table}

\begin{table}[!t]\centering
\setlength{\abovecaptionskip}{-0.3em}
	\caption{Sparse testbed results for link prediction (top) and node classification (bottom) tasks.
		Higher~($ \uparrow $) scores are better. The \textbf{best} result is in bold, and the \underline{second best} result is underlined.
	}
	\label{tab:results:sparse}
	\setlength{\tabcolsep}{3pt}
\fontsize{7.5}{8.5}\selectfont
	
	\begin{subtable}[h]{1.0\textwidth}
		\setlength{\abovecaptionskip}{-0.5em}
		\subcaption{Link prediction}
		\label{tab:results:sparse:linkpred}
		\begin{center}
			
			\begin{tabular}{cccccccccccc}\toprule
				\textbf{Perf. Metric} & \textbf{Sparsity} &\makecell[c]{\textbf{\rs}} &\makecell[c]{\textbf{\gbperf}} &\makecell[c]{\textbf{\gbrank}} &\makecell[c]{\textbf{\isac}} &\makecell[c]{\textbf{\as}} &\makecell[c]{\textbf{\ss}} &\makecell[c]{\textbf{\alors}} &\makecell[c]{\textbf{\ncf}} &\makecell[c]{\textbf{\metaod}} &\makecell[c]{\textbf{\metagl}} \\\midrule
				\multirow{5}{*}{\textbf{AUC} ($ \uparrow $)} & 10\%	& 0.524 & 0.733 & 0.732 & 0.804 & 0.829 & 0.813 & \underline{0.831} & 0.735 & 0.743 & \textbf{0.865} \\ 
				& 30\% & 0.524 & 0.728 & 0.738 & 0.798 & 0.763 & 0.811 & \underline{0.827} & 0.739 & 0.703 & \textbf{0.871} \\ 
				& 50\% & 0.524 & 0.704 & 0.730 & 0.790 & 0.690 & \underline{0.839} & 0.814 & 0.739 & 0.669 & \textbf{0.866} \\ 
				& 70\% & 0.524 & 0.708 & 0.730 & 0.778 & 0.618 & \underline{0.814} & 0.795 & 0.757 & 0.630 & \textbf{0.866} \\ 
				& 90\% & 0.524 & 0.717 & \underline{0.732} & 0.720 & 0.547 & 0.464 & 0.687 & 0.656 & 0.599 & \textbf{0.811} \\
				\bottomrule
			\end{tabular}
			
			\vspace{-0.5em}
		\end{center}
	\end{subtable}
	\begin{subtable}[h]{1.0\textwidth}
		\setlength{\abovecaptionskip}{-0.5em}
		\subcaption{Node classification}
		\label{tab:results:sparse:nodeclass}
		\begin{center}
			
			\begin{tabular}{cccccccccccc}\toprule
				\textbf{Perf. Metric} & \textbf{Sparsity} &\makecell[c]{\textbf{\rs}} &\makecell[c]{\textbf{\gbperf}} &\makecell[c]{\textbf{\gbrank}} &\makecell[c]{\textbf{\isac}} &\makecell[c]{\textbf{\as}} &\makecell[c]{\textbf{\ss}} &\makecell[c]{\textbf{\alors}} &\makecell[c]{\textbf{\ncf}} &\makecell[c]{\textbf{\metaod}} &\makecell[c]{\textbf{\metagl}} \\\midrule
				\multirow{5}{*}{\textbf{AUC} ($ \uparrow $)} & 10\%	& 0.518 & 0.746 & 0.744 & 0.744 & 0.748 & \textbf{0.766} & 0.727 & 0.727 & 0.575 & \underline{0.761} \\ 
				& 30\% & 0.518 & \underline{0.743} & 0.738 & 0.734 & 0.680 & \textbf{0.769} & 0.741 & 0.735 & 0.533 & 0.736 \\ 
				& 50\% & 0.518 & 0.726 & \textbf{0.739} & 0.687 & 0.592 & \underline{0.739} & 0.730 & 0.713 & 0.485 & 0.709 \\ 
				& 70\% & 0.518 & 0.692 & \textbf{0.738} & 0.653 & 0.571 & 0.684 & 0.694 & \underline{0.709} & 0.483 & 0.662 \\ 
				& 90\% & 0.518 & 0.626 & \textbf{0.697} & 0.592 & 0.535 & 0.620 & 0.654 & \underline{0.660} & 0.490 & 0.659 \\
				\bottomrule
			\end{tabular}
			
		\end{center}
	\end{subtable}
	\vspace{-2.0em}
\end{table}

\textbf{Fully-Observed Testbed} (\Cref{tab:results:fully-observed}).
Comparison between methods where meta-graph features are either used (\eg, \as, \metagl) or not used (\eg, \gbperf)
shows the benefits of utilizing meta-graph features for GL model selection.
While optimizable methods (\eg, \ncf, \metaod) have the additional flexibility to adaptively tune their behavior based on data,
they are outperformed by relatively simple methods like \isac and \as.
At the same time, the best results on link prediction in the majority of metrics are achieved by another optimizable method, \metagl,
which shows the promising potential of optimizable framework for model selection.
In node classification results, the performance decrease of optimizable methods are notable (\eg, \metagl).
One potential reason for this is that the graph set for node classification is relatively small 
compared to the graphs applicable for link prediction, which limits the effectiveness of optimizable algorithms that are more prone to overfitting in such cases.

\textbf{Sparse Testbed} (\Cref{tab:results:sparse}).
As the sparsity of the performance matrix $ \mP $ increases, model selection methods perform increasingly worse.
In particular, while \as achieves the best or the second best results in the Fully-Observed testbed, 
its performance quickly declines as sparsity increases. Since \as performs model selection based on the most similar observed graph, 
it cannot operate effectively in a highly sparse setting.
Global averaging methods (\eg, \gbperf), or more sophisticated optimizable methods show more stable results.
Due to the additional requirement for node labels, 
node classification task in this setup presents the most data sparse, yet practically important regime.

\textbf{Out-Of-Domain Testbed} (\Cref{tab:results:out-of-domain}).
Graphs in the same or similar domains are often more similar to each other than graphs in different domains.
As this testbed requires addressing additional challenges to achieve out-of-distribution generalization, 
most methods perform worse than in other in-distribution testbeds.
For instance, \as, which are sensitive to the availability of observed graphs similar to the query graph, perform worse than in \Cref{tab:results:fully-observed}.
On the other hand, optimizable methods show more promising results as they learn to extrapolate into new domains by learning from observed domains.

\textbf{Small-To-Large Testbed} (\Cref{tab:results:small-to-large}).
In comparison to the Fully-Observed testbed, the performance decreases overall in this testbed. 
However, considering that methods learn only from small graphs, 
model selection for large graphs still performs quite well, often achieving a similar level of performance.
Successful methods in this testbed can make the model selection pipeline much more efficient 
as performance collection for small graphs can be done much more efficiently than for large graphs.

\textbf{Additional Results.}
We provide additional results in \Cref{app:results}, including the results of the Cross-Task testbed, 
and results obtained with other meta-graph features, \eg, \textit{M\textsubscript{graphlets}} and \textit{M\textsubscript{compact}}.

\begin{table}[!t]\centering
\setlength{\abovecaptionskip}{-0.5em}
	\caption{Out-Of-Domain testbed results for link prediction (top) and node classification (bottom)~tasks.
		Higher~($ \uparrow $) scores are better. The \textbf{best} result is in bold, and the \underline{second best} result is underlined.}
	\label{tab:results:out-of-domain}
	\setlength{\tabcolsep}{3pt}
\fontsize{7.5}{8.5}\selectfont
	
\begin{subtable}[h]{1.0\textwidth}
	\setlength{\abovecaptionskip}{-0.5em}
	\subcaption{Link prediction}
	\label{tab:results:out-of-domain:linkpred}
	\begin{center}
		\begin{tabular}{lcccccccccc}\toprule
			\textbf{Perf. Metric}&\makecell[c]{\textbf{\rs}} &\makecell[c]{\textbf{\gbperf}} &\makecell[c]{\textbf{\gbrank}} &\makecell[c]{\textbf{\isac}} &\makecell[c]{\textbf{\as}} &\makecell[c]{\textbf{\ss}} &\makecell[c]{\textbf{\alors}} &\makecell[c]{\textbf{\ncf}} &\makecell[c]{\textbf{\metaod}} &\makecell[c]{\textbf{\metagl}} \\\midrule
			\textbf{AUC} ($ \uparrow $)	& 0.517 & 0.809 & 0.811 & \underline{0.850} & 0.786 & 0.837 & 0.820 & 0.837 & 0.681 & \textbf{0.871} \\ 
			\textbf{MRR} ($ \uparrow $) & 0.018 & 0.110 & 0.101 & 0.125 & \textbf{0.237} & 0.116 & 0.109 & 0.109 & 0.047 & \underline{0.148} \\ 
			\textbf{NDCG@1} ($ \uparrow $) & 0.820 & \textbf{0.956} & \underline{0.954} & 0.951 & 0.935 & 0.953 & 0.953 & 0.952 & 0.918 & 0.951 \\
			\bottomrule
		\end{tabular}
		\vspace{-0.5em}
	\end{center}
\end{subtable}
\begin{subtable}[h]{1.0\textwidth}
	\setlength{\abovecaptionskip}{-0.5em}
	\subcaption{Node classification}
	\label{tab:results:out-of-domain:nodeclass}
	\begin{center}
		\begin{tabular}{lcccccccccc}\toprule
			\textbf{Perf. Metric}&\makecell[c]{\textbf{\rs}} &\makecell[c]{\textbf{\gbperf}} &\makecell[c]{\textbf{\gbrank}} &\makecell[c]{\textbf{\isac}} &\makecell[c]{\textbf{\as}} &\makecell[c]{\textbf{\ss}} &\makecell[c]{\textbf{\alors}} &\makecell[c]{\textbf{\ncf}} &\makecell[c]{\textbf{\metaod}} &\makecell[c]{\textbf{\metagl}} \\\midrule
			\textbf{AUC} ($ \uparrow $)	& 0.495 & 0.726 & 0.727 & 0.701 & 0.684 & \textbf{0.750} & 0.668 & \underline{0.741} & 0.571 & 0.705 \\ 
			\textbf{MRR} ($ \uparrow $) & 0.019 & 0.074 & \underline{0.086} & 0.046 & 0.060 & \textbf{0.089} & 0.056 & 0.066 & 0.044 & 0.082 \\ 
			\textbf{NDCG@1} ($ \uparrow $) & 0.722 & 0.828 & 0.836 & \underline{0.848} & 0.828 & \textbf{0.901} & 0.796 & 0.842 & 0.810 & 0.848 \\
			\bottomrule
		\end{tabular}
	\end{center}
\end{subtable}
\vspace{-0.75em}
\end{table}

\begin{table}[!t]\centering
	\par\vspace{-1.0em}\par
	\setlength{\abovecaptionskip}{-0.5em}
	\caption{Small-To-Large testbed results for link prediction (top) and node classification (bottom)~tasks.
		Higher~($ \uparrow $) scores are better. The \textbf{best} result is in bold, and the \underline{second best} result is underlined.}
	\label{tab:results:small-to-large}
	\setlength{\tabcolsep}{3pt}
\fontsize{7.5}{8.5}\selectfont
	
\begin{subtable}[h]{1.0\textwidth}
	\setlength{\abovecaptionskip}{-0.5em}
	\subcaption{Link prediction}
	\label{tab:results:small-to-large:linkpred}
	\begin{center}
		\begin{tabular}{lcccccccccc}\toprule
			\textbf{Perf. Metric}&\makecell[c]{\textbf{\rs}} &\makecell[c]{\textbf{\gbperf}} &\makecell[c]{\textbf{\gbrank}} &\makecell[c]{\textbf{\isac}} &\makecell[c]{\textbf{\as}} &\makecell[c]{\textbf{\ss}} &\makecell[c]{\textbf{\alors}} &\makecell[c]{\textbf{\ncf}} &\makecell[c]{\textbf{\metaod}} &\makecell[c]{\textbf{\metagl}} \\\midrule
			\textbf{AUC} ($ \uparrow $)	& 0.522 & 0.798 & 0.797 & \underline{0.842} & 0.827 & 0.767 & 0.812 & 0.796 & 0.667 & \textbf{0.870} \\ 
			\textbf{MRR} ($ \uparrow $) & 0.031 & 0.072 & 0.061 & 0.132 & \textbf{0.368} & 0.074 & 0.209 & 0.047 & 0.075 & \underline{0.260} \\ 
			\textbf{NDCG@1} ($ \uparrow $) & 0.841 & 0.958 & \underline{0.960} & 0.957 & 0.951 & 0.953 & 0.947 & 0.956 & 0.921 & \textbf{0.964} \\
			\bottomrule
		\end{tabular}
		\vspace{-0.5em}
	\end{center}
\end{subtable}
\begin{subtable}[h]{1.0\textwidth}
	\setlength{\abovecaptionskip}{-0.5em}
	\subcaption{Node classification}
	\label{tab:results:small-to-large:nodeclass}
	\begin{center}
		\begin{tabular}{lcccccccccc}\toprule
			\textbf{Perf. Metric}&\makecell[c]{\textbf{\rs}} &\makecell[c]{\textbf{\gbperf}} &\makecell[c]{\textbf{\gbrank}} &\makecell[c]{\textbf{\isac}} &\makecell[c]{\textbf{\as}} &\makecell[c]{\textbf{\ss}} &\makecell[c]{\textbf{\alors}} &\makecell[c]{\textbf{\ncf}} &\makecell[c]{\textbf{\metaod}} &\makecell[c]{\textbf{\metagl}} \\\midrule
			\textbf{AUC} ($ \uparrow $)	& 0.508 & 0.724 & 0.726 & 0.711 & \textbf{0.761} & 0.664 & 0.701 & 0.697 & 0.467 & \underline{0.736} \\ 
			\textbf{MRR} ($ \uparrow $) & 0.011 & 0.058 & 0.082 & \textbf{0.109} & \underline{0.095} & 0.036 & 0.042 & 0.034 & 0.016 & 0.071 \\ 
			\textbf{NDCG@1} ($ \uparrow $) & 0.795 & 0.861 & \underline{0.896} & \textbf{0.902} & 0.883 & 0.855 & 0.862 & 0.864 & 0.830 & 0.857 \\
			\bottomrule
		\end{tabular}
	\end{center}
\end{subtable}
\vspace{-1.0em}
\end{table}

\subsection{Discussion on Limitations and Future Directions}\label{sec:exp:discussion}
\vspace{-0.4em}

\textbf{Limitations.} In principle, using \benchmark to select a GL model to employ for a new graph is 
based on the assumption that similar graph datasets exist in the benchmark. 
Therefore, it may not be very effective if the new graph is significantly different from all graphs in the benchmark 
(\eg, the new graph is from a completely new domain).
However, as the benchmark data continue to grow over time, such cases will be increasingly less likely, 
while model selection performances will likely improve with the addition of more data.
Furthermore, while \benchmark currently supports two fundamental GL tasks, namely, node classification and link prediction,
it can be further extended with additional tasks (\eg, graph classification).
Incorporating them into \benchmark is one of the future plans.

\textbf{Future Directions.} 
Below we list promising research directions to further improve the algorithms as well as the benchmark for instantaneous GL model selection.

\setlength{\parskip}{0.15\baselineskip}

$\sq~$\textit{\textbf{Direction 1: enabling model selection methods to use additional graph data.}}
While existing methods utilize model performances and graph structural information captured by meta-features,
they currently do not take other available graph data into account, such as
node and edge features, timestamps in the case of dynamic graphs, and node and edge types (\eg, knowledge graphs).
These data can be useful for modeling graph similarities, and the benchmark can further be enriched with such additional data.

$\sq~$\textit{\textbf{Direction 2: developing data augmentation techniques.}}
Adding new performance records to the benchmark can improve the effectiveness of model selection methods.
However, it is often computationally expensive to train and evaluate GL models on non-trivial graphs.
Data augmentation techniques for GL model performances can be helpful in this data sparse regime,
especially for optimizable methods that require a lot of data to learn effectively.

$\sq~$\textit{\textbf{Direction 3: handling out-of-distribution settings.}}
Existing model selection methods are mainly designed for an in-distribution setup, 
as they assume that there exist observed graphs similar to a query graph.
Thus their performance is suboptimal when a query graph comes from a new~distribution.
Investigating how to achieve generalization in such an out-of-distribution scenario would be beneficial.

$\sq~$\textit{\textbf{Direction 4: effective performance collection.}}
When we have a limited budget for performance measurements on new graphs, 
selecting which pairs of graphs and models to evaluate and include~in~the benchmark
can greatly influence the learning of model selection methods.
Thus the ability to find a small set of representative pairs can lead to a fast and effective performance collection.
Challenges include how to make such selections from a heterogeneous model set with multiple GL~methods.

\vspace{-0.6em}
\section{Conclusion}\label{sec:conclusion}
\vspace{-0.4em}

The choice of a GL model has a significant impact on the performance of downstream tasks.
Despite recent efforts to tackle this important problem, there exists no benchmark environment
to evaluate the performance of GL model selection methods, and to support the development of new methods.
In this work, we develop \benchmark, the first benchmark environment for instantaneous GL model selection.
\begin{itemize}[leftmargin=0.90em,nosep,topsep=-0.4em]
\item \textbf{Extensive Benchmark Data.} 
Among others, \benchmark provides an extensive collection of model performances on fundamental GL tasks,
\ie, link prediction and node classification, which is by far the largest and most comprehensive benchmark for Prob.~\ref{prob} to the best of our knowledge.

\item \textbf{Algorithms and Testbeds.}
\benchmark provides representative algorithms for Prob.~\ref{prob},
as well as multiple testbeds to assess model selection performance in practical usage scenarios.

\item \textbf{Extensible Open Source Environment.}
\benchmark is designed to be easily extended with new GL models, new graphs, new performance records, and new GL tasks, while allowing reproducibility.

\end{itemize}

{

\bibliographystyle{plain}
}

\clearpage
\appendix

In the appendix, we provide 
additional experimental results (\Cref{app:results});
details of model selection algorithms (\Cref{app:algorithms}), and meta-graph features (\Cref{app:metafeats});
experimental settings (\Cref{app:expsettings});
a discussion on the usage and extensibility of \benchmark (\Cref{app:extensibility});
details of the data \benchmark provides (\Cref{app:data}); and 
hosting, licensing, and maintenance plan (\Cref{app:hlm}).

\section{Additional Results}\label{app:results}
\subsection{Cross-Task Testbed Results}\label{app:results:crosstask}

In \Cref{appendix:tab:results:cross-task}, we report the cross-task testbed results in two transfer learning settings, \ie,
(a) node~classification to link prediction (\Cref{tab:results:cross-task:node_to_link}) and (b) link prediction to node classification (\Cref{tab:results:cross-task:link_to_node}).
Compared to other testbeds that operate on the performances measured for only one type of GL~task, 
nearly all methods exhibit performance decline in this challenging setup,
which indicates that GL models that are good for one type of task may not be as effective for another type of GL task.
In contrast to other testbeds, sophisticated algorithms (\eg, \metagl) tend to experience more performance decrease in this testbed 
than simple averaging methods (\eg, \gbrank), which perform close to the best.
By designing mechanisms that can model how performance characteristics on one task would translate to those on another,
optimizable algorithms can be made much more effective in this setting.

\begin{table}[!htbp]\centering
\setlength{\abovecaptionskip}{-0.2em}
	\caption{Cross-Task testbed results for node classification-to-link prediction (top) and link prediction-to-node classification (bottom) settings.
		Higher~($ \uparrow $) scores are better. The numbers in the parentheses denote one standard error.
		The \textbf{best} result is in bold, and the \underline{second best} result is underlined.}
	\label{appendix:tab:results:cross-task}
	\setlength{\tabcolsep}{3pt}
\fontsize{7.5}{8.5}\selectfont
	
	\begin{subtable}[h]{1.0\textwidth}
		\setlength{\abovecaptionskip}{-0.5em}
		\subcaption{Node classification $ \rightarrow $ Link prediction}
		\label{tab:results:cross-task:node_to_link}
		\begin{center}
			\begin{tabular}{lcccccccccc}\toprule
				\textbf{Perf. Metric}&\makecell[c]{\textbf{\rs}} &\makecell[c]{\textbf{\gbperf}} &\makecell[c]{\textbf{\gbrank}} &\makecell[c]{\textbf{\isac}} &\makecell[c]{\textbf{\as}} &\makecell[c]{\textbf{\ss}} &\makecell[c]{\textbf{\alors}} &\makecell[c]{\textbf{\ncf}} &\makecell[c]{\textbf{\metaod}} &\makecell[c]{\textbf{\metagl}} \\\midrule
				\textbf{AUC} ($ \uparrow $)	& \makecell{0.479\\(0.016)} & \makecell{\underline{0.652}\\(0.013)} & \makecell{\textbf{0.671}\\(0.013)} & \makecell{0.626\\(0.014)} & \makecell{0.520\\(0.015)} & \makecell{0.636\\(0.014)} & \makecell{0.594\\(0.015)} & \makecell{0.650\\(0.013)} & \makecell{0.465\\(0.015)} & \makecell{0.553\\(0.014)} \\ \midrule[0.2pt]
				\textbf{MRR} ($ \uparrow $) & \makecell{0.020\\(0.003)} & \makecell{0.022\\(0.002)} & \makecell{0.022\\(0.002)} & \makecell{0.020\\(0.002)} & \makecell{\underline{0.026}\\(0.006)} & \makecell{0.022\\(0.002)} & \makecell{\textbf{0.029}\\(0.005)} & \makecell{0.024\\(0.002)} & \makecell{0.015\\(0.003)} & \makecell{0.018\\(0.002)} \\ \midrule[0.2pt]
				\textbf{NDCG@1} ($ \uparrow $) & \makecell{0.830\\(0.008)} & \makecell{0.806\\(0.008)} & \makecell{\underline{0.848}\\(0.006)} & \makecell{0.833\\(0.007)} & \makecell{0.825\\(0.008)} & \makecell{0.815\\(0.007)} & \makecell{\textbf{0.858}\\(0.007)} & \makecell{0.812\\(0.008)} & \makecell{0.807\\(0.008)} & \makecell{0.827\\(0.008)} \\
				\bottomrule
			\end{tabular}
			
			\vspace{-0.5em}
		\end{center}
	\end{subtable}
	\begin{subtable}[h]{1.0\textwidth}
		\setlength{\abovecaptionskip}{-0.5em}
		\subcaption{Link prediction $ \rightarrow $ Node classification}
		\label{tab:results:cross-task:link_to_node}
		\begin{center}
			
			\begin{tabular}{lcccccccccc}\toprule
				\textbf{Perf. Metric}&\makecell[c]{\textbf{\rs}} &\makecell[c]{\textbf{\gbperf}} &\makecell[c]{\textbf{\gbrank}} &\makecell[c]{\textbf{\isac}} &\makecell[c]{\textbf{\as}} &\makecell[c]{\textbf{\ss}} &\makecell[c]{\textbf{\alors}} &\makecell[c]{\textbf{\ncf}} &\makecell[c]{\textbf{\metaod}} &\makecell[c]{\textbf{\metagl}} \\\midrule
				\textbf{AUC} ($ \uparrow $)	& \makecell{0.542\\(0.025)} & \makecell{0.595\\(0.025)} & \makecell{0.596\\(0.025)} & \makecell{0.581\\(0.026)} & \makecell{0.517\\(0.025)} & \makecell{0.536\\(0.024)} & \makecell{\underline{0.606}\\(0.025)} & \makecell{\textbf{0.608}\\(0.025)} & \makecell{0.477\\(0.024)} & \makecell{0.542\\(0.026)} \\ \midrule[0.2pt]
				\textbf{MRR} ($ \uparrow $) & \makecell{0.023\\(0.008)} & \makecell{\underline{0.053}\\(0.016)} & \makecell{0.047\\(0.011)} & \makecell{0.043\\(0.012)} & \makecell{0.027\\(0.009)} & \makecell{0.034\\(0.010)} & \makecell{0.032\\(0.006)} & \makecell{\textbf{0.057}\\(0.016)} & \makecell{0.011\\(0.001)} & \makecell{0.038\\(0.012)} \\ \midrule[0.2pt]
				\textbf{NDCG@1} ($ \uparrow $) & \makecell{0.731\\(0.022)} & \makecell{0.817\\(0.018)} & \makecell{\underline{0.822}\\(0.019)} & \makecell{0.795\\(0.021)} & \makecell{0.780\\(0.020)} & \makecell{0.789\\(0.021)} & \makecell{0.747\\(0.022)} & \makecell{\textbf{0.823}\\(0.017)} & \makecell{0.730\\(0.021)} & \makecell{0.779\\(0.021)} \\
				\bottomrule
			\end{tabular}
			
		\end{center}
	\end{subtable}
	
\end{table}

\subsection{Results With Different Sets of Meta-Graph Features}\label{app:results:metagraphfeats}

{\setlength{\parskip}{0.4\baselineskip}

We present three sets of meta-graph features in the main text, \ie, 
\Mregular, \Mgraphlets, and \Mcompact, which consist of different types of graph structural features.
Here we provide these different sets of meta-graph features to model selection algorithms, and evaluate their performance using each set.
In addition to the above three sets, we also use the concatenation of \Mregular and \Mgraphlets, denoted \Mreggraph, 
which augments the regular structural features with graphlet-based features, forming the largest set with 1074 meta-graph features.
In \Cref{appendix:tab:results:fully-observed:metafeats,appendix:tab:results:sparse:metafeats,appendix:tab:results:out-of-domain:metafeats,appendix:tab:results:small-to-large:metafeats,appendix:tab:results:cross-task:metafeats}, 
we report the performance of model selection algorithms in terms of their ROC AUC scores for the five testbeds.
Since the Random Selection and Global Best (GB) algorithms are independent of meta-features, 
their performances are the same across different features. From the results below, we make the following observations.

\textbf{\textit{Using additional meta-graph features can improve model selection results.}}
For instance, in 
Sparse testbed (\Cref{appendix:tab:results:sparse:metafeats}), 
the best performing method, \metagl, achieves the highest AUC by using an augmented feature set \Mreggraph.
As distinct graph features may capture different aspects of graph structural properties,
they can provide further information to find a better model.

\textbf{\textit{More features do not always lead to a better performance.}}
For example, in \Cref{appendix:tab:results:fully-observed:metafeats,appendix:tab:results:sparse:metafeats}, we see mixed results with optimizable methods (\eg, \ncf).
In some cases, they experience some~performance improvements, 
while in others, their performance declines as they use more features.
The capability to adaptively utilize meta-features for the given context could further improve their performances.

\textbf{\textit{The impact of different meta-graph features is more pronounced in the more challenging transfer learning settings}},
\ie, Out-Of-Domain~(\Cref{appendix:tab:results:out-of-domain:metafeats}), Small-To-Large~(\Cref{appendix:tab:results:small-to-large:metafeats}), and 
Cross-Task~(\Cref{appendix:tab:results:cross-task:metafeats}) testbeds.
These testbeds present additional challenges for model selection methods to achieve an effective generalization 
(\eg, large differences exist in graph data distributions or graph sizes between training and testing phases).
As existing methods do not take such challenges into account, 
they are prone to overfitting and thus may not generalize well in the testing phase. 
For example, the performance of \metagl in the Cross-Task testbed~(\Cref{appendix:tab:results:cross-task:metafeats}) is the lowest when using the largest feature set \Mreggraph.
A stronger and more robust transfer capability would be needed to enable a better use of additional meta-graph features in such cases.
On the other hand,
we also observe that using more meta-features leads to a significant performance improvement
for relatively simple methods,
\eg, \isac in Small-To-Large testbed (\Cref{appendix:tab:results:small-to-large:metafeats}),
which shows the promises and potential of meta-graph features 
to handle these challenging transfer learning settings.

\begin{table}[!t]\centering
\setlength{\abovecaptionskip}{-0.5em}
	\caption{Fully-Observed testbed results, obtained with different meta-graph features,
		for link prediction (top) and node classification (bottom) tasks.
		\textit{\textbf{M}\textsubscript{reg+graphlets}} denotes a concatenation of \textit{\textbf{M}\textsubscript{regular}} and \textit{\textbf{M}\textsubscript{graphlets}} meta-graph features.
		Higher~($ \uparrow $) scores are better. The numbers in the parentheses denote one standard error.
		The \textbf{best} result is in bold, and the \underline{second best} result is underlined.}
	\label{appendix:tab:results:fully-observed:metafeats}
	\setlength{\tabcolsep}{3pt}
\fontsize{7.5}{8.5}\selectfont
	
	\begin{subtable}[h]{1.0\textwidth}
		\setlength{\abovecaptionskip}{-0.5em}
		\subcaption{Link prediction}
		\label{appendix:tab:results:fully-observed:metafeats:linkpred}
		\begin{center}
			
			\begin{tabular}{cccccccccccc}\toprule
				\textbf{Perf. Metric} & \textbf{Meta-Feature} &\makecell[c]{\textbf{\rs}} &\makecell[c]{\textbf{\gbperf}} &\makecell[c]{\textbf{\gbrank}} &\makecell[c]{\textbf{\isac}} &\makecell[c]{\textbf{\as}} &\makecell[c]{\textbf{\ss}} &\makecell[c]{\textbf{\alors}} &\makecell[c]{\textbf{\ncf}} &\makecell[c]{\textbf{\metaod}} &\makecell[c]{\textbf{\metagl}} \\\midrule
				\multirow{8}{*}{\textbf{AUC} ($ \uparrow $)} & \textit{\textbf{M}\textsubscript{compact}} & \makecell{0.524\\(0.013)} & \makecell{0.735\\(0.011)} & \makecell{0.730\\(0.010)} & \makecell{0.757\\(0.011)} & \makecell{\underline{0.870}\\(0.010)} & \makecell{0.831\\(0.009)} & \makecell{0.847\\(0.009)} & \makecell{0.789\\(0.011)} & \makecell{0.726\\(0.015)} & \makecell{\textbf{0.875}\\(0.009)} \\ \cmidrule(lr){2-12}
				& \textit{\textbf{M}\textsubscript{regular}} & \makecell{0.524\\(0.013)} & \makecell{0.735\\(0.011)} & \makecell{0.730\\(0.010)} & \makecell{0.807\\(0.011)} & \makecell{\underline{0.864}\\(0.010)} & \makecell{0.809\\(0.011)} & \makecell{0.843\\(0.010)} & \makecell{0.728\\(0.011)} & \makecell{0.764\\(0.014)} & \makecell{\textbf{0.875}\\(0.009)} \\ \cmidrule(lr){2-12}
				& \textit{\textbf{M}\textsubscript{graphlets}} & \makecell{0.524\\(0.013)} & \makecell{0.735\\(0.011)} & \makecell{0.730\\(0.010)} & \makecell{0.781\\(0.011)} & \makecell{\underline{0.850}\\(0.011)} & \makecell{0.806\\(0.011)} & \makecell{0.830\\(0.010)} & \makecell{0.791\\(0.011)} & \makecell{0.740\\(0.014)} & \makecell{\textbf{0.873}\\(0.009)} \\ \cmidrule(lr){2-12}
				& \textit{\textbf{M}\textsubscript{reg+graphlets}} & \makecell{0.524\\(0.013)} & \makecell{0.735\\(0.011)} & \makecell{0.730\\(0.010)} & \makecell{0.803\\(0.011)} & \makecell{\underline{0.867}\\(0.010)} & \makecell{0.833\\(0.010)} & \makecell{0.843\\(0.010)} & \makecell{0.794\\(0.010)} & \makecell{0.740\\(0.014)} & \makecell{\textbf{0.874}\\(0.009)} \\
				\bottomrule
			\end{tabular}
			
			\vspace{-0.5em}
		\end{center}
	\end{subtable}
	\begin{subtable}[h]{1.0\textwidth}
		\setlength{\abovecaptionskip}{-0.5em}
		\subcaption{Node classification}
		\label{appendix:tab:results:fully-observed:metafeats:nodeclass}
		\begin{center}
			
			\begin{tabular}{cccccccccccc}\toprule
				\textbf{Perf. Metric} & \textbf{Meta-Feature} &\makecell[c]{\textbf{\rs}} &\makecell[c]{\textbf{\gbperf}} &\makecell[c]{\textbf{\gbrank}} &\makecell[c]{\textbf{\isac}} &\makecell[c]{\textbf{\as}} &\makecell[c]{\textbf{\ss}} &\makecell[c]{\textbf{\alors}} &\makecell[c]{\textbf{\ncf}} &\makecell[c]{\textbf{\metaod}} &\makecell[c]{\textbf{\metagl}} \\\midrule
				\multirow{8}{*}{\textbf{AUC} ($ \uparrow $)} & \textit{\textbf{M}\textsubscript{compact}} & \makecell{0.518\\(0.026)} & \makecell{0.747\\(0.024)} & \makecell{0.744\\(0.024)} & \makecell{0.749\\(0.024)} & \makecell{\textbf{0.786}\\(0.023)} & \makecell{\underline{0.775}\\(0.022)} & \makecell{0.763\\(0.023)} & \makecell{0.744\\(0.024)} & \makecell{0.602\\(0.029)} & \makecell{0.765\\(0.023)} \\ \cmidrule(lr){2-12}
				& \textit{\textbf{M}\textsubscript{regular}} & \makecell{0.518\\(0.026)} & \makecell{0.747\\(0.024)} & \makecell{0.744\\(0.024)} & \makecell{0.746\\(0.023)} & \makecell{\underline{0.762}\\(0.023)} & \makecell{\textbf{0.772}\\(0.022)} & \makecell{0.734\\(0.023)} & \makecell{0.745\\(0.025)} & \makecell{0.581\\(0.028)} & \makecell{0.740\\(0.024)} \\ \cmidrule(lr){2-12}
				& \textit{\textbf{M}\textsubscript{graphlets}} & \makecell{0.518\\(0.026)} & \makecell{\underline{0.747}\\(0.024)} & \makecell{0.744\\(0.024)} & \makecell{0.746\\(0.024)} & \makecell{0.729\\(0.026)} & \makecell{0.747\\(0.023)} & \makecell{0.715\\(0.025)} & \makecell{0.743\\(0.024)} & \makecell{0.629\\(0.029)} & \makecell{\textbf{0.763}\\(0.023)} \\ \cmidrule(lr){2-12}
				& \textit{\textbf{M}\textsubscript{reg+graphlets}} & \makecell{0.518\\(0.026)} & \makecell{\underline{0.747}\\(0.024)} & \makecell{0.744\\(0.024)} & \makecell{\textbf{0.758}\\(0.023)} & \makecell{0.728\\(0.026)} & \makecell{0.742\\(0.025)} & \makecell{0.744\\(0.023)} & \makecell{0.735\\(0.025)} & \makecell{0.600\\(0.029)} & \makecell{0.734\\(0.025)} \\
				\bottomrule
			\end{tabular}
			
		\end{center}
	\end{subtable}
	
\end{table}

\begin{table}[!htbp]\centering
\setlength{\abovecaptionskip}{-0.2em}
	\caption{Sparse testbed results, obtained with different meta-graph features, and performance matrices with a sparsity of $ 50\% $,
		for link prediction (top) and node classification (bottom) tasks.
		\textit{\textbf{M}\textsubscript{reg+graphlets}} denotes a concatenation of \textit{\textbf{M}\textsubscript{regular}} and \textit{\textbf{M}\textsubscript{graphlets}} meta-graph features.
		Higher~($ \uparrow $) scores are better. The numbers in the parentheses denote one standard error.
		The \textbf{best} result is in bold, and the \underline{second best} result is~underlined.}
	\label{appendix:tab:results:sparse:metafeats}
	\setlength{\tabcolsep}{3pt}
\fontsize{7.5}{8.5}\selectfont
	
	\begin{subtable}[h]{1.0\textwidth}
		\setlength{\abovecaptionskip}{-0.5em}
		\subcaption{Link prediction}
		\label{appendix:tab:results:sparse:metafeats:linkpred}
		\begin{center}
			\begin{tabular}{cccccccccccc}\toprule
				\textbf{Perf. Metric} & \textbf{Meta-Feature} &\makecell[c]{\textbf{\rs}} &\makecell[c]{\textbf{\gbperf}} &\makecell[c]{\textbf{\gbrank}} &\makecell[c]{\textbf{\isac}} &\makecell[c]{\textbf{\as}} &\makecell[c]{\textbf{\ss}} &\makecell[c]{\textbf{\alors}} &\makecell[c]{\textbf{\ncf}} &\makecell[c]{\textbf{\metaod}} &\makecell[c]{\textbf{\metagl}} \\\midrule
				\multirow{8}{*}{\textbf{AUC} ($ \uparrow $)} & \textit{\textbf{M}\textsubscript{compact}} & \makecell{0.524\\(0.013)} & \makecell{0.704\\(0.011)} & \makecell{0.730\\(0.010)} & \makecell{0.741\\(0.011)} & \makecell{0.682\\(0.012)} & \makecell{\underline{0.829}\\(0.009)} & \makecell{0.802\\(0.010)} & \makecell{0.780\\(0.011)} & \makecell{0.678\\(0.014)} & \makecell{\textbf{0.865}\\(0.010)} \\ \cmidrule(lr){2-12}
				& \textit{\textbf{M}\textsubscript{regular}} & \makecell{0.524\\(0.013)} & \makecell{0.704\\(0.011)} & \makecell{0.730\\(0.010)} & \makecell{0.790\\(0.011)} & \makecell{0.690\\(0.012)} & \makecell{\underline{0.839}\\(0.010)} & \makecell{0.814\\(0.010)} & \makecell{0.739\\(0.011)} & \makecell{0.669\\(0.015)} & \makecell{\textbf{0.866}\\(0.010)} \\ \cmidrule(lr){2-12}
				& \textit{\textbf{M}\textsubscript{graphlets}} & \makecell{0.524\\(0.013)} & \makecell{0.704\\(0.011)} & \makecell{0.730\\(0.010)} & \makecell{0.762\\(0.011)} & \makecell{0.690\\(0.013)} & \makecell{\underline{0.814}\\(0.010)} & \makecell{0.802\\(0.010)} & \makecell{0.775\\(0.011)} & \makecell{0.646\\(0.015)} & \makecell{\textbf{0.871}\\(0.009)} \\ \cmidrule(lr){2-12}
				& \textit{\textbf{M}\textsubscript{reg+graphlets}} & \makecell{0.524\\(0.013)} & \makecell{0.704\\(0.011)} & \makecell{0.730\\(0.010)} & \makecell{0.787\\(0.011)} & \makecell{0.679\\(0.012)} & \makecell{\underline{0.838}\\(0.010)} & \makecell{0.808\\(0.010)} & \makecell{0.778\\(0.010)} & \makecell{0.676\\(0.014)} & \makecell{\textbf{0.875}\\(0.010)} \\
				\bottomrule
			\end{tabular}
			
			\vspace{-0.5em}
		\end{center}
	\end{subtable}
	\begin{subtable}[h]{1.0\textwidth}
		\setlength{\abovecaptionskip}{-0.5em}
		\subcaption{Node classification}
		\label{appendix:tab:results:sparse:metafeats:nodeclass}
		\begin{center}
			\begin{tabular}{cccccccccccc}\toprule
				\textbf{Perf. Metric} & \textbf{Meta-Feature} &\makecell[c]{\textbf{\rs}} &\makecell[c]{\textbf{\gbperf}} &\makecell[c]{\textbf{\gbrank}} &\makecell[c]{\textbf{\isac}} &\makecell[c]{\textbf{\as}} &\makecell[c]{\textbf{\ss}} &\makecell[c]{\textbf{\alors}} &\makecell[c]{\textbf{\ncf}} &\makecell[c]{\textbf{\metaod}} &\makecell[c]{\textbf{\metagl}} \\\midrule
				\multirow{8}{*}{\textbf{AUC} ($ \uparrow $)} & \textit{\textbf{M}\textsubscript{compact}} & \makecell{0.518\\(0.026)} & \makecell{0.726\\(0.024)} & \makecell{\underline{0.739}\\(0.024)} & \makecell{0.694\\(0.025)} & \makecell{0.652\\(0.023)} & \makecell{0.727\\(0.023)} & \makecell{0.731\\(0.023)} & \makecell{0.731\\(0.024)} & \makecell{0.471\\(0.030)} & \makecell{\textbf{0.748}\\(0.024)} \\ \cmidrule(lr){2-12}
				& \textit{\textbf{M}\textsubscript{regular}} & \makecell{0.518\\(0.026)} & \makecell{0.726\\(0.024)} & \makecell{\textbf{0.739}\\(0.024)} & \makecell{0.687\\(0.022)} & \makecell{0.592\\(0.021)} & \makecell{\underline{0.739}\\(0.024)} & \makecell{0.730\\(0.024)} & \makecell{0.713\\(0.025)} & \makecell{0.485\\(0.031)} & \makecell{0.709\\(0.023)} \\ \cmidrule(lr){2-12}
				& \textit{\textbf{M}\textsubscript{graphlets}} & \makecell{0.518\\(0.026)} & \makecell{0.726\\(0.024)} & \makecell{\underline{0.739}\\(0.024)} & \makecell{0.687\\(0.024)} & \makecell{0.591\\(0.024)} & \makecell{0.682\\(0.025)} & \makecell{\textbf{0.751}\\(0.023)} & \makecell{0.721\\(0.023)} & \makecell{0.497\\(0.030)} & \makecell{0.731\\(0.025)} \\ \cmidrule(lr){2-12}
				& \textit{\textbf{M}\textsubscript{reg+graphlets}} & \makecell{0.518\\(0.026)} & \makecell{0.726\\(0.024)} & \makecell{\textbf{0.739}\\(0.024)} & \makecell{0.694\\(0.024)} & \makecell{0.606\\(0.025)} & \makecell{0.721\\(0.024)} & \makecell{\underline{0.731}\\(0.022)} & \makecell{0.705\\(0.024)} & \makecell{0.494\\(0.029)} & \makecell{0.711\\(0.023)} \\
				\bottomrule
			\end{tabular}
			
		\end{center}
	\end{subtable}
	
	\vspace{-0.5em}
\end{table}

\begin{table}[!htbp]\centering
\setlength{\abovecaptionskip}{-0.2em}
	\caption{Out-Of-Domain testbed results, obtained with different meta-graph features,
		for link prediction (top) and node classification (bottom) tasks.
		\textit{\textbf{M}\textsubscript{reg+graphlets}} denotes a concatenation of \textit{\textbf{M}\textsubscript{regular}} and \textit{\textbf{M}\textsubscript{graphlets}} meta-graph features.
		Higher~($ \uparrow $) scores are better. The numbers in the parentheses denote one standard error.
		The \textbf{best} result is in bold, and the \underline{second best} result is underlined.}
	\label{appendix:tab:results:out-of-domain:metafeats}
	\setlength{\tabcolsep}{3pt}
\fontsize{7.5}{8.5}\selectfont
	
	\begin{subtable}[h]{1.0\textwidth}
		\setlength{\abovecaptionskip}{-0.5em}
		\subcaption{Link prediction}
		\label{appendix:tab:results:out-of-domain:metafeats:linkpred}
		\begin{center}
			\begin{tabular}{cccccccccccc}\toprule
				\textbf{Perf. Metric} & \textbf{Meta-Feature} &\makecell[c]{\textbf{\rs}} &\makecell[c]{\textbf{\gbperf}} &\makecell[c]{\textbf{\gbrank}} &\makecell[c]{\textbf{\isac}} &\makecell[c]{\textbf{\as}} &\makecell[c]{\textbf{\ss}} &\makecell[c]{\textbf{\alors}} &\makecell[c]{\textbf{\ncf}} &\makecell[c]{\textbf{\metaod}} &\makecell[c]{\textbf{\metagl}} \\\midrule
				\multirow{8}{*}{\textbf{AUC} ($ \uparrow $)} & \textit{\textbf{M}\textsubscript{compact}} & \makecell{0.517\\(0.013)} & \makecell{0.809\\(0.010)} & \makecell{0.811\\(0.010)} & \makecell{0.805\\(0.012)} & \makecell{0.779\\(0.012)} & \makecell{0.813\\(0.010)} & \makecell{\underline{0.832}\\(0.009)} & \makecell{0.816\\(0.010)} & \makecell{0.659\\(0.015)} & \makecell{\textbf{0.867}\\(0.009)} \\ \cmidrule(lr){2-12}
				& \textit{\textbf{M}\textsubscript{regular}} & \makecell{0.517\\(0.013)} & \makecell{0.809\\(0.010)} & \makecell{0.811\\(0.010)} & \makecell{\underline{0.850}\\(0.009)} & \makecell{0.786\\(0.012)} & \makecell{0.837\\(0.010)} & \makecell{0.820\\(0.009)} & \makecell{0.837\\(0.009)} & \makecell{0.681\\(0.015)} & \makecell{\textbf{0.871}\\(0.009)} \\ \cmidrule(lr){2-12}
				& \textit{\textbf{M}\textsubscript{graphlets}} & \makecell{0.517\\(0.013)} & \makecell{0.809\\(0.010)} & \makecell{0.811\\(0.010)} & \makecell{\underline{0.828}\\(0.011)} & \makecell{0.788\\(0.011)} & \makecell{0.777\\(0.012)} & \makecell{0.807\\(0.010)} & \makecell{0.828\\(0.010)} & \makecell{0.689\\(0.015)} & \makecell{\textbf{0.864}\\(0.009)} \\ \cmidrule(lr){2-12}
				& \textit{\textbf{M}\textsubscript{reg+graphlets}} & \makecell{0.517\\(0.013)} & \makecell{0.809\\(0.010)} & \makecell{0.811\\(0.010)} & \makecell{\underline{0.843}\\(0.011)} & \makecell{0.790\\(0.011)} & \makecell{0.839\\(0.009)} & \makecell{0.830\\(0.009)} & \makecell{0.832\\(0.010)} & \makecell{0.688\\(0.015)} & \makecell{\textbf{0.864}\\(0.009)} \\
				\bottomrule
			\end{tabular}
			
			\vspace{-0.5em}
		\end{center}
	\end{subtable}
	\begin{subtable}[h]{1.0\textwidth}
		\setlength{\abovecaptionskip}{-0.5em}
		\subcaption{Node classification}
		\label{appendix:tab:results:out-of-domain:metafeats:nodeclass}
		\begin{center}
			\begin{tabular}{cccccccccccc}\toprule
				\textbf{Perf. Metric} & \textbf{Meta-Feature} &\makecell[c]{\textbf{\rs}} &\makecell[c]{\textbf{\gbperf}} &\makecell[c]{\textbf{\gbrank}} &\makecell[c]{\textbf{\isac}} &\makecell[c]{\textbf{\as}} &\makecell[c]{\textbf{\ss}} &\makecell[c]{\textbf{\alors}} &\makecell[c]{\textbf{\ncf}} &\makecell[c]{\textbf{\metaod}} &\makecell[c]{\textbf{\metagl}} \\\midrule
				\multirow{8}{*}{\textbf{AUC} ($ \uparrow $)} & \textit{\textbf{M}\textsubscript{compact}} & \makecell{0.495\\(0.025)} & \makecell{0.726\\(0.025)} & \makecell{0.727\\(0.025)} & \makecell{0.718\\(0.025)} & \makecell{0.646\\(0.025)} & \makecell{\underline{0.728}\\(0.022)} & \makecell{0.690\\(0.023)} & \makecell{\textbf{0.736}\\(0.023)} & \makecell{0.468\\(0.027)} & \makecell{0.715\\(0.024)} \\ \cmidrule(lr){2-12}
				& \textit{\textbf{M}\textsubscript{regular}} & \makecell{0.495\\(0.025)} & \makecell{0.726\\(0.025)} & \makecell{0.727\\(0.025)} & \makecell{0.701\\(0.025)} & \makecell{0.684\\(0.023)} & \makecell{\textbf{0.750}\\(0.023)} & \makecell{0.668\\(0.026)} & \makecell{\underline{0.741}\\(0.024)} & \makecell{0.571\\(0.027)} & \makecell{0.705\\(0.023)} \\ \cmidrule(lr){2-12}
				& \textit{\textbf{M}\textsubscript{graphlets}} & \makecell{0.495\\(0.025)} & \makecell{0.726\\(0.025)} & \makecell{\underline{0.727}\\(0.025)} & \makecell{0.697\\(0.025)} & \makecell{0.673\\(0.026)} & \makecell{\textbf{0.746}\\(0.022)} & \makecell{0.677\\(0.024)} & \makecell{0.717\\(0.024)} & \makecell{0.537\\(0.030)} & \makecell{0.688\\(0.024)} \\ \cmidrule(lr){2-12}
				& \textit{\textbf{M}\textsubscript{reg+graphlets}} & \makecell{0.495\\(0.025)} & \makecell{0.726\\(0.025)} & \makecell{\underline{0.727}\\(0.025)} & \makecell{0.691\\(0.024)} & \makecell{0.667\\(0.025)} & \makecell{\textbf{0.732}\\(0.023)} & \makecell{0.628\\(0.025)} & \makecell{0.712\\(0.026)} & \makecell{0.536\\(0.029)} & \makecell{0.660\\(0.025)} \\
				\bottomrule
			\end{tabular}
			
		\end{center}
	\end{subtable}
	
	\vspace{-0.5em}
\end{table}

\begin{table}[!htbp]\centering
\setlength{\abovecaptionskip}{-0.2em}
	\caption{Small-To-Large testbed results, obtained with different meta-graph features,
		for link prediction (top) and node classification (bottom) tasks.
		\textit{\textbf{M}\textsubscript{reg+graphlets}} denotes a concatenation of \textit{\textbf{M}\textsubscript{regular}} and \textit{\textbf{M}\textsubscript{graphlets}} meta-graph features.
		Higher~($ \uparrow $) scores are better. The numbers in the parentheses denote one standard error.
		The \textbf{best} result is in bold, and the \underline{second best} result is underlined.}
	\label{appendix:tab:results:small-to-large:metafeats}
	\setlength{\tabcolsep}{3pt}
\fontsize{7.5}{8.5}\selectfont
	
	\begin{subtable}[h]{1.0\textwidth}
		\setlength{\abovecaptionskip}{-0.5em}
		\subcaption{Link prediction}
		\label{appendix:tab:results:small-to-large:metafeats:linkpred}
		\begin{center}
			\begin{tabular}{cccccccccccc}\toprule
				\textbf{Perf. Metric} & \textbf{Meta-Feature} &\makecell[c]{\textbf{\rs}} &\makecell[c]{\textbf{\gbperf}} &\makecell[c]{\textbf{\gbrank}} &\makecell[c]{\textbf{\isac}} &\makecell[c]{\textbf{\as}} &\makecell[c]{\textbf{\ss}} &\makecell[c]{\textbf{\alors}} &\makecell[c]{\textbf{\ncf}} &\makecell[c]{\textbf{\metaod}} &\makecell[c]{\textbf{\metagl}} \\\midrule
				\multirow{8}{*}{\textbf{AUC} ($ \uparrow $)} & \textit{\textbf{M}\textsubscript{compact}} & \makecell{0.522\\(0.027)} & \makecell{0.798\\(0.017)} & \makecell{0.797\\(0.017)} & \makecell{0.772\\(0.022)} & \makecell{0.835\\(0.020)} & \makecell{0.715\\(0.021)} & \makecell{\underline{0.837}\\(0.018)} & \makecell{0.783\\(0.017)} & \makecell{0.480\\(0.029)} & \makecell{\textbf{0.875}\\(0.018)} \\ \cmidrule(lr){2-12}
				& \textit{\textbf{M}\textsubscript{regular}} & \makecell{0.522\\(0.027)} & \makecell{0.798\\(0.017)} & \makecell{0.797\\(0.017)} & \makecell{\underline{0.842}\\(0.018)} & \makecell{0.827\\(0.022)} & \makecell{0.767\\(0.020)} & \makecell{0.812\\(0.024)} & \makecell{0.796\\(0.017)} & \makecell{0.667\\(0.029)} & \makecell{\textbf{0.870}\\(0.018)} \\ \cmidrule(lr){2-12}
				& \textit{\textbf{M}\textsubscript{graphlets}} & \makecell{0.522\\(0.027)} & \makecell{0.798\\(0.017)} & \makecell{0.797\\(0.017)} & \makecell{0.841\\(0.016)} & \makecell{0.830\\(0.019)} & \makecell{0.750\\(0.021)} & \makecell{\underline{0.842}\\(0.020)} & \makecell{0.783\\(0.018)} & \makecell{0.700\\(0.028)} & \makecell{\textbf{0.875}\\(0.016)} \\ \cmidrule(lr){2-12}
				& \textit{\textbf{M}\textsubscript{reg+graphlets}} & \makecell{0.522\\(0.027)} & \makecell{0.798\\(0.017)} & \makecell{0.797\\(0.017)} & \makecell{\underline{0.843}\\(0.016)} & \makecell{0.841\\(0.020)} & \makecell{0.806\\(0.018)} & \makecell{0.831\\(0.021)} & \makecell{0.795\\(0.018)} & \makecell{0.750\\(0.027)} & \makecell{\textbf{0.858}\\(0.018)} \\
				\bottomrule
			\end{tabular}
			
			\vspace{-0.5em}
		\end{center}
	\end{subtable}
	\begin{subtable}[h]{1.0\textwidth}
		\setlength{\abovecaptionskip}{-0.5em}
		\subcaption{Node classification}
		\label{appendix:tab:results:small-to-large:metafeats:nodeclass}
		\begin{center}
			\begin{tabular}{cccccccccccc}\toprule
				\textbf{Perf. Metric} & \textbf{Meta-Feature} &\makecell[c]{\textbf{\rs}} &\makecell[c]{\textbf{\gbperf}} &\makecell[c]{\textbf{\gbrank}} &\makecell[c]{\textbf{\isac}} &\makecell[c]{\textbf{\as}} &\makecell[c]{\textbf{\ss}} &\makecell[c]{\textbf{\alors}} &\makecell[c]{\textbf{\ncf}} &\makecell[c]{\textbf{\metaod}} &\makecell[c]{\textbf{\metagl}} \\\midrule
				\multirow{8}{*}{\textbf{AUC} ($ \uparrow $)} & \textit{\textbf{M}\textsubscript{compact}} & \makecell{0.508\\(0.039)} & \makecell{\underline{0.724}\\(0.042)} & \makecell{\textbf{0.726}\\(0.042)} & \makecell{0.655\\(0.043)} & \makecell{0.723\\(0.038)} & \makecell{0.682\\(0.044)} & \makecell{0.696\\(0.041)} & \makecell{0.693\\(0.039)} & \makecell{0.586\\(0.039)} & \makecell{0.714\\(0.041)} \\ \cmidrule(lr){2-12}
				& \textit{\textbf{M}\textsubscript{regular}} & \makecell{0.508\\(0.039)} & \makecell{0.724\\(0.042)} & \makecell{0.726\\(0.042)} & \makecell{0.711\\(0.044)} & \makecell{\textbf{0.761}\\(0.037)} & \makecell{0.664\\(0.041)} & \makecell{0.701\\(0.035)} & \makecell{0.697\\(0.042)} & \makecell{0.467\\(0.043)} & \makecell{\underline{0.736}\\(0.037)} \\ \cmidrule(lr){2-12}
				& \textit{\textbf{M}\textsubscript{graphlets}} & \makecell{0.508\\(0.039)} & \makecell{0.724\\(0.042)} & \makecell{0.726\\(0.042)} & \makecell{\textbf{0.743}\\(0.039)} & \makecell{0.725\\(0.037)} & \makecell{0.712\\(0.039)} & \makecell{0.658\\(0.042)} & \makecell{\underline{0.728}\\(0.039)} & \makecell{0.494\\(0.044)} & \makecell{0.722\\(0.040)} \\ \cmidrule(lr){2-12}
				& \textit{\textbf{M}\textsubscript{reg+graphlets}} & \makecell{0.508\\(0.039)} & \makecell{0.724\\(0.042)} & \makecell{\underline{0.726}\\(0.042)} & \makecell{\textbf{0.744}\\(0.039)} & \makecell{0.677\\(0.044)} & \makecell{0.613\\(0.042)} & \makecell{0.700\\(0.038)} & \makecell{0.707\\(0.041)} & \makecell{0.493\\(0.043)} & \makecell{0.726\\(0.038)} \\
				\bottomrule
			\end{tabular}
			
		\end{center}
	\end{subtable}
	
	\vspace{-0.5em}
\end{table}

\begin{table}[!htbp]\centering
\setlength{\abovecaptionskip}{-0.2em}
	\caption{Cross-Task testbed results, obtained with different meta-graph features, 
		for node classification-to-link prediction (top) and link prediction-to-node classification (bottom) settings.
\textit{\textbf{M}\textsubscript{reg+graphlets}} denotes a concatenation of \textit{\textbf{M}\textsubscript{regular}} and \textit{\textbf{M}\textsubscript{graphlets}} meta-graph features.
		Higher~($ \uparrow $) scores are better. The numbers in the parentheses denote one standard error.
		The \textbf{best} result is in bold, and the \underline{second best} result is underlined.}
	\label{appendix:tab:results:cross-task:metafeats}
	\setlength{\tabcolsep}{3pt}
\fontsize{7.5}{8.5}\selectfont
	
	\begin{subtable}[h]{1.0\textwidth}
		\setlength{\abovecaptionskip}{-0.5em}
		\subcaption{Node classification $ \rightarrow $ Link prediction}
		\label{appendix:tab:results:cross-task:node_to_link:metafeats}
		\begin{center}
			\begin{tabular}{cccccccccccc}\toprule
				\textbf{Perf. Metric} & \textbf{Meta-Feature} &\makecell[c]{\textbf{\rs}} &\makecell[c]{\textbf{\gbperf}} &\makecell[c]{\textbf{\gbrank}} &\makecell[c]{\textbf{\isac}} &\makecell[c]{\textbf{\as}} &\makecell[c]{\textbf{\ss}} &\makecell[c]{\textbf{\alors}} &\makecell[c]{\textbf{\ncf}} &\makecell[c]{\textbf{\metaod}} &\makecell[c]{\textbf{\metagl}} \\\midrule
				\multirow{8}{*}{\textbf{AUC} ($ \uparrow $)} & \textit{\textbf{M}\textsubscript{compact}} & \makecell{0.479\\(0.016)} & \makecell{0.652\\(0.013)} & \makecell{\textbf{0.671}\\(0.013)} & \makecell{0.646\\(0.013)} & \makecell{0.536\\(0.016)} & \makecell{0.626\\(0.014)} & \makecell{0.553\\(0.014)} & \makecell{\underline{0.670}\\(0.013)} & \makecell{0.376\\(0.016)} & \makecell{0.601\\(0.014)} \\ \cmidrule(lr){2-12}
				& \textit{\textbf{M}\textsubscript{regular}} & \makecell{0.479\\(0.016)} & \makecell{\underline{0.652}\\(0.013)} & \makecell{\textbf{0.671}\\(0.013)} & \makecell{0.626\\(0.014)} & \makecell{0.520\\(0.015)} & \makecell{0.636\\(0.014)} & \makecell{0.594\\(0.015)} & \makecell{0.650\\(0.013)} & \makecell{0.465\\(0.015)} & \makecell{0.553\\(0.014)} \\ \cmidrule(lr){2-12}
				& \textit{\textbf{M}\textsubscript{graphlets}} & \makecell{0.479\\(0.016)} & \makecell{0.652\\(0.013)} & \makecell{\textbf{0.671}\\(0.013)} & \makecell{0.614\\(0.014)} & \makecell{0.510\\(0.016)} & \makecell{0.624\\(0.014)} & \makecell{0.526\\(0.015)} & \makecell{\underline{0.660}\\(0.014)} & \makecell{0.409\\(0.015)} & \makecell{0.564\\(0.014)} \\ \cmidrule(lr){2-12}
				& \textit{\textbf{M}\textsubscript{reg+graphlets}} & \makecell{0.479\\(0.016)} & \makecell{\underline{0.652}\\(0.013)} & \makecell{\textbf{0.671}\\(0.013)} & \makecell{0.608\\(0.015)} & \makecell{0.525\\(0.015)} & \makecell{0.601\\(0.014)} & \makecell{0.582\\(0.015)} & \makecell{0.640\\(0.013)} & \makecell{0.414\\(0.015)} & \makecell{0.546\\(0.015)} \\
				\bottomrule
			\end{tabular}
			
			\vspace{-0.5em}
		\end{center}
	\end{subtable}
	
	\begin{subtable}[h]{1.0\textwidth}
		\setlength{\abovecaptionskip}{-0.5em}
		\subcaption{Link prediction $ \rightarrow $ Node classification}
		\label{appendix:tab:results:cross-task:link_to_node:metafeats}
		\begin{center}
			\begin{tabular}{cccccccccccc}\toprule
				\textbf{Perf. Metric} & \textbf{Meta-Feature} &\makecell[c]{\textbf{\rs}} &\makecell[c]{\textbf{\gbperf}} &\makecell[c]{\textbf{\gbrank}} &\makecell[c]{\textbf{\isac}} &\makecell[c]{\textbf{\as}} &\makecell[c]{\textbf{\ss}} &\makecell[c]{\textbf{\alors}} &\makecell[c]{\textbf{\ncf}} &\makecell[c]{\textbf{\metaod}} &\makecell[c]{\textbf{\metagl}} \\\midrule
				\multirow{8}{*}{\textbf{AUC} ($ \uparrow $)} & \textit{\textbf{M}\textsubscript{compact}} & \makecell{0.542\\(0.025)} & \makecell{0.595\\(0.025)} & \makecell{\underline{0.596}\\(0.025)} & \makecell{0.557\\(0.027)} & \makecell{0.445\\(0.027)} & \makecell{0.490\\(0.025)} & \makecell{0.543\\(0.025)} & \makecell{\textbf{0.617}\\(0.025)} & \makecell{0.407\\(0.024)} & \makecell{0.554\\(0.024)} \\ \cmidrule(lr){2-12}
				& \textit{\textbf{M}\textsubscript{regular}} & \makecell{0.542\\(0.025)} & \makecell{0.595\\(0.025)} & \makecell{0.596\\(0.025)} & \makecell{0.581\\(0.026)} & \makecell{0.517\\(0.025)} & \makecell{0.536\\(0.024)} & \makecell{\underline{0.606}\\(0.025)} & \makecell{\textbf{0.608}\\(0.025)} & \makecell{0.477\\(0.024)} & \makecell{0.542\\(0.026)} \\ \cmidrule(lr){2-12}
				& \textit{\textbf{M}\textsubscript{graphlets}} & \makecell{0.542\\(0.025)} & \makecell{\underline{0.595}\\(0.025)} & \makecell{\textbf{0.596}\\(0.025)} & \makecell{0.517\\(0.029)} & \makecell{0.477\\(0.026)} & \makecell{0.573\\(0.025)} & \makecell{0.505\\(0.024)} & \makecell{0.593\\(0.025)} & \makecell{0.460\\(0.025)} & \makecell{0.545\\(0.025)} \\ \cmidrule(lr){2-12}
				& \textit{\textbf{M}\textsubscript{reg+graphlets}} & \makecell{0.542\\(0.025)} & \makecell{0.595\\(0.025)} & \makecell{\underline{0.596}\\(0.025)} & \makecell{0.534\\(0.028)} & \makecell{0.494\\(0.026)} & \makecell{0.594\\(0.025)} & \makecell{0.518\\(0.025)} & \makecell{\textbf{0.607}\\(0.026)} & \makecell{0.448\\(0.026)} & \makecell{0.528\\(0.027)} \\
				\bottomrule
			\end{tabular}
			
		\end{center}
	\end{subtable}
	
	\vspace{-0.5em}
\end{table}

\clearpage
\subsection{Results on Time Cost}

\begin{figure*}[!t]
\centering
	\makebox[0.4\textwidth][c]{
		\includegraphics[width=0.75\linewidth]{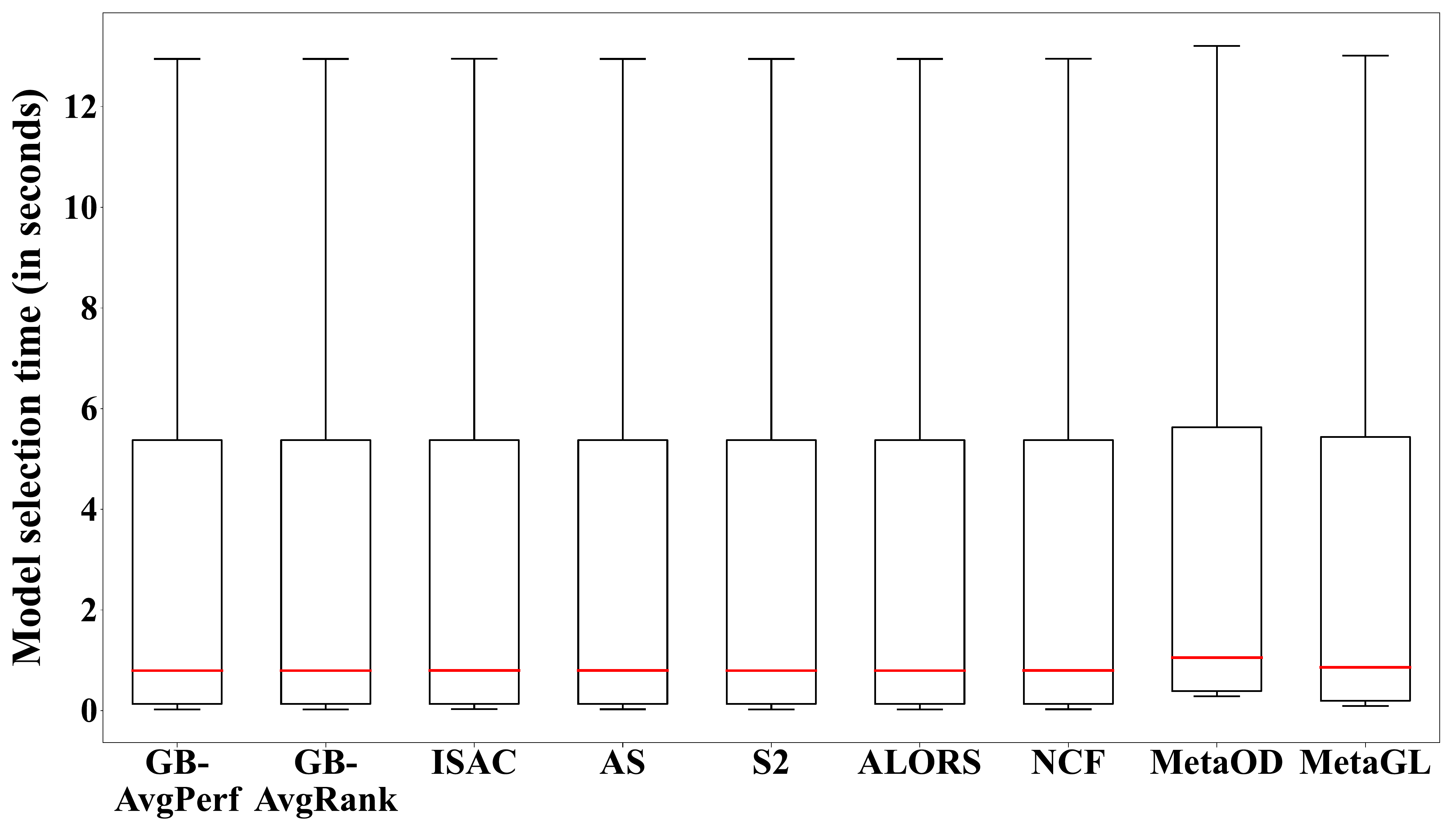}
	}
	\setlength{\abovecaptionskip}{0.5em}
\caption{
		Time taken for selecting the best model (in seconds), 
		which includes the time for generating meta-graph features for the new graph, and 
		the time for model selection algorithms to infer the best model for the given graph based on them.
}
	\label{fig:exp:inferencetime}
\end{figure*}

\textbf{Model Selection Runtime.} 
In \Cref{fig:exp:inferencetime}, we report the distribution of runtime (in seconds) for model selection, measured over the graphs in the benchmark.
The runtime includes the time for generating meta-graph features for the new graph, and 
the time for a model selection method to infer the best model based on them.
The median runtime (shown by the red line) is less than one second, and for a majority of the graphs, it takes at most five to six seconds.
Note that the distributions are mostly the same for different methods, 
as the time for different model selection algorithms to infer the best model is very short (close to zero second), and similar to each other.
We measured the time for meta feature generation via sequential processing for simplicity,
while these features can be processed in parallel as they are independent of each other.

\textbf{Training Runtime.} 
\Cref{appendix:tab:results:fully-observed:traintime} reports the time (in seconds) taken for training model selection algorithms until convergence
on the Fully-Observed testbed for the link prediction task.
Note that Global Best algorithms and \as are excluded as they do not require model training.
\isac~takes the least amount of time as it only performs clustering without model parameter updates during the training phase.
Algorithms that rely on neural networks require more training time. 
Yet a majority of them can still be trained quickly, in just a few seconds to a few minutes.
\metaod takes the largest amount of time with its current optimization framework.

\begin{table}[!h]\centering
\setlength{\abovecaptionskip}{0.5em}
	\caption{
		Time (in seconds) taken for training model selection algorithms on the Fully-Observed testbed for the link prediction task.
		For each algorithm, we show the training runtime averaged over the splits in the testbed, and the standard deviation in the parentheses.
	}
	\label{appendix:tab:results:fully-observed:traintime}
\fontsize{8.5}{9.5}\selectfont

	\begin{center}
		\begin{tabular}{cccccc}\toprule
			\makecell[c]{\textbf{\isac}} &\makecell[c]{\textbf{\ss}} &\makecell[c]{\textbf{\alors}} &\makecell[c]{\textbf{\ncf}} &\makecell[c]{\textbf{\metaod}} &\makecell[c]{\textbf{\metagl}} \\\midrule
			\makecell{0.0330\\(0.0279)} & 
			\makecell{0.6162\\(0.6473)} & 
			\makecell{0.7530\\(0.0897)} & 
			\makecell{9.2510\\(2.3227)} & 
			\makecell{8136.1713\\(89.6785)} & 
			\makecell{172.0742\\(50.9201)} \\
			\bottomrule
		\end{tabular}
	\end{center}
\end{table}

\subsection{Testbed Results With Standard Error}\label{app:results:error}

In the main text, we provide the average model selection performances in four testbeds,
\ie, Fully-Observed, Sparse, Out-Of-Domain, and Small-To-Large testbeds,
but their standard errors could not be shown due to space constraint.
In this subsection, we present the standard error along with the average performance in those four testbeds 
(\Cref{appendix:tab:results:fully-observed,appendix:tab:results:sparse,appendix:tab:results:out-of-domain,appendix:tab:results:small-to-large}).
Please refer to the main text for the discussion of the results of these testbeds.
Cross-Task testbed results are discussed in \Cref{app:results:crosstask}

\begin{table}[!htbp]\centering
\setlength{\abovecaptionskip}{-0.2em}
	\caption{Fully-Observed testbed results for link prediction (top) and node classification (bottom)~tasks.
	Higher~($ \uparrow $) scores are better. The numbers in the parentheses denote one standard error.
	The \textbf{best} result is in bold, and the \underline{second best} result is underlined.}
	\label{appendix:tab:results:fully-observed}
	\setlength{\tabcolsep}{3pt}
\fontsize{7.5}{8.5}\selectfont
	
	\begin{subtable}[h]{1.0\textwidth}
		\setlength{\abovecaptionskip}{-0.5em}
		\subcaption{Link prediction}
		\label{appendix:tab:results:fully-observed:linkpred}
		\begin{center}
			\begin{tabular}{lcccccccccc}\toprule
				\textbf{Perf. Metric}&\makecell[c]{\textbf{\rs}} &\makecell[c]{\textbf{\gbperf}} &\makecell[c]{\textbf{\gbrank}} &\makecell[c]{\textbf{\isac}} &\makecell[c]{\textbf{\as}} &\makecell[c]{\textbf{\ss}} &\makecell[c]{\textbf{\alors}} &\makecell[c]{\textbf{\ncf}} &\makecell[c]{\textbf{\metaod}} &\makecell[c]{\textbf{\metagl}} \\\midrule
				\textbf{AUC} ($ \uparrow $)	& \makecell{0.524\\(0.013)} & \makecell{0.735\\(0.011)} & \makecell{0.730\\(0.010)} & \makecell{0.807\\(0.011)} & \makecell{\underline{0.864}\\(0.010)} & \makecell{0.809\\(0.011)} & \makecell{0.843\\(0.010)} & \makecell{0.728\\(0.011)} & \makecell{0.764\\(0.014)} & \makecell{\textbf{0.875}\\(0.009)} \\ \midrule[0.2pt]
				\textbf{MRR} ($ \uparrow $) & \makecell{0.016\\(0.001)} & \makecell{0.087\\(0.010)} & \makecell{0.064\\(0.007)} & \makecell{0.134\\(0.011)} & \makecell{\textbf{0.371}\\(0.019)} & \makecell{0.198\\(0.015)} & \makecell{0.201\\(0.015)} & \makecell{0.073\\(0.008)} & \makecell{0.096\\(0.008)} & \makecell{\underline{0.295}\\(0.017)} \\ \midrule[0.2pt]
				\textbf{NDCG@1} ($ \uparrow $) & \makecell{0.813\\(0.006)} & \makecell{0.942\\(0.003)} & \makecell{0.934\\(0.004)} & \makecell{0.944\\(0.004)} & \makecell{0.957\\(0.004)} & \makecell{0.950\\(0.004)} & \makecell{\underline{0.961}\\(0.003)} & \makecell{0.943\\(0.003)} & \makecell{0.937\\(0.003)} & \makecell{\textbf{0.969}\\(0.003)} \\
				\bottomrule
			\end{tabular}
			
			\vspace{-0.5em}
		\end{center}
	\end{subtable}
	\begin{subtable}[h]{1.0\textwidth}
		\setlength{\abovecaptionskip}{-0.5em}
		\subcaption{Node classification}
		\label{appendix:tab:results:fully-observed:nodeclass}
		\begin{center}
			\begin{tabular}{lcccccccccc}\toprule
				\textbf{Perf. Metric}&\makecell[c]{\textbf{\rs}} &\makecell[c]{\textbf{\gbperf}} &\makecell[c]{\textbf{\gbrank}} &\makecell[c]{\textbf{\isac}} &\makecell[c]{\textbf{\as}} &\makecell[c]{\textbf{\ss}} &\makecell[c]{\textbf{\alors}} &\makecell[c]{\textbf{\ncf}} &\makecell[c]{\textbf{\metaod}} &\makecell[c]{\textbf{\metagl}} \\\midrule
				\textbf{AUC} ($ \uparrow $)	& \makecell{0.518\\(0.026)} & \makecell{0.747\\(0.024)} & \makecell{0.744\\(0.024)} & \makecell{0.746\\(0.023)} & \makecell{\underline{0.762}\\(0.023)} & \makecell{\textbf{0.772}\\(0.022)} & \makecell{0.734\\(0.023)} & \makecell{0.745\\(0.025)} & \makecell{0.581\\(0.028)} & \makecell{0.740\\(0.024)} \\ \midrule[0.2pt]
				\textbf{MRR} ($ \uparrow $) & \makecell{0.029\\(0.008)} & \makecell{0.102\\(0.018)} & \makecell{0.124\\(0.022)} & \makecell{0.118\\(0.021)} & \makecell{\textbf{0.181}\\(0.029)} & \makecell{0.110\\(0.019)} & \makecell{0.103\\(0.019)} & \makecell{0.124\\(0.023)} & \makecell{0.041\\(0.010)} & \makecell{\underline{0.129}\\(0.023)} \\ \midrule[0.2pt]
				\textbf{NDCG@1} ($ \uparrow $) & \makecell{0.747\\(0.021)} & \makecell{0.865\\(0.015)} & \makecell{0.860\\(0.015)} & \makecell{0.885\\(0.013)} & \makecell{\underline{0.892}\\(0.013)} & \makecell{\textbf{0.916}\\(0.010)} & \makecell{0.886\\(0.014)} & \makecell{0.883\\(0.013)} & \makecell{0.839\\(0.017)} & \makecell{0.863\\(0.019)} \\
				\bottomrule
			\end{tabular}
			
		\end{center}
	\end{subtable}

\end{table}

\begin{table}[!htbp]\centering
\setlength{\abovecaptionskip}{-0.3em}
	\caption{Sparse testbed results for link prediction (top) and node classification (bottom)~tasks.
		Higher~($ \uparrow $) scores are better. The numbers in the parentheses denote one standard error.
		The \textbf{best} result is in bold, and the \underline{second best} result is underlined.}
	\label{appendix:tab:results:sparse}
	\setlength{\tabcolsep}{3pt}
\fontsize{7.5}{8.5}\selectfont
	
	\begin{subtable}[h]{1.0\textwidth}
		\setlength{\abovecaptionskip}{-0.5em}
		\subcaption{Link prediction}
		\label{appendix:tab:results:sparse:linkpred}
		\begin{center}
			\begin{tabular}{cccccccccccc}\toprule
				\textbf{Perf. Metric} & \textbf{Sparsity} &\makecell[c]{\textbf{\rs}} &\makecell[c]{\textbf{\gbperf}} &\makecell[c]{\textbf{\gbrank}} &\makecell[c]{\textbf{\isac}} &\makecell[c]{\textbf{\as}} &\makecell[c]{\textbf{\ss}} &\makecell[c]{\textbf{\alors}} &\makecell[c]{\textbf{\ncf}} &\makecell[c]{\textbf{\metaod}} &\makecell[c]{\textbf{\metagl}} \\\midrule
				\multirow{10}{*}{\textbf{AUC} ($ \uparrow $)} & 10\%	& \makecell{0.524\\(0.013)} & \makecell{0.733\\(0.011)} & \makecell{0.732\\(0.010)} & \makecell{0.804\\(0.011)} & \makecell{0.829\\(0.011)} & \makecell{0.813\\(0.011)} & \makecell{\underline{0.831}\\(0.010)} & \makecell{0.735\\(0.011)} & \makecell{0.743\\(0.014)} & \makecell{\textbf{0.865}\\(0.010)} \\ \cmidrule(lr){2-12}
				& 30\% & \makecell{0.524\\(0.013)} & \makecell{0.728\\(0.011)} & \makecell{0.738\\(0.010)} & \makecell{0.798\\(0.011)} & \makecell{0.763\\(0.012)} & \makecell{0.811\\(0.011)} & \makecell{\underline{0.827}\\(0.010)} & \makecell{0.739\\(0.011)} & \makecell{0.703\\(0.014)} & \makecell{\textbf{0.871}\\(0.010)} \\ \cmidrule(lr){2-12}
				& 50\% & \makecell{0.524\\(0.013)} & \makecell{0.704\\(0.011)} & \makecell{0.730\\(0.010)} & \makecell{0.790\\(0.011)} & \makecell{0.690\\(0.012)} & \makecell{\underline{0.839}\\(0.010)} & \makecell{0.814\\(0.010)} & \makecell{0.739\\(0.011)} & \makecell{0.669\\(0.015)} & \makecell{\textbf{0.866}\\(0.010)} \\ \cmidrule(lr){2-12}
				& 70\% & \makecell{0.524\\(0.013)} & \makecell{0.708\\(0.010)} & \makecell{0.730\\(0.010)} & \makecell{0.778\\(0.011)} & \makecell{0.618\\(0.011)} & \makecell{\underline{0.814}\\(0.011)} & \makecell{0.795\\(0.010)} & \makecell{0.757\\(0.010)} & \makecell{0.630\\(0.015)} & \makecell{\textbf{0.866}\\(0.010)} \\ \cmidrule(lr){2-12}
				& 90\% & \makecell{0.524\\(0.013)} & \makecell{0.717\\(0.011)} & \makecell{\underline{0.732}\\(0.010)} & \makecell{0.720\\(0.012)} & \makecell{0.547\\(0.007)} & \makecell{0.464\\(0.012)} & \makecell{0.687\\(0.012)} & \makecell{0.656\\(0.012)} & \makecell{0.599\\(0.015)} & \makecell{\textbf{0.811}\\(0.011)} \\
				\bottomrule
			\end{tabular}
			
			\vspace{-0.5em}
		\end{center}
	\end{subtable}
	\begin{subtable}[h]{1.0\textwidth}
		\setlength{\abovecaptionskip}{-0.5em}
		\subcaption{Node classification}
		\label{appendix:tab:results:sparse:nodeclass}
		\begin{center}
			\begin{tabular}{cccccccccccc}\toprule
				\textbf{Perf. Metric} & \textbf{Sparsity} &\makecell[c]{\textbf{\rs}} &\makecell[c]{\textbf{\gbperf}} &\makecell[c]{\textbf{\gbrank}} &\makecell[c]{\textbf{\isac}} &\makecell[c]{\textbf{\as}} &\makecell[c]{\textbf{\ss}} &\makecell[c]{\textbf{\alors}} &\makecell[c]{\textbf{\ncf}} &\makecell[c]{\textbf{\metaod}} &\makecell[c]{\textbf{\metagl}} \\\midrule
				\multirow{10}{*}{\textbf{AUC} ($ \uparrow $)} & 10\%	& \makecell{0.518\\(0.026)} & \makecell{0.746\\(0.024)} & \makecell{0.744\\(0.024)} & \makecell{0.744\\(0.023)} & \makecell{0.748\\(0.023)} & \makecell{\textbf{0.766}\\(0.023)} & \makecell{0.727\\(0.026)} & \makecell{0.727\\(0.024)} & \makecell{0.575\\(0.029)} & \makecell{\underline{0.761}\\(0.021)} \\ \cmidrule(lr){2-12}
				& 30\% & \makecell{0.518\\(0.026)} & \makecell{\underline{0.743}\\(0.023)} & \makecell{0.738\\(0.024)} & \makecell{0.734\\(0.023)} & \makecell{0.680\\(0.023)} & \makecell{\textbf{0.769}\\(0.022)} & \makecell{0.741\\(0.024)} & \makecell{0.735\\(0.024)} & \makecell{0.533\\(0.031)} & \makecell{0.736\\(0.023)} \\ \cmidrule(lr){2-12}
				& 50\% & \makecell{0.518\\(0.026)} & \makecell{0.726\\(0.024)} & \makecell{\textbf{0.739}\\(0.024)} & \makecell{0.687\\(0.022)} & \makecell{0.592\\(0.021)} & \makecell{\underline{0.739}\\(0.024)} & \makecell{0.730\\(0.024)} & \makecell{0.713\\(0.025)} & \makecell{0.485\\(0.031)} & \makecell{0.709\\(0.023)} \\ \cmidrule(lr){2-12}
				& 70\% & \makecell{0.518\\(0.026)} & \makecell{0.692\\(0.024)} & \makecell{\textbf{0.738}\\(0.023)} & \makecell{0.653\\(0.023)} & \makecell{0.571\\(0.018)} & \makecell{0.684\\(0.023)} & \makecell{0.694\\(0.024)} & \makecell{\underline{0.709}\\(0.023)} & \makecell{0.483\\(0.029)} & \makecell{0.662\\(0.026)} \\ \cmidrule(lr){2-12}
				& 90\% & \makecell{0.518\\(0.026)} & \makecell{0.626\\(0.026)} & \makecell{\textbf{0.697}\\(0.025)} & \makecell{0.592\\(0.026)} & \makecell{0.535\\(0.012)} & \makecell{0.620\\(0.027)} & \makecell{0.654\\(0.026)} & \makecell{\underline{0.660}\\(0.026)} & \makecell{0.490\\(0.030)} & \makecell{0.659\\(0.024)} \\
				\bottomrule
			\end{tabular}
			
		\end{center}
	\end{subtable}

\end{table}

\begin{table}[!t]\centering
\setlength{\abovecaptionskip}{-0.5em}
	\caption{Out-Of-Domain testbed results for link prediction (top) and node classification (bottom)~tasks.
		Higher~($ \uparrow $) scores are better. The numbers in the parentheses denote one standard error.
		The \textbf{best} result is in bold, and the \underline{second best} result is underlined.}
	\label{appendix:tab:results:out-of-domain}
	\setlength{\tabcolsep}{3pt}
\fontsize{7.5}{8.5}\selectfont
	
	\begin{subtable}[h]{1.0\textwidth}
		\setlength{\abovecaptionskip}{-0.5em}
		\subcaption{Link prediction}
		\label{appendix:tab:results:out-of-domain:linkpred}
		\begin{center}
			\begin{tabular}{lcccccccccc}\toprule
				\textbf{Perf. Metric}&\makecell[c]{\textbf{\rs}} &\makecell[c]{\textbf{\gbperf}} &\makecell[c]{\textbf{\gbrank}} &\makecell[c]{\textbf{\isac}} &\makecell[c]{\textbf{\as}} &\makecell[c]{\textbf{\ss}} &\makecell[c]{\textbf{\alors}} &\makecell[c]{\textbf{\ncf}} &\makecell[c]{\textbf{\metaod}} &\makecell[c]{\textbf{\metagl}} \\\midrule
				\textbf{AUC} ($ \uparrow $)	& \makecell{0.517\\(0.013)} & \makecell{0.809\\(0.010)} & \makecell{0.811\\(0.010)} & \makecell{\underline{0.850}\\(0.009)} & \makecell{0.786\\(0.012)} & \makecell{0.837\\(0.010)} & \makecell{0.820\\(0.009)} & \makecell{0.837\\(0.009)} & \makecell{0.681\\(0.015)} & \makecell{\textbf{0.871}\\(0.009)} \\ \midrule[0.2pt]
				\textbf{MRR} ($ \uparrow $) & \makecell{0.018\\(0.002)} & \makecell{0.110\\(0.010)} & \makecell{0.101\\(0.008)} & \makecell{0.125\\(0.010)} & \makecell{\textbf{0.237}\\(0.017)} & \makecell{0.116\\(0.010)} & \makecell{0.109\\(0.010)} & \makecell{0.109\\(0.010)} & \makecell{0.047\\(0.005)} & \makecell{\underline{0.148}\\(0.011)} \\ \midrule[0.2pt]
				\textbf{NDCG@1} ($ \uparrow $) & \makecell{0.820\\(0.007)} & \makecell{\textbf{0.956}\\(0.003)} & \makecell{\underline{0.954}\\(0.003)} & \makecell{0.951\\(0.003)} & \makecell{0.935\\(0.004)} & \makecell{0.953\\(0.003)} & \makecell{0.953\\(0.003)} & \makecell{0.952\\(0.003)} & \makecell{0.918\\(0.004)} & \makecell{0.951\\(0.004)} \\
				\bottomrule
			\end{tabular}

			\vspace{-0.5em}
		\end{center}
	\end{subtable}
	\begin{subtable}[h]{1.0\textwidth}
		\setlength{\abovecaptionskip}{-0.5em}
		\subcaption{Node classification}
		\label{appendix:tab:results:out-of-domain:nodeclass}
		\begin{center}
			\begin{tabular}{lcccccccccc}\toprule
				\textbf{Perf. Metric}&\makecell[c]{\textbf{\rs}} &\makecell[c]{\textbf{\gbperf}} &\makecell[c]{\textbf{\gbrank}} &\makecell[c]{\textbf{\isac}} &\makecell[c]{\textbf{\as}} &\makecell[c]{\textbf{\ss}} &\makecell[c]{\textbf{\alors}} &\makecell[c]{\textbf{\ncf}} &\makecell[c]{\textbf{\metaod}} &\makecell[c]{\textbf{\metagl}} \\\midrule
				\textbf{AUC} ($ \uparrow $)	& \makecell{0.495\\(0.025)} & \makecell{0.726\\(0.025)} & \makecell{0.727\\(0.025)} & \makecell{0.701\\(0.025)} & \makecell{0.684\\(0.023)} & \makecell{\textbf{0.750}\\(0.023)} & \makecell{0.668\\(0.026)} & \makecell{\underline{0.741}\\(0.024)} & \makecell{0.571\\(0.027)} & \makecell{0.705\\(0.023)} \\ \midrule[0.2pt]
				\textbf{MRR} ($ \uparrow $) & \makecell{0.019\\(0.005)} & \makecell{0.074\\(0.015)} & \makecell{\underline{0.086}\\(0.019)} & \makecell{0.046\\(0.009)} & \makecell{0.060\\(0.016)} & \makecell{\textbf{0.089}\\(0.017)} & \makecell{0.056\\(0.012)} & \makecell{0.066\\(0.011)} & \makecell{0.044\\(0.012)} & \makecell{0.082\\(0.017)} \\ \midrule[0.2pt]
				\textbf{NDCG@1} ($ \uparrow $) & \makecell{0.722\\(0.023)} & \makecell{0.828\\(0.016)} & \makecell{0.836\\(0.015)} & \makecell{\underline{0.848}\\(0.013)} & \makecell{0.828\\(0.017)} & \makecell{\textbf{0.901}\\(0.011)} & \makecell{0.796\\(0.019)} & \makecell{0.842\\(0.015)} & \makecell{0.810\\(0.019)} & \makecell{0.848\\(0.016)} \\
				\bottomrule
			\end{tabular}
			
		\end{center}
	\end{subtable}

\end{table}

\begin{table}[!t]\centering
\setlength{\abovecaptionskip}{-0.5em}
	\caption{Small-To-Large testbed results for link prediction (top) and node classification (bottom)~tasks.
		Higher~($ \uparrow $) scores are better. The numbers in the parentheses denote one standard error.
		The \textbf{best} result is in bold, and the \underline{second best} result is underlined.}
	\label{appendix:tab:results:small-to-large}
	\setlength{\tabcolsep}{3pt}
\fontsize{7.5}{8.5}\selectfont
	
	\begin{subtable}[h]{1.0\textwidth}
		\setlength{\abovecaptionskip}{-0.5em}
		\subcaption{Link prediction}
		\label{appendix:tab:results:small-to-large:linkpred}
		\begin{center}
			\begin{tabular}{lcccccccccc}\toprule
				\textbf{Perf. Metric}&\makecell[c]{\textbf{\rs}} &\makecell[c]{\textbf{\gbperf}} &\makecell[c]{\textbf{\gbrank}} &\makecell[c]{\textbf{\isac}} &\makecell[c]{\textbf{\as}} &\makecell[c]{\textbf{\ss}} &\makecell[c]{\textbf{\alors}} &\makecell[c]{\textbf{\ncf}} &\makecell[c]{\textbf{\metaod}} &\makecell[c]{\textbf{\metagl}} \\\midrule
				\textbf{AUC} ($ \uparrow $)	& \makecell{0.522\\(0.027)} & \makecell{0.798\\(0.017)} & \makecell{0.797\\(0.017)} & \makecell{\underline{0.842}\\(0.018)} & \makecell{0.827\\(0.022)} & \makecell{0.767\\(0.020)} & \makecell{0.812\\(0.024)} & \makecell{0.796\\(0.017)} & \makecell{0.667\\(0.029)} & \makecell{\textbf{0.870}\\(0.018)} \\ \midrule[0.2pt]
				\textbf{MRR} ($ \uparrow $) & \makecell{0.031\\(0.009)} & \makecell{0.072\\(0.011)} & \makecell{0.061\\(0.011)} & \makecell{0.132\\(0.019)} & \makecell{\textbf{0.368}\\(0.036)} & \makecell{0.074\\(0.016)} & \makecell{0.209\\(0.029)} & \makecell{0.047\\(0.005)} & \makecell{0.075\\(0.012)} & \makecell{\underline{0.260}\\(0.031)} \\ \midrule[0.2pt]
				\textbf{NDCG@1} ($ \uparrow $) & \makecell{0.841\\(0.011)} & \makecell{0.958\\(0.004)} & \makecell{\underline{0.960}\\(0.004)} & \makecell{0.957\\(0.005)} & \makecell{0.951\\(0.008)} & \makecell{0.953\\(0.005)} & \makecell{0.947\\(0.007)} & \makecell{0.956\\(0.004)} & \makecell{0.921\\(0.008)} & \makecell{\textbf{0.964}\\(0.006)} \\
				\bottomrule
			\end{tabular}
			
			\vspace{-0.5em}
		\end{center}
	\end{subtable}
	\begin{subtable}[h]{1.0\textwidth}
		\setlength{\abovecaptionskip}{-0.5em}
		\subcaption{Node classification}
		\label{appendix:tab:results:small-to-large:nodeclass}
		\begin{center}
			\begin{tabular}{lcccccccccc}\toprule
				\textbf{Perf. Metric}&\makecell[c]{\textbf{\rs}} &\makecell[c]{\textbf{\gbperf}} &\makecell[c]{\textbf{\gbrank}} &\makecell[c]{\textbf{\isac}} &\makecell[c]{\textbf{\as}} &\makecell[c]{\textbf{\ss}} &\makecell[c]{\textbf{\alors}} &\makecell[c]{\textbf{\ncf}} &\makecell[c]{\textbf{\metaod}} &\makecell[c]{\textbf{\metagl}} \\\midrule
				\textbf{AUC} ($ \uparrow $)	& \makecell{0.508\\(0.039)} & \makecell{0.724\\(0.042)} & \makecell{0.726\\(0.042)} & \makecell{0.711\\(0.044)} & \makecell{\textbf{0.761}\\(0.037)} & \makecell{0.664\\(0.041)} & \makecell{0.701\\(0.035)} & \makecell{0.697\\(0.042)} & \makecell{0.467\\(0.043)} & \makecell{\underline{0.736}\\(0.037)} \\ \midrule[0.2pt]
				\textbf{MRR} ($ \uparrow $) & \makecell{0.011\\(0.002)} & \makecell{0.058\\(0.021)} & \makecell{0.082\\(0.023)} & \makecell{\textbf{0.109}\\(0.025)} & \makecell{\underline{0.095}\\(0.029)} & \makecell{0.036\\(0.009)} & \makecell{0.042\\(0.013)} & \makecell{0.034\\(0.008)} & \makecell{0.016\\(0.007)} & \makecell{0.071\\(0.024)} \\ \midrule[0.2pt]
				\textbf{NDCG@1} ($ \uparrow $) & \makecell{0.795\\(0.032)} & \makecell{0.861\\(0.028)} & \makecell{\underline{0.896}\\(0.025)} & \makecell{\textbf{0.902}\\(0.024)} & \makecell{0.883\\(0.023)} & \makecell{0.855\\(0.024)} & \makecell{0.862\\(0.028)} & \makecell{0.864\\(0.029)} & \makecell{0.830\\(0.031)} & \makecell{0.857\\(0.027)} \\
				\bottomrule
			\end{tabular}
			
		\end{center}
	\end{subtable}

\end{table}

\section{Model Selection Algorithms}\label{app:algorithms}

Here we provide details of the model selection algorithms included in \benchmark.

\textbf{Random Selection} assigns random scores to the models, and
thus chooses the best model purely randomly without considering prior model performances and meta-graph features.

\textbf{Global Best (GB)-AvgPerf}~\cite{DBLP:conf/nips/ZhaoRA21} computes the average performance of each model over all observed graphs, and 
selects the one with the largest average performance.
Thus this algorithm is independent of meta-graph features.

\textbf{Global Best (GB)-AvgRank}~\cite{park2023metagl} computes the rank (in percentile) of all models for each graph,
where a higher performance is assigned a larger rank percentile, and 
selects the model that has the largest average rank percentile over observed graphs.
Thus this algorithm is also independent of meta-graph features.

\textbf{\isac}~\cite{DBLP:conf/ecai/KadiogluMST10} clusters observed graphs into $ k $ groups in the space of meta-graph features, and
when given a new graph, finds the cluster nearest to the given graph and 
selects the model that obtained the highest average performance over the observed graphs in that cluster.
$ k $ is set to 5 in our experiments.

\textbf{\asfull (\as)}~\cite{nikolic2013simple} finds the observed graph, which is the most similar to the test graph 
in terms of meta-graph features, and selects the model that had the best performance on that graph.

\textbf{Supervised Surrogates (\ss)}~\cite{xu2012satzilla2012} optimizes a surrogate model (\ie, a regressor)
to transform meta-graph features into model performances. 
We use a two-layer feedforward neural network with ReLU nonlinearity as the surrogate model.

\textbf{\alors}~\cite{DBLP:journals/ai/MisirS17} learns latent factors of the observed graphs and models 
by performing low-rank nonnegative matrix factorization on the performance matrix, and then 
optimizes a non-linear regressor to map meta-graph features into latent graph factors.
Given a new graph, it predicts the new graph's latent factor, and 
estimates the model performances on the new graph to be the dot product between the estimated graph factor and the learned model factor.

\textbf{\ncf}~\cite{DBLP:conf/www/HeLZNHC17} adapts \alors by replacing the dot product operation used in \alors 
with a more flexible neural network model,
which predicts model performances by jointly employing the linearity of matrix factorization and nonlinearity of deep neural networks.

\textbf{\metaod}~\cite{DBLP:conf/nips/ZhaoRA21} improves \alors for the model selection task,
\eg, by employing a rank-based meta-learning objective instead of the usual reconstruction objective. 
As the first method for unsupervised outlier model selection, 
it also presents specialized meta-features to capture the outlying characteristics of non-graph data.

\textbf{\metagl}~\cite{park2023metagl} extends \metaod for GL model selection
by designing meta-graph features to capture the structural characteristics of a graph, 
employing the top-1 probability as a meta-training objective, and 
modeling the relations between graphs and models in the form of a heterogeneous graph and learning latent factors by applying graph neural networks on it.

\section{Meta-Graph Features}\label{app:metafeats}

\subsection{Time Complexity Analysis
}
\label{appendix:metafeats:timecomplexity}

Let $ G=(V, E) $ be a graph, where $ V $ and $ E $ denote the set of nodes and edges in graph~$ G $, respectively.
Overall, the time complexity of generating meta-graph features for graph $ G $ is 
\begin{align}
	\mathcal{O}(k|E|\Delta)
\end{align}
where $ k $ is the number of feature extractors in the extractor set $ |\Psi| $, and $ \Delta $ denotes a small constant.

Each feature extractor in $ \Psi $ extracts a specific graph structural feature, 
such as network motifs and PageRank scores, as well as other graph statistics such as the density of a graph.
We estimate the frequency of all network motifs with $\{2,3,4\}$-nodes, 
which can be done in $\mathcal{O}(|E|\Delta)$ time, 
where $ \Delta $ is a small constant representing the maximum sampled degree, which can be set by the user. 
For further details, please refer to~\cite{ahmed2016estimation}.
Other structural features, such as PageRank, take $\mathcal{O}(|E|)$ time, and graph 
statistics, such as the number of nodes and edges and the graph density, can be obtained even more efficiently.
Assuming $ k $ such feature extractors, overall it takes $ \mathcal{O}(k|E|\Delta) $ time.

Note that \benchmark aims to provide a representative and diverse set of graph features,
which have been proven effective in previous studies.
At the same time, our framework is flexible, and can support any set of meta-graph features.
Thus, depending on the task, more computationally expensive, yet informative new features might be additionally used, 
or the set of features may be limited to only those that can be efficiently computed in time strictly linear in the number of edges, \ie, $\mathcal{O}(|E|)$,
in which case the $ \Delta $ term would be dropped, and we have $ \mathcal{O}(k|E|) $.
Further, note that feature extractors are independent of each other, and can be computed in parallel.

\vspace{-0.5em}
\subsection{Statistical Functions}
\vspace{-0.5em}

\Cref{table-meta-features} lists the statistical functions $ \Sigma $, which derives a set of meta-graph features 
that summarize the statistical distribution of graph invariants,
such as the node degree distribution, or $ k $-core numbers. 
Such vectors have different sizes for different graphs as their size is determined by the number of nodes or edges of the graph.
The statistical functions $ \Sigma $ in \Cref{table-meta-features} transforms those vectors of varying length
into fixed-size meta-feature vectors.

\begin{table}[!t]
\centering
\caption{
	Statistical functions $ \Sigma $ to derive a set of meta-graph features from a graph invariant, \eg, degree distribution or $ k $-core numbers.
$\vx$ denotes a vector of arbitrary graph invariants  for some graph $G_i=(V_i,E_i)$, 
	such as node degree vector, and PageRank vector (\ie, PageRank scores of each node in $ G_i $).
	$\pi(\vx)$ denotes the sorted vector of $\vx$.
	} 
	\label{table-meta-features}
	\vspace{4mm}
	\small
	\begin{tabular}{@{}l l @{} H @{}H HHH  @{}} 
		\toprule
		\textbf{Function}  &
		\textbf{Equation} &
\textbf{Rationale} & \textbf{Variants} & 
\textbf{Representation} ($\vx$, ...) & 
\textbf{Variable Type} (Q,C,T) &
\\
		\midrule

        Min, Max & $\min(\vx)$, $\max(\vx)$ &  $-$ \\
Median  & $\mathsf{med}(\vx)$ &  Variable normality \\
        
        Geometric Mean  & $\abs{\vx}^{-1} \prod_i x_i $ & Variable normality & \\
        Harmonic Mean & $\abs{\vx} / \sum_i \frac{1}{x_i}$ & Variable normality & \\
Mean, Stdev, Variance  & $\mu_{\vx}$, $\sigma_{\vx}$, $\sigma^2_{\vx}$ & Variable normality & \\
Skewness  & $\nicefrac{\mathbb{E}(\vx - \mu_{\vx})^3}{\sigma^3_{\vx}}$ & Variable normality & \\
        Pearson Kurtosis  & $\textrm{Kurt}[\vx]=\nicefrac{\mathbb{E}(\vx - \mu_{\vx})^4}{\sigma^4_{\vx}}$ (biased/unbiased) & Variable normality &
        Fisher/Pearson, and bias/unbiased. \\
        Fisher Kurtosis  & $ \textrm{Kurt}[\vx]-3.0 $ (biased/unbiased) & Variable normality &
        Fisher/Pearson, and bias/unbiased. \\

        \midrule

Quartile Dispersion Coeff. & $\frac{Q_3-Q_1}{Q_3+Q_1}$ & Dispersion & \\
        Median Absolute Deviation & $\mathsf{med}(\abs{\vx - \mathsf{med}(\vx)})$ & Dispersion & \\
Avg. Absolute Deviation & $\frac{1}{\abs{\vx}} \ve^T\!\abs{\vx - \mu_{\vx}} $ & Dispersion & \\

        Coeff. of Variation & $\nicefrac{\sigma_{\vx}}{\mu_{\vx}}$ & Dispersion & \\

        Efficiency Ratio & $\nicefrac{\sigma^2_{\vx}}{\mu^2_{\vx}}$ & Dispersion & \\
        Variance-to-Mean Ratio & $\nicefrac{\sigma^2_{\vx}}{\mu_{\vx}}$ & Dispersion & \\
\midrule
        
        Signal-to-Noise Ratio (SNR)  & $\nicefrac{\mu^2_{\vx}}{\sigma^2_{\vx}}$ & Noisiness of data  & \\
Entropy & $H(\vx) = -\sum_{i} \; x_i \log x_i$ & Variable Informativeness  & \\
        Norm. Entropy & ${H(\vx)} / {\log_2 \abs{\vx}}$ & Variable Informativeness  & \\
        Gini Coefficient & 
        $\nicefrac{\sum_{i=1}^{\abs{\vx}} (2i-\abs{\vx}-1) \pi(\vx)_i}{n \sum_{i=1}^{\abs{\vx}} \pi(\vx)_i} $
        & Variable Informativeness & \\

		\midrule
$Q_1$, $Q_3$  & median of the $\abs{\vx}/2$ smallest (largest) values & $-$ & \\
IQR  & $Q_3 - Q_1$ & $-$ & \\
		Outlier LB $\alpha \in \{1.5,3\}$ & $ Q_1-\alpha IQR$ & Data noisiness & \\
		Outlier UB $\alpha \in \{1.5,3\}$ & $ Q_3+\alpha IQR$ & Data noisiness & \\
		\makecell[l]{Outliers Count $\alpha \in \{1.5,3\}$,\\$~~~~~~(\beta,\gamma) \in \{(1,0),(0,1),(1,1)\}$} & 
		$\beta \cdot \sum_{i} \mathbb{I}(x_i \!<\! Q_1\!-\!\alpha IQR) + \gamma \cdot \sum_{i} \mathbb{I}(x_i \!>\! Q_3+\alpha IQR)$ 
		& Data noisiness & \\
		\makecell[l]{Outliers Frac. $\alpha \in \{1.5,3\}$,\\$~~~~~~(\beta,\gamma) \in \{(1,0),(0,1),(1,1)\}$} & 
		$(\beta \cdot \sum_{i} \mathbb{I}(x_i \!<\! Q_1\!-\!\alpha IQR) + \gamma \cdot \sum_{i} \mathbb{I}(x_i \!>\! Q_3+\alpha IQR)) / \abs{\vx}$ 
		& Data noisiness & \\
		
		\makecell[l]{($\alpha$-std) Outliers Count $\alpha \in \{1,2,3\}$,\\ $~~~~~~(\beta,\gamma) \in \{(1,0),(0,1),(1,1)\}$}        &  
		$\beta \cdot \sum_{i} \mathbb{I}(x_i \!<\! \mu_{\vx} - \alpha \sigma_{\vx}) + \gamma \cdot \sum_{i} \mathbb{I}(x_i \!>\! \mu_{\vx} + \alpha \sigma_{\vx})$ 
		& Data noisiness             & $o/\abs{\vx}$, lb, ub, total \\
		
		\makecell[l]{($\alpha$-std) Outliers Frac. $\alpha \in \{1,2,3\}$,\\ $~~~~~~(\beta,\gamma) \in \{(1,0),(0,1),(1,1)\}$}        &  
		$(\beta \cdot \sum_{i} \mathbb{I}(x_i \!<\! \mu_{\vx} - \alpha \sigma_{\vx}) + \gamma \cdot \sum_{i} \mathbb{I}(x_i \!>\! \mu_{\vx} + \alpha \sigma_{\vx})) / \abs{\vx}$ 
		& Data noisiness             & $o/\abs{\vx}$, lb, ub, total \\

        \midrule
        
        Mode & modal (most common) value in $ \vx $ \\
        Mode Count & count for the modal value in $ \vx $ \\
        Mode Frac. & mode count of $ \vx $ / $ \abs{\vx} $ \\

        \bottomrule
	\end{tabular}
\end{table}

\section{Experimental Settings and Details}\label{app:expsettings}

\textbf{Hardware.} Experiments were performed on a Linux server on AWS, running Ubuntu 20.04.5 LTS
with Intel Xeon Platinum 8275CL CPUs @ 3.00GHz, 1.1TB RAM, and NVIDIA A100 SXM4 GPUs with 40GB memory.

\textbf{Performance Evaluation.}
To evaluate the performance of optimizable GL methods, such as GCN~\cite{DBLP:conf/iclr/KipfW17} and GraphSAGE~\cite{DBLP:conf/nips/HamiltonYL17}, 
we trained these methods for up to 300 epochs, using Adam optimizer with a learning rate of 0.001, and
applying a validation-based early stopping with a patience of 30 epochs.
A few graphs have multiple labels for each node. For those graphs, we found that using a larger learning rate and patience leads to a better performance.
So for the multi-label node classification datasets, we used a learning rate of 0.01 and a patience of 60 epochs.
As an early stopping criterion, we used ROC AUC for link prediction, and 
used accuracy for node classification (or weighted average precision for multi-label node classification).
Not all graphs come with input node features. 
For those graphs without input node features, we used randomly initialized embeddings of size 32 as input node embeddings, 
and let those embeddings optimized during model training.

\textbf{Meta-Graph Features.}
We present three sets of meta-graph features in the main text, \ie, \Mregular, \Mgraphlets, and \Mcompact.
The experimental results reported in the main text were obtained using \Mregular.
We also report results obtained with three different sets of meta-graph features, 
\ie, \Mgraphlets, \Mcompact, and \Mreggraph, in \Cref{app:results:metagraphfeats}.

\textbf{Graph Learning (GL) Methods.} 
In the evaluation using the proposed testbeds, model selection algorithms aim to predict the best model from the set 
of differently configured GL models, that~is,
GCN~\cite{DBLP:conf/iclr/KipfW17}, GraphSAGE~\cite{DBLP:conf/nips/HamiltonYL17}, GAT~\cite{DBLP:conf/iclr/VelickovicCCRLB18},
GIN~\cite{DBLP:conf/iclr/XuHLJ19}, EGC~\cite{DBLP:conf/iclr/TailorOLL22}, SGC~\cite{DBLP:conf/icml/WuSZFYW19},
ChebNet~\cite{DBLP:conf/nips/DefferrardBV16}, PNA~\cite{DBLP:conf/nips/CorsoCBLV20}, 
DGI~\cite{DBLP:conf/iclr/VelickovicFHLBH19}, 
spectral embedding~\cite{DBLP:journals/pr/LuoWH03},
GraRep~\cite{DBLP:conf/cikm/CaoLX15}, 
node2vec~\cite{DBLP:conf/kdd/GroverL16},
label propagation~\cite{Zhu2002LearningFL},
Jaccard's Coeff.~\cite{DBLP:journals/jasis/Liben-NowellK07},
Resource Alloc.~\cite{zhou2009predicting},
Adamic/Adar~\cite{DBLP:journals/jasis/Liben-NowellK07}, and
SEAL~\cite{DBLP:conf/nips/ZhangC18}.
Their hyperparameter settings are provided in Table 2 in the main text.
We implemented spectral embedding~\cite{DBLP:journals/pr/LuoWH03} and GraRep~\cite{DBLP:conf/cikm/CaoLX15},
using NumPy\footnote{\label{numpy}\url{https://numpy.org/}} and SciPy\footnote{\url{https://scipy.org/}}.
We used GRAPE~\cite{cappelletti2021grape} for the implementation of node2vec~\cite{DBLP:conf/kdd/GroverL16}.
We used NetworkX~\cite{hagberg2008exploring} for the implementation of classical link prediction methods, \ie, 
Jaccard's Coeff.~\cite{DBLP:journals/jasis/Liben-NowellK07},
Resource Alloc.~\cite{zhou2009predicting}, and
Adamic/Adar~\cite{DBLP:journals/jasis/Liben-NowellK07}.
We used PyTorch Geometric (PyG)~\cite{DBLP:journals/corr/abs-1903-02428} for the implementation of other GL methods,
\eg, GCN~\cite{DBLP:conf/iclr/KipfW17} and GraphSAGE~\cite{DBLP:conf/nips/HamiltonYL17}.
Spectral embedding and GraRep use SciPy's functionalities to find eigenvalue/eigenvectors and perform singular value decomposition, respectively;
for these methods, we used the default parameter values specified by SciPy, except for the parameters that we vary in creating the model set $ \modelSet $.
Similarly, for methods supported by PyG, we used their default parameter settings in the corresponding package,
while varying a few important parameters to create the model set $ \modelSet $.

\textbf{Model Selection Algorithms.}
In experiments, we evaluate \numBaselinesInWords model selection algorithms, that is,
Random Selection, \gbperffull~\cite{DBLP:conf/nips/ZhaoRA21}, \gbrankfull~\cite{park2023metagl}, 
\isac~\cite{DBLP:conf/ecai/KadiogluMST10}, \asfull (\as)~\cite{nikolic2013simple}, 
Supervised Surrogates (\ss)~\cite{xu2012satzilla2012}, \alors~\cite{DBLP:journals/ai/MisirS17}, \ncf~\cite{DBLP:conf/www/HeLZNHC17}, \metaod~\cite{DBLP:conf/nips/ZhaoRA21}, and \metagl~\cite{park2023metagl}.
For \metagl~\cite{park2023metagl} and \metaod~\cite{DBLP:conf/nips/ZhaoRA21}, we used the authors' implementation.
We adapted \metaod's implementation so that it can work with a sparse performance matrix.
We implemented other model selection algorithms included in \benchmark in python 
using open source libraries such as NumPy\footref{numpy} and DGL\footnote{\url{http://dgl.ai/}}.
Global Best methods perform a global averaging of the performance matrix. 
Given sparse performance matrices, they average over observed entries alone and ignore missing entries.
\isac~\cite{DBLP:conf/ecai/KadiogluMST10} applies $ k $-means algorithm to meta-graph features to cluster observed graphs into $ 5 $ groups.
\as~\cite{nikolic2013simple} uses cosine similarity scoring to find the $ 1 $-NN observed graph.
\ss~\cite{xu2012satzilla2012} uses Adam optimizer to train an MLP regressor with two hidden layers, 
which is optimized to transform meta-graph features into model performances.
\alors~\cite{DBLP:journals/ai/MisirS17} learns latent embeddings by using nonnegative matrix factorization, and 
uses an MLP regressor with two hidden layers to transform meta-graph features into latent graph factors.
\ncf~\cite{DBLP:conf/www/HeLZNHC17} produces latent graph and model embedding by using an MLP regressor with two hidden layers,
which is optimized via Adam optimizer with a learning rate of 0.01 and a weight decay of 0.0001.
\metaod~\cite{DBLP:conf/nips/ZhaoRA21} uses the default parameter settings given by the original implementation, 
\eg, a random forest regressor with 100 estimators and a max depth of 10.
\metagl~\cite{park2023metagl} uses heterogeneous graph transformer (HGT)~\cite{DBLP:conf/www/HuDWS20} as a graph encoder, 
which contains 2~layers and 4~attention heads per layer. In its G-M network, nodes are connected to their top-30 most similar nodes.
The \metagl model is optimized with Adam optimizer with a learning rate of 0.00075 and a weight decay of 0.0001.
For optimizable algorithms discussed above, which involve learning low-dimensional embeddings, we consistently set the embedding size to 32.

\vspace{-1.0em}
\section{Usage and Extensibility}\label{app:extensibility}
\vspace{-0.5em}

\definecolor{codegreen}{rgb}{0,0.6,0}
\definecolor{codegray}{rgb}{0.5,0.5,0.5}
\definecolor{codepurple}{rgb}{0.58,0,0.82}
\definecolor{backcolour}{rgb}{0.95,0.95,0.92}

\lstdefinestyle{mystyle}{
	backgroundcolor=\color{backcolour},   
	commentstyle=\color{codegreen},
	keywordstyle=\color{magenta},
	numberstyle=\tiny\color{codegray},
	stringstyle=\color{codepurple},
	basicstyle=\ttfamily\footnotesize,
	breakatwhitespace=false,         
	breaklines=true,                 
	captionpos=b,                    
	keepspaces=true,                 
	numbers=left,                    
	numbersep=5pt,                  
	showspaces=false,                
	showstringspaces=false,
	showtabs=false,                  
	tabsize=2
}

\lstset{style=mystyle}

We describe how \benchmark can be used and extended with graphs, models, and performance records.

\vspace{-0.5em}
\subsection{Graphs}\label{app:extensibility:graphs}
To support graph data from multiple data sources, graphs in \benchmark are represented~by the
\Colorbox{backcolour}{\lstinline[language=python]|graphs.graphset.Graph|} class,
which is a wrapper class that holds the graph data from different sources, and 
provides auxiliary methods serving as a common interface to different types of graph~data.

The set of graphs in \benchmark is represented by the 
\Colorbox{backcolour}{\lstinline[language=python]|graphs.graphset.GraphSet|} class,
which provides the functionality to load graphs for a certain graph learning task like node classification, 
as well as graphs from specific data sources, 
such as the Network Repository (NetRepo)~\cite{DBLP:conf/aaai/RossiA15} and PyTorch Geometric (PyG)~\cite{DBLP:journals/corr/abs-1903-02428},
as shown below.
\begin{lstlisting}[language=Python,upquote=true]
from graphs.graphset import GraphSet
graph_set = GraphSet(['netrepo', 'pyg'])
node_classification_graphs = graph_set.node_classification_graphs()
\end{lstlisting}

\textbf{Adding New Graphs From Existing Sources.} 
We currently provide code to incorporate~graphs from the two popular graph repositories, \ie, NetRepo and PyG.
Adding new graphs from these repositories can be done easily by instantiating a 
\Colorbox{backcolour}{\lstinline[language=python]|graphs.graphset.Graph|}
object for the new graph, and adding it to the list returned by the 
\Colorbox{backcolour}{\lstinline[language=python]|load_graphs|} functions (e.g,. 
\Colorbox{backcolour}{\lstinline[language=python]|graphs.pyg_graphs.load_graphs|})

\textbf{Adding New Graphs From New Sources} (\eg, a new graph repository or your own graph data) can be done
by (1) adding a script for the new data source in the \Colorbox{backcolour}{\lstinline[language=python]|graphs|} package, 
which will parse the raw graph data, and construct a \Colorbox{backcolour}{\lstinline[language=python]|graphs.graphset.Graph|} object for the new graph, 
and (2) registering the new data source in the \Colorbox{backcolour}{\lstinline[language=python]|graphs.graphset.GraphSet|} class.

\vspace{-0.5em}
\subsection{Models}\label{app:extensibility:models}
\vspace{-0.5em}
The set of models included in \benchmark is defined by the \Colorbox{backcolour}{\lstinline[language=python]|models.modelset|} package.
Each graph learning method and its hyperparameter settings to be searched over are defined by a separate class that inherits from the 
\Colorbox{backcolour}{\lstinline[language=python]|models.modelset.ModelSettings|} class. 
For example, the following code defines the GAT model set.
\begin{lstlisting}[language=Python,upquote=true]
class GATModelSettings(ModelSettings):
	def __init__(self):
		super().__init__()
		self.variable_hyperparams = ModelSettings.alpha_ordered_dict({
			'hidden_channels': [16, 64],
			'num_layers': [1, 2, 3],
			'dropout': [0.0, 0.5],
			'heads': [1, 4],
			'concat': [True, False],
		})
	
	@classmethod
	def load_model(cls, in_channels, out_channels, **params):
		return GAT(in_channels=in_channels, out_channels=out_channels, **params)
\end{lstlisting}
In the above code snippet, a GAT model instance, instantiated with particular hyperparameter settings, is obtained via the 
\Colorbox{backcolour}{\lstinline[language=python]|models.modelset.ModelSettings.load_model|} method.

\textbf{Adding New Hyperparameter Settings to Existing Models} can be done by simply adding additional hyperparameter settings
to the corresponding \Colorbox{backcolour}{\lstinline[language=python]|models.modelset.ModelSettings|} class.

\textbf{Adding New Models} can be done by 
(1) creating a new \Colorbox{backcolour}{\lstinline[language=python]|ModelSettings|} class for the new graph learning model,
(2) specifying the set of hyperparameter settings to be searched over, and
(3) completing the \Colorbox{backcolour}{\lstinline[language=python]|load_model|} method, such that new instantiated model object is to be returned,
as in the above code.

\vspace{-0.5em}
\subsection{Performance Records}\label{app:extensibility:perf}
\vspace{-0.25em}

\textbf{Data Splits.} For consistent comparisons among different models, \benchmark provides the data splits used in the evaluation.
The \Colorbox{backcolour}{\lstinline[language=python]|graphs.datasplit.DataSplit|} class provides functionalities
to generate and load data splits (\eg, node splits for node classification, and edge splits for link prediction), and
the generated data splits are included in \benchmark.
Then these previously generated data splits are used when evaluating new GL models on the existing graphs in the benchmark.

\textbf{Model Training and Evaluation.}
Given the instantiated GL model and the graph data, 
model training and evaluation is taken care of by the 
\Colorbox{backcolour}{\lstinline[language=python]|models.trainer.Trainer|} class.
The same code can be used to train and evaluate new GL models on either existing or new graphs, 

\textbf{Parallel Processing.}
To perform the training and evaluation of multiple <model, graph> pairs,
the aforementioned \Colorbox{backcolour}{\lstinline[language=python]|models.trainer.Trainer|} class is repeatedly executed by the 
\Colorbox{backcolour}{\lstinline[language=python]|performances.taskrunner.Runner|} class.
The \Colorbox{backcolour}{\lstinline[language=python]|Runner|} class is designed to support parallel processing,
such that multiple processes pick up an unevaluated <model, graph> pair, and perform model training and evaluation in parallel.

\section{Additional Details of the Dataset}\label{app:data}

\subsection{Data Overview}\label{app:data:overview}

\benchmark provides the following sets of data.
\begin{enumerate}[label=(\alph*),leftmargin=1.6em,topsep=0pt,itemsep=0pt]
\item\label{app:data:overview:perf} \textbf{Performance records}: performance of graph learning (GL) models on various graphs, measured in multiple metrics for the following GL tasks.
\begin{itemize}[leftmargin=1em,topsep=0pt,itemsep=0pt]
	\item Node classification performances
	\item Link prediction performances
\end{itemize}
\item\label{app:data:overview:graphsplits} \textbf{Graph data splits} used for evaluating GL models.
	\begin{itemize}[leftmargin=1em,topsep=0pt,itemsep=0pt]
	\item node splits for node classification
	\item edge splits for link prediction
	\end{itemize}
\item\label{app:data:overview:testbedsplits} \textbf{Testbed data splits} (\ie, splitting over the performance matrix and the meta-graph features) to evaluate model selection algorithms.
\item\label{app:data:overview:metafeats} \textbf{Meta-graph features}: we provide the following sets of meta-graph features.
	\begin{itemize}[leftmargin=1em,topsep=0pt,itemsep=0pt]
		\item \Mregular
		\item \Mgraphlets
		\item \Mcompact
		\item \Mreggraph
	\end{itemize}
\end{enumerate}

Performance records~\ref{app:data:overview:perf},
splitting of graph data~\ref{app:data:overview:graphsplits} and testbed data~\ref{app:data:overview:testbedsplits}, and
meta-graph features~\ref{app:data:overview:metafeats} were generated by \benchmark
by processing the graph data~(\Cref{app:data:graph}).
For details of these four types of data, please refer to the main text.
We give further description of the graph data used in the benchmark in the next subsection.

\textbf{No Personal or Offensive Contents.}
Note that \benchmark does not include personal data (\eg, personally identifiable information) or offensive contents in all the data described above (\ref{app:data:overview:perf} to \ref{app:data:overview:metafeats}).

\begin{figure}[b!]
\centering
	\includegraphics[width=0.6\linewidth]{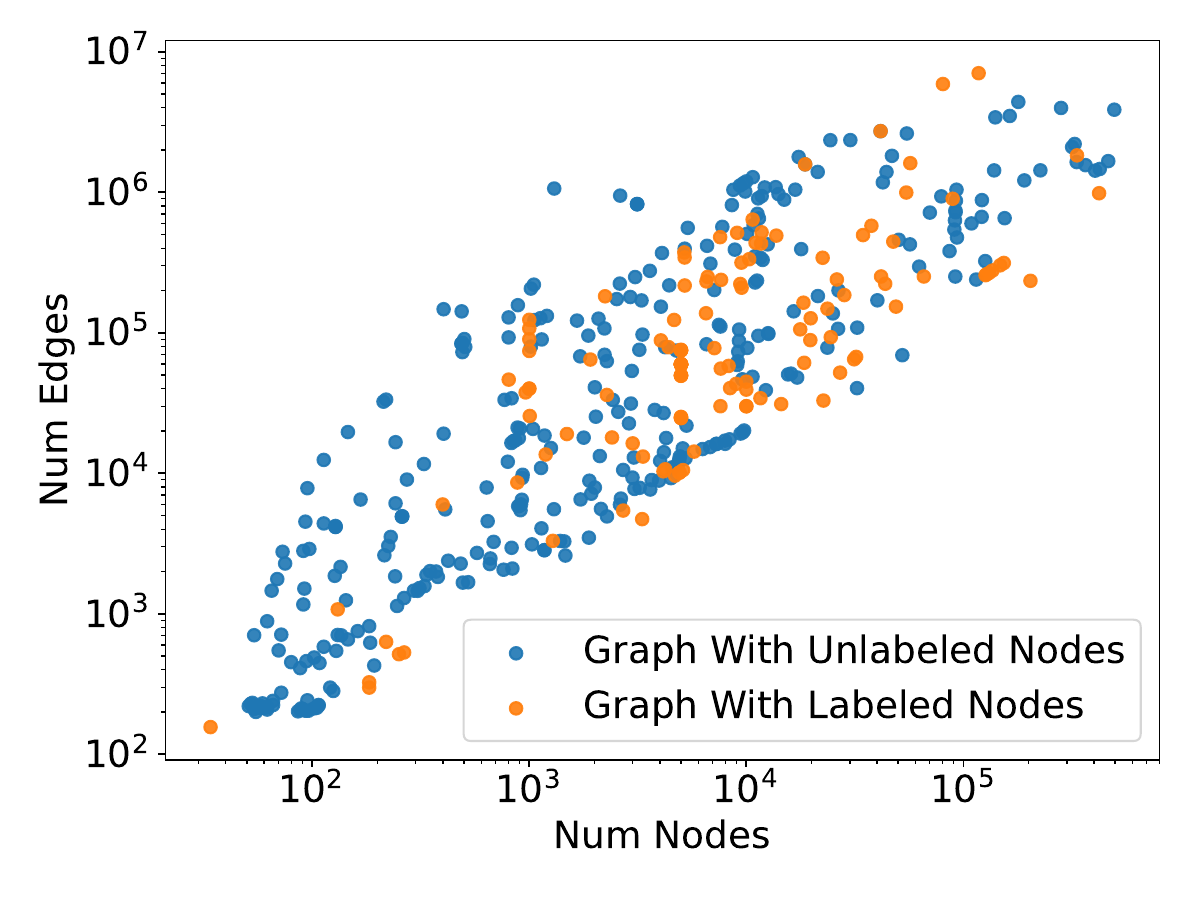}
\setlength{\abovecaptionskip}{0.1em}
	\caption{Distribution of the graphs in terms of the number of nodes (x-axis) and edges~(y-axis).
		Each dot corresponds to a graph, and is colored depending on whether the nodes in the graph have class labels or not.} 
	\label{fig:graphsizedist}
\end{figure}

\begin{table}[!h]\centering
\setlength{\abovecaptionskip}{0.5em}
	\renewcommand{\arraystretch}{1.1} \caption{
		Distribution of the graphs in \benchmark per graph domain. 
		In total, \benchmark currently covers \numGraphs graphs drawn from \numGraphDomains domains.
	}
	\label{tab:graphs_per_domain}
\fontsize{8.5}{9.5}\selectfont
\begin{tabular}{c c c c c c}\toprule
			\makecell[c]{\textbf{Graph Domain}} & \makecell[c]{\textbf{Count}}  & \makecell[c]{\textbf{Graph Domain}} & \makecell[c]{\textbf{Count}}  & \makecell[c]{\textbf{Graph Domain}} & \makecell[c]{\textbf{Count}}  \\\midrule
			Social Networks & 53& Synthetic-ER & 15& Synthetic-SBM & 6\\ 
			Chemical & 40& Temporal Reachability & 14& Road & 5\\ 
			Protein & 26& Synthetic-RandPart & 12& Recommendation & 5\\ 
			Retweet & 26& Interaction & 10& Computer Vision & 4\\ 
			Biological Networks & 24& Proximity & 10& Flight & 3\\ 
			Web Graphs & 22& Brain Networks & 9& Misc & 3\\ 
			Economic Networks & 17& Power Networks & 8& Co-Purchase & 3\\ 
			Synthetic-BA & 17& Citation & 8& Scientific Computing & 3\\ 
			Friendship & 16& Email & 7& Infrastructure & 2\\ 
			Synthetic-CL & 16& Technology & 7& Coauthor & 2\\ 
			Collaboration & 15& Ecology & 6& Phone Call Networks & 1\\ 
			Synthetic-KPGM & 15& Wikipedia & 6&  & \\ 
			Synthetic-Others & 15& Knowledgebase & 6&  & \\ 
			
			\bottomrule
			\multicolumn{5}{r}{\textbf{Total Number of Graphs}} &\textbf{457} \\
		\end{tabular}
\end{table}

\subsection{Graph Data}\label{app:data:graph}

\textbf{Graph Data Sources.}
We collect graph data from two widely used public graph repositories, \ie, 
Network Repository (NetRepo)~\cite{DBLP:conf/aaai/RossiA15} and PyTorch Geometric (PyG)~\cite{DBLP:journals/corr/abs-1903-02428}.
In~\Cref{tab:graphs}, we provide the list of graphs in \benchmark, and specify the data source for each graph.

\textbf{Graph Data Overview.}
\Cref{fig:graphsizedist} shows the distribution of the graph in \benchmark in terms of the number of nodes (x-axis) and
the number of edge (y-axis), where each dot corresponds to a graph, and is colored depending on whether nodes are labeled or not.
\Cref{tab:graphs_per_domain} shows the distribution of the graphs per graph domain. 
In total, \numGraphs graphs drawn from \numGraphDomains domains are currently used for 
model performance evaluation.

\textbf{List of Graphs.}
\Cref{tab:graphs} lists the graphs in \benchmark, and provides information on the graph size, 
the number of node classes, data source, and graph domain.

\begin{normalsize}
\fontsize{7.0}{8.0}\selectfont

\begin{longtable}{r c r r r c c}
	\caption{List of the graphs.}\label{tab:graphs}\\
	\toprule
	\makecell[c]{\textbf{Id}} & \makecell[c]{\textbf{Graph}} & \makecell[c]{\textbf{\# Nodes}} & \makecell[c]{\textbf{\# Edges}} & \makecell[c]{\textbf{\# Node}\\\textbf{Classes}} & \makecell[c]{\textbf{Data}\\\textbf{Source}} & \makecell[c]{\textbf{Domain}}  \\
	\midrule
	\endfirsthead
	\multicolumn{7}{c}
	{\tablename\ \thetable\ -- \textit{Continued from the previous page}} \\
	\midrule
	\makecell[c]{\textbf{Id}} & \makecell[c]{\textbf{Graph}} & \makecell[c]{\textbf{\# Nodes}} & \makecell[c]{\textbf{\# Edges}} & \makecell[c]{\textbf{\# Node}\\\textbf{Classes}} & \makecell[c]{\textbf{Data}\\\textbf{Source}} & \makecell[c]{\textbf{Domain}}  \\
	\midrule
	\endhead
	\midrule \multicolumn{7}{r}{\textit{Continued on the next page}} \\
	\endfoot
	\bottomrule
	\endlastfoot
	1 & PLC-60-30-L2 & 117,572 & 7,045,181 & 2 & NetRepo & Synthetic-Others \\
	2 & soc-Flickr-ASU & 80,513 & 5,899,882 & 195 & NetRepo & Social Networks \\
	3 & sc-shipsec5 & 179,104 & 4,400,152 & \textsc{n/a} & NetRepo & Scientific Computing \\
	4 & web-Stanford & 281,903 & 3,985,272 & \textsc{n/a} & NetRepo & Web Graphs \\
	5 & soc-youtube & 495,957 & 3,873,496 & \textsc{n/a} & NetRepo & Social Networks \\
	6 & web-arabic-2005 & 163,598 & 3,494,538 & \textsc{n/a} & NetRepo & Web Graphs \\
	7 & sc-shipsec1 & 140,385 & 3,415,518 & \textsc{n/a} & NetRepo & Scientific Computing \\
	8 & LINKX-penn94 & 41,554 & 2,724,458 & 2 & PyG & Social Networks \\
	9 & socfb-Penn94 & 41,536 & 2,724,440 & \textsc{n/a} & NetRepo & Social Networks \\
	10 & sc-nasasrb & 54,870 & 2,622,454 & \textsc{n/a} & NetRepo & Scientific Computing \\
	11 & socfb-Michigan23 & 30,147 & 2,353,032 & \textsc{n/a} & NetRepo & Social Networks \\
	12 & socfb-UGA50 & 24,389 & 2,348,114 & \textsc{n/a} & NetRepo & Social Networks \\
	13 & web-NotreDame & 325,729 & 2,207,671 & \textsc{n/a} & NetRepo & Web Graphs \\
	14 & ca-dblp-2012 & 317,080 & 2,099,732 & \textsc{n/a} & NetRepo & Collaboration \\
	15 & Entities-BGS & 333,845 & 1,832,398 & 2 & PyG & Knowledgebase \\
	16 & KPGM-log16-16-trial1 & 46,872 & 1,818,168 & \textsc{n/a} & NetRepo & Synthetic-KPGM \\
	17 & socfb-Oklahoma97 & 17,425 & 1,785,056 & \textsc{n/a} & NetRepo & Social Networks \\
	18 & soc-twitter-follows-mun & 465,017 & 1,667,082 & \textsc{n/a} & NetRepo & Social Networks \\
	19 & ca-MathSciNet & 332,689 & 1,641,288 & \textsc{n/a} & NetRepo & Collaboration \\
	20 & AttributedGraph-PPI & 56,944 & 1,612,348 & 121 & PyG & Protein \\
	21 & socfb-Cornell5 & 18,660 & 1,581,554 & \textsc{n/a} & NetRepo & Social Networks \\
	22 & LINKX-cornell5 & 18,660 & 1,581,554 & 2 & PyG & Friendship \\
	23 & ia-dbpedia-team-bi & 365,492 & 1,560,266 & \textsc{n/a} & NetRepo & Interaction \\
	24 & rt-higgs & 425,008 & 1,465,617 & \textsc{n/a} & NetRepo & Retweet \\
	25 & ca-dblp-2010 & 226,413 & 1,432,920 & \textsc{n/a} & NetRepo & Collaboration \\
	26 & ia-wiki-trust-dir & 138,592 & 1,432,057 & \textsc{n/a} & NetRepo & Interaction \\
	27 & soc-twitter-follows & 404,719 & 1,426,638 & \textsc{n/a} & NetRepo & Social Networks \\
	28 & socfb-Virginia63 & 21,325 & 1,396,356 & \textsc{n/a} & NetRepo & Social Networks \\
	29 & KPGM-log16-12-trial1 & 44,241 & 1,395,310 & \textsc{n/a} & NetRepo & Synthetic-KPGM \\
	30 & BA-2\_3\_60 & 10,708 & 1,281,300 & \textsc{n/a} & NetRepo & Synthetic-BA \\
	31 & tech-RL-caida & 190,914 & 1,215,220 & \textsc{n/a} & NetRepo & Technology \\
	32 & BA-2\_21\_60 & 9,993 & 1,195,500 & \textsc{n/a} & NetRepo & Synthetic-BA \\
	33 & KPGM-log16-10-trial1 & 42,551 & 1,177,546 & \textsc{n/a} & NetRepo & Synthetic-KPGM \\
	34 & BA-2\_20\_60 & 9,691 & 1,159,260 & \textsc{n/a} & NetRepo & Synthetic-BA \\
	35 & BA-2\_15\_60 & 9,340 & 1,117,140 & \textsc{n/a} & NetRepo & Synthetic-BA \\
	36 & socfb-Syracuse56 & 13,653 & 1,087,964 & \textsc{n/a} & NetRepo & Social Networks \\
	37 & socfb-NotreDame57 & 12,155 & 1,082,678 & \textsc{n/a} & NetRepo & Social Networks \\
	38 & scc\_fb-messages & 1,303 & 1,063,786 & \textsc{n/a} & NetRepo & Temporal Reachability \\
	39 & socfb-UC33 & 16,808 & 1,044,294 & \textsc{n/a} & NetRepo & Social Networks \\
	40 & CL-100000-1d7-trial3 & 92,967 & 1,043,480 & \textsc{n/a} & NetRepo & Synthetic-CL \\
	41 & BA-2\_9\_60 & 8,717 & 1,042,380 & \textsc{n/a} & NetRepo & Synthetic-BA \\
	42 & socfb-Duke14 & 9,885 & 1,012,874 & \textsc{n/a} & NetRepo & Social Networks \\
	43 & GemsecDeezer-HR & 54,573 & 996,404 & 84 & PyG & Friendship \\
	44 & LINKX-genius & 421,961 & 984,979 & 2 & PyG & Social Networks \\
	45 & socfb-JMU79 & 14,070 & 971,128 & \textsc{n/a} & NetRepo & Social Networks \\
	46 & scc\_twitter-copen & 2,623 & 947,228 & \textsc{n/a} & NetRepo & Temporal Reachability \\
	47 & BA-2\_23\_40 & 11,770 & 939,960 & \textsc{n/a} & NetRepo & Synthetic-BA \\
	48 & soc-slashdot-zoo & 79,120 & 935,738 & \textsc{n/a} & NetRepo & Social Networks \\
	49 & BA-2\_11\_40 & 11,337 & 905,320 & \textsc{n/a} & NetRepo & Synthetic-BA \\
	50 & Flickr & 89,250 & 899,756 & 7 & PyG & Computer Vision \\
	51 & socfb-UCSD34 & 14,948 & 886,442 & \textsc{n/a} & NetRepo & Social Networks \\
	52 & rec-github & 121,709 & 879,770 & \textsc{n/a} & NetRepo & Recommendation \\
	53 & CL-100000-1d8-trial3 & 92,402 & 871,580 & \textsc{n/a} & NetRepo & Synthetic-CL \\
	54 & econ-psmigr3 & 3,140 & 824,702 & \textsc{n/a} & NetRepo & Economic Networks \\
	55 & econ-psmigr1 & 3,140 & 824,702 & \textsc{n/a} & NetRepo & Economic Networks \\
	56 & econ-psmigr2 & 3,140 & 821,562 & \textsc{n/a} & NetRepo & Economic Networks \\
	57 & socfb-Yale4 & 8,578 & 810,900 & \textsc{n/a} & NetRepo & Social Networks \\
	58 & CL-100000-1d9-trial3 & 91,889 & 742,469 & \textsc{n/a} & NetRepo & Synthetic-CL \\
	59 & ia-wiki-Talk & 92,117 & 721,534 & \textsc{n/a} & NetRepo & Interaction \\
	60 & soc-slashdot & 70,068 & 717,294 & \textsc{n/a} & NetRepo & Social Networks \\
	61 & socfb-Cal65 & 11,247 & 702,716 & \textsc{n/a} & NetRepo & Social Networks \\
	62 & web-sk-2005 & 121,422 & 668,838 & \textsc{n/a} & NetRepo & Web Graphs \\
	63 & soc-douban & 154,908 & 654,324 & \textsc{n/a} & NetRepo & Social Networks \\
	64 & ER-2\_2\_50 & 11,429 & 651,760 & \textsc{n/a} & NetRepo & Synthetic-ER \\
	65 & BA-2\_24\_60-L2 & 10,693 & 639,750 & 2 & NetRepo & Synthetic-BA \\
	66 & CL-100000-2d0-trial1 & 91,471 & 631,153 & \textsc{n/a} & NetRepo & Synthetic-CL \\
	67 & ER-3\_19\_50 & 108,999 & 600,554 & \textsc{n/a} & NetRepo & Synthetic-ER \\
	68 & ER-2\_21\_50 & 10,804 & 584,286 & \textsc{n/a} & NetRepo & Synthetic-ER \\
	69 & GitHub & 37,700 & 578,006 & 2 & PyG & Friendship \\
	70 & socfb-Tulane29 & 7,752 & 567,836 & \textsc{n/a} & NetRepo & Social Networks \\
	71 & socfb-Wake73 & 5,372 & 558,382 & \textsc{n/a} & NetRepo & Social Networks \\
	72 & CL-100000-2d1-trial2 & 90,880 & 543,035 & \textsc{n/a} & NetRepo & Synthetic-CL \\
	73 & HeterophilousGraph-Tolokers & 11,758 & 519,000 & 2 & PyG & Collaboration \\
	74 & web-spam-detection & 9,072 & 514,700 & 3 & NetRepo & Web Graphs \\
	75 & ER-2\_8\_50 & 10,070 & 506,096 & \textsc{n/a} & NetRepo & Synthetic-ER \\
	76 & Coauthor-Physics & 34,493 & 495,924 & 5 & PyG & Coauthor \\
	77 & Amazon-Computers & 13,752 & 491,722 & 10 & PyG & Co-Purchase \\
	78 & AttributedGraph-Flickr & 7,575 & 479,476 & 9 & PyG & Social Networks \\
	79 & ia-wikiquote-user-edits & 93,445 & 476,865 & \textsc{n/a} & NetRepo & Interaction \\
	80 & rec-yelp-user-business & 50,395 & 459,208 & \textsc{n/a} & NetRepo & Recommendation \\
	81 & GemsecDeezer-HU & 47,538 & 445,774 & 84 & PyG & Friendship \\
	82 & PLC-40-30-L5 & 11,025 & 437,979 & 5 & NetRepo & Synthetic-Others \\
	83 & WikiCS & 11,701 & 431,726 & 10 & PyG & Wikipedia \\
	84 & soc-brightkite & 56,739 & 425,890 & \textsc{n/a} & NetRepo & Social Networks \\
	85 & KPGM-log14-16-trial3 & 12,545 & 425,872 & \textsc{n/a} & NetRepo & Synthetic-KPGM \\
	86 & socfb-UChicago30 & 6,591 & 416,206 & \textsc{n/a} & NetRepo & Social Networks \\
	87 & web-wiki-squirrel & 5,201 & 396,846 & \textsc{n/a} & NetRepo & Web Graphs \\
	88 & ca-AstroPh & 17,903 & 393,944 & \textsc{n/a} & NetRepo & Collaboration \\
	89 & ER-2\_14\_50 & 8,851 & 390,774 & \textsc{n/a} & NetRepo & Synthetic-ER \\
	90 & ER-3\_18\_50 & 86,337 & 381,446 & \textsc{n/a} & NetRepo & Synthetic-ER \\
	91 & LINKX-johnshopkins55 & 5,180 & 373,172 & 2 & PyG & Social Networks \\
	92 & socfb-Rice & 4,087 & 369,656 & \textsc{n/a} & NetRepo & Social Networks \\
	93 & scc\_infect-dublin & 10,972 & 351,146 & \textsc{n/a} & NetRepo & Temporal Reachability \\
	94 & AttributedGraph-BlogCatalog & 5,196 & 343,486 & 6 & PyG & Social Networks \\
	95 & FacebookPagePage & 22,470 & 342,004 & 4 & PyG & Web Graphs \\
	96 & web-wiki-crocodile & 11,631 & 341,691 & \textsc{n/a} & NetRepo & Web Graphs \\
	97 & soc-BlogCatalog-ASU & 10,312 & 333,983 & 39 & NetRepo & Social Networks \\
	98 & KPGM-log14-12-trial1 & 11,893 & 330,072 & \textsc{n/a} & NetRepo & Synthetic-KPGM \\
	99 & road-usroads-48 & 126,146 & 323,900 & \textsc{n/a} & NetRepo & Road \\
	100 & Twitch-DE & 9,498 & 315,774 & 2 & PyG & Friendship \\
	101 & Tox21-p53 & 153,563 & 314,046 & 47 & NetRepo & Chemical \\
	102 & socfb-UC64 & 6,833 & 310,664 & \textsc{n/a} & NetRepo & Social Networks \\
	103 & Tox21-AHR & 147,772 & 302,188 & 49 & NetRepo & Chemical \\
	104 & tech-p2p-gnutella & 62,561 & 295,756 & \textsc{n/a} & NetRepo & Technology \\
	105 & Tox21-HSE & 136,239 & 277,682 & 47 & NetRepo & Chemical \\
	106 & socfb-Wesleyan43 & 3,593 & 276,070 & \textsc{n/a} & NetRepo & Social Networks \\
	107 & Mutagenicity & 131,488 & 266,894 & 14 & NetRepo & Chemical \\
	108 & Tox21-MMP & 127,998 & 260,962 & 47 & NetRepo & Chemical \\
	109 & Tox21-aromatase & 126,483 & 257,092 & 46 & NetRepo & Chemical \\
	110 & GemsecDeezer-RO & 41,773 & 251,652 & 84 & PyG & Friendship \\
	111 & NELL & 65,755 & 251,550 & 186 & PyG & Knowledgebase \\
	112 & rec-amazon & 91,813 & 251,408 & \textsc{n/a} & NetRepo & Recommendation \\
	113 & fb-CMU-Carnegie49 & 6,637 & 249,967 & 3 & NetRepo & Social Networks \\
	114 & socfb-Middlebury45 & 3,075 & 249,220 & \textsc{n/a} & NetRepo & Social Networks \\
	115 & DBLP & 26,128 & 239,566 & 4 & PyG & Knowledgebase \\
	116 & road-luxembourg-osm & 114,599 & 239,332 & \textsc{n/a} & NetRepo & Road \\
	117 & Amazon-Photo & 7,650 & 238,162 & 8 & PyG & Co-Purchase \\
	118 & ca-HepPh & 11,204 & 235,238 & \textsc{n/a} & NetRepo & Collaboration \\
	119 & EllipticBitcoin & 203,769 & 234,355 & 3 & PyG & Economic Networks \\
	120 & Twitch-FR & 6,551 & 231,883 & 2 & PyG & Friendship \\
	121 & KPGM-log14-8-trial1 & 10,978 & 227,990 & \textsc{n/a} & NetRepo & Synthetic-KPGM \\
	122 & socfb-Trinity100 & 2,613 & 223,992 & \textsc{n/a} & NetRepo & Social Networks \\
	123 & MSRC-21 & 43,644 & 223,312 & 22 & NetRepo & Computer Vision \\
	124 & DHFR-MD & 9,380 & 222,452 & 7 & NetRepo & Chemical \\
	125 & ER-1\_5\_20 & 1,050 & 220,058 & \textsc{n/a} & NetRepo & Synthetic-ER \\
	126 & bio-HS-CX & 4,413 & 217,636 & \textsc{n/a} & NetRepo & Biological Networks \\
	127 & WikipediaNetwork-Squirrel & 5,201 & 217,073 & 5 & PyG & Wikipedia \\
	128 & ER-MD & 9,512 & 209,482 & 10 & NetRepo & Chemical \\
	129 & ER-1\_3\_20 & 1,017 & 206,432 & \textsc{n/a} & NetRepo & Synthetic-ER \\
	130 & soc-wiki-elec & 7,118 & 201,564 & \textsc{n/a} & NetRepo & Social Networks \\
	131 & soc-epinions & 26,588 & 200,240 & \textsc{n/a} & NetRepo & Social Networks \\
	132 & DeezerEurope & 28,281 & 185,504 & 2 & PyG & Friendship \\
	133 & ca-CondMat & 21,363 & 182,572 & \textsc{n/a} & NetRepo & Collaboration \\
	134 & LINKX-amherst41 & 2,235 & 181,908 & 2 & PyG & Friendship \\
	135 & socfb-Oberlin44 & 2,920 & 179,824 & \textsc{n/a} & NetRepo & Social Networks \\
	136 & econ-orani678 & 2,529 & 173,747 & \textsc{n/a} & NetRepo & Economic Networks \\
	137 & tech-internet-as & 40,164 & 170,246 & \textsc{n/a} & NetRepo & Technology \\
	138 & bio-DR-CX & 3,289 & 169,880 & \textsc{n/a} & NetRepo & Biological Networks \\
	139 & Coauthor-CS & 18,333 & 163,788 & 15 & PyG & Coauthor \\
	140 & ER-1\_25\_20 & 886 & 157,146 & \textsc{n/a} & NetRepo & Synthetic-ER \\
	141 & HeterophilousGraph-Questions & 48,921 & 153,540 & 2 & PyG & Interaction \\
	142 & bio-DM-CX & 4,040 & 153,434 & \textsc{n/a} & NetRepo & Biological Networks \\
	143 & Entities-MUTAG & 23,644 & 148,454 & 2 & PyG & Knowledgebase \\
	144 & copresence-SFHH & 403 & 147,114 & \textsc{n/a} & NetRepo & Proximity \\
	145 & rec-movielens-tag-movies-10m & 16,528 & 142,148 & \textsc{n/a} & NetRepo & Recommendation \\
	146 & scc\_fb-forum & 488 & 142,022 & \textsc{n/a} & NetRepo & Temporal Reachability \\
	147 & BZR-MD & 6,519 & 137,734 & 8 & NetRepo & Chemical \\
	148 & ia-frwikinews-user-edits & 25,042 & 137,354 & \textsc{n/a} & NetRepo & Interaction \\
	149 & scc\_retweet & 1,206 & 131,980 & \textsc{n/a} & NetRepo & Temporal Reachability \\
	150 & ER-1\_6\_20 & 803 & 128,654 & \textsc{n/a} & NetRepo & Synthetic-ER \\
	151 & ER-1\_16\_10 & 1,126 & 127,004 & \textsc{n/a} & NetRepo & Synthetic-ER \\
	152 & CitationFull-Cora & 19,793 & 126,842 & 70 & PyG & Citation \\
	153 & bio-SC-HT & 2,084 & 126,054 & \textsc{n/a} & NetRepo & Biological Networks \\
	154 & StochasticBlockModel-3.0 & 1,000 & 123,752 & 4 & PyG & Synthetic-SBM \\
	155 & Twitch-ES & 4,648 & 123,412 & 2 & PyG & Friendship \\
	156 & BA-1\_9\_60 & 1,056 & 123,060 & \textsc{n/a} & NetRepo & Synthetic-BA \\
	157 & socfb-Swarthmore42 & 1,659 & 122,100 & \textsc{n/a} & NetRepo & Social Networks \\
	158 & tech-WHOIS & 7,476 & 113,886 & \textsc{n/a} & NetRepo & Technology \\
	159 & rec-movielens-user-movies-10m & 7,601 & 110,779 & \textsc{n/a} & NetRepo & Recommendation \\
	160 & email-EU & 32,430 & 108,794 & \textsc{n/a} & NetRepo & Email \\
	161 & StochasticBlockModel-2.5 & 1,000 & 107,416 & 4 & PyG & Synthetic-SBM \\
	162 & bio-CE-GN & 2,220 & 107,366 & \textsc{n/a} & NetRepo & Biological Networks \\
	163 & tech-as-caida2007 & 26,475 & 106,762 & \textsc{n/a} & NetRepo & Technology \\
	164 & CitationFull-DBLP & 17,716 & 105,734 & 4 & PyG & Citation \\
	165 & CL-10000-1d7-trial3 & 9,267 & 105,485 & \textsc{n/a} & NetRepo & Synthetic-CL \\
	166 & cit-DBLP & 12,591 & 99,255 & \textsc{n/a} & NetRepo & Citation \\
	167 & soc-anybeat & 12,645 & 98,264 & \textsc{n/a} & NetRepo & Social Networks \\
	168 & KPGM-log12-16-trial3 & 3,324 & 97,110 & \textsc{n/a} & NetRepo & Synthetic-KPGM \\
	169 & bio-CE-PG & 1,871 & 95,508 & \textsc{n/a} & NetRepo & Biological Networks \\
	170 & web-indochina-2004 & 11,358 & 95,212 & \textsc{n/a} & NetRepo & Web Graphs \\
	171 & HeterophilousGraph-Amazon-ratings & 24,492 & 93,050 & 5 & PyG & Co-Purchase \\
	172 & BA-1\_6\_60 & 803 & 92,700 & \textsc{n/a} & NetRepo & Synthetic-BA \\
	173 & econ-beaflw & 502 & 90,202 & \textsc{n/a} & NetRepo & Economic Networks \\
	174 & StochasticBlockModel-2.0 & 1,000 & 89,892 & 4 & PyG & Synthetic-SBM \\
	175 & BA-1\_18\_40 & 1,141 & 89,640 & \textsc{n/a} & NetRepo & Synthetic-BA \\
	176 & CitationFull-PubMed & 19,717 & 88,648 & 3 & PyG & Citation \\
	177 & AttributedGraph-Facebook & 4,039 & 88,234 & 193 & PyG & Social Networks \\
	178 & CL-10000-1d8-trial3 & 9,251 & 87,601 & \textsc{n/a} & NetRepo & Synthetic-CL \\
	179 & econ-beacxc & 492 & 84,754 & \textsc{n/a} & NetRepo & Economic Networks \\
	180 & econ-mbeacxc & 487 & 83,776 & \textsc{n/a} & NetRepo & Economic Networks \\
	181 & econ-mbeaflw & 487 & 83,776 & \textsc{n/a} & NetRepo & Economic Networks \\
	182 & soc-advogato & 6,551 & 82,859 & \textsc{n/a} & NetRepo & Social Networks \\
	183 & BA-1\_3\_40 & 1,017 & 79,720 & \textsc{n/a} & NetRepo & Synthetic-BA \\
	184 & econ-beause & 507 & 79,254 & \textsc{n/a} & NetRepo & Economic Networks \\
	185 & Twitch-RU & 4,385 & 78,993 & 2 & PyG & Friendship \\
	186 & bio-HS-LC & 4,227 & 78,968 & \textsc{n/a} & NetRepo & Biological Networks \\
	187 & soc-gplus & 23,628 & 78,388 & \textsc{n/a} & NetRepo & Social Networks \\
	188 & ia-escorts-dynamic & 10,106 & 78,040 & \textsc{n/a} & NetRepo & Interaction \\
	189 & Twitch-EN & 7,126 & 77,774 & 2 & PyG & Friendship \\
	190 & RandomPartitionGraph-hr0.5-ad15 & 5,000 & 75,702 & 10 & PyG & Synthetic-RandPart \\
	191 & KPGM-log12-12-trial2 & 3,214 & 75,682 & \textsc{n/a} & NetRepo & Synthetic-KPGM \\
	192 & RandomPartitionGraph-hr0.7-ad15 & 5,000 & 75,042 & 10 & PyG & Synthetic-RandPart \\
	193 & RandomPartitionGraph-hr0.1-ad15 & 5,000 & 75,026 & 10 & PyG & Synthetic-RandPart \\
	194 & RandomPartitionGraph-hr0.3-ad15 & 5,000 & 74,978 & 10 & PyG & Synthetic-RandPart \\
	195 & web-spam & 4,767 & 74,750 & \textsc{n/a} & NetRepo & Web Graphs \\
	196 & StochasticBlockModel-1.5 & 1,000 & 74,218 & 4 & PyG & Synthetic-SBM \\
	197 & CL-10000-1d9-trial1 & 9,177 & 73,296 & \textsc{n/a} & NetRepo & Synthetic-CL \\
	198 & econ-mbeause & 492 & 72,818 & \textsc{n/a} & NetRepo & Economic Networks \\
	199 & bio-SC-CC & 2,223 & 69,758 & \textsc{n/a} & NetRepo & Biological Networks \\
	200 & ER-3\_25\_5 & 52,336 & 69,246 & \textsc{n/a} & NetRepo & Synthetic-ER \\
	201 & bio-SC-GT & 1,716 & 67,974 & \textsc{n/a} & NetRepo & Biological Networks \\
	202 & DHFR & 32,075 & 67,352 & 9 & NetRepo & Chemical \\
	203 & AIDS & 31,385 & 64,780 & 38 & NetRepo & Biological Networks \\
	204 & Twitch-PT & 1,912 & 64,510 & 2 & PyG & Friendship \\
	205 & web-wiki-chameleon & 2,277 & 62,792 & \textsc{n/a} & NetRepo & Web Graphs \\
	206 & CL-10000-2d0-trial1 & 9,130 & 62,615 & \textsc{n/a} & NetRepo & Synthetic-CL \\
	207 & soc-political-retweet & 18,470 & 61,157 & 2 & NetRepo & Retweet \\
	208 & MixHopSynthetic-Homophily-0.7 & 5,000 & 59,596 & 10 & PyG & Synthetic-Others \\
	209 & MixHopSynthetic-Homophily-0.3 & 5,000 & 59,596 & 10 & PyG & Synthetic-Others \\
	210 & MixHopSynthetic-Homophily-0.6 & 5,000 & 59,596 & 10 & PyG & Synthetic-Others \\
	211 & MixHopSynthetic-Homophily-0.8 & 5,000 & 59,596 & 10 & PyG & Synthetic-Others \\
	212 & MixHopSynthetic-Homophily-0.9 & 5,000 & 59,596 & 10 & PyG & Synthetic-Others \\
	213 & MixHopSynthetic-Homophily-0.1 & 5,000 & 59,596 & 10 & PyG & Synthetic-Others \\
	214 & MixHopSynthetic-Homophily-0.0 & 5,000 & 59,596 & 10 & PyG & Synthetic-Others \\
	215 & MixHopSynthetic-Homophily-0.4 & 5,000 & 59,596 & 10 & PyG & Synthetic-Others \\
	216 & MixHopSynthetic-Homophily-0.5 & 5,000 & 59,596 & 10 & PyG & Synthetic-Others \\
	217 & MixHopSynthetic-Homophily-0.2 & 5,000 & 59,596 & 10 & PyG & Synthetic-Others \\
	218 & CL-10000-2d1-trial2 & 9,078 & 59,026 & \textsc{n/a} & NetRepo & Synthetic-CL \\
	219 & Entities-AIFB & 8,285 & 58,086 & 4 & PyG & Knowledgebase \\
	220 & LastFMAsia & 7,624 & 55,612 & 18 & PyG & Friendship \\
	221 & KPGM-log12-8-trial3 & 2,968 & 53,510 & \textsc{n/a} & NetRepo & Synthetic-KPGM \\
	222 & reality-call & 27,045 & 52,050 & 2 & NetRepo & Phone Call Networks \\
	223 & web-webbase-2001 & 16,062 & 51,186 & \textsc{n/a} & NetRepo & Web Graphs \\
	224 & econ-poli-large & 15,575 & 50,511 & \textsc{n/a} & NetRepo & Economic Networks \\
	225 & RandomPartitionGraph-hr0.7-ad10 & 5,000 & 49,974 & 10 & PyG & Synthetic-RandPart \\
	226 & RandomPartitionGraph-hr0.5-ad10 & 5,000 & 49,688 & 10 & PyG & Synthetic-RandPart \\
	227 & RandomPartitionGraph-hr0.3-ad10 & 5,000 & 49,548 & 10 & PyG & Synthetic-RandPart \\
	228 & RandomPartitionGraph-hr0.1-ad10 & 5,000 & 49,434 & 10 & PyG & Synthetic-RandPart \\
	229 & tech-pgp & 10,680 & 48,632 & \textsc{n/a} & NetRepo & Technology \\
	230 & scc\_retweet-crawl & 17,151 & 48,030 & \textsc{n/a} & NetRepo & Temporal Reachability \\
	231 & ER-2\_7\_5 & 9,583 & 46,642 & \textsc{n/a} & NetRepo & Synthetic-ER \\
	232 & BA-1\_10\_60-L5 & 804 & 46,410 & 5 & NetRepo & Synthetic-BA \\
	233 & CL-10K-1d8-L5 & 10,000 & 44,896 & 5 & NetRepo & Synthetic-CL \\
	234 & MSRC-9 & 8,968 & 43,288 & 10 & NetRepo & Computer Vision \\
	235 & bio-SC-LC & 2,004 & 40,904 & \textsc{n/a} & NetRepo & Biological Networks \\
	236 & MSRC-21C & 8,418 & 40,380 & 21 & NetRepo & Computer Vision \\
	237 & ER-3\_16\_5 & 32,358 & 40,346 & \textsc{n/a} & NetRepo & Synthetic-ER \\
	238 & StochasticBlockModel-0.5 & 1,000 & 40,020 & 4 & PyG & Synthetic-SBM \\
	239 & StochasticBlockModel-1.0 & 1,000 & 40,020 & 4 & PyG & Synthetic-SBM \\
	240 & HeterophilousGraph-Minesweeper & 10,000 & 39,402 & 2 & PyG & Synthetic-Others \\
	241 & web-BerkStan & 12,305 & 39,000 & \textsc{n/a} & NetRepo & Web Graphs \\
	242 & LINKX-reed98 & 962 & 37,624 & 2 & PyG & Friendship \\
	243 & WikipediaNetwork-Chameleon & 2,277 & 36,101 & 5 & PyG & Wikipedia \\
	244 & IMDB & 11,616 & 34,212 & 3 & PyG & Knowledgebase \\
	245 & ER-1\_14\_5 & 829 & 34,184 & \textsc{n/a} & NetRepo & Synthetic-ER \\
	246 & copresence-InVS15 & 219 & 33,450 & \textsc{n/a} & NetRepo & Proximity \\
	247 & socfb-Caltech & 769 & 33,312 & \textsc{n/a} & NetRepo & Social Networks \\
	248 & soc-hamsterster & 2,426 & 33,260 & \textsc{n/a} & NetRepo & Social Networks \\
	249 & HeterophilousGraph-Roman-empire & 22,662 & 32,927 & 18 & PyG & Wikipedia \\
	250 & bn-mouse-brain1 & 213 & 32,331 & \textsc{n/a} & NetRepo & Brain Networks \\
	251 & inf-openflights & 2,939 & 31,354 & \textsc{n/a} & NetRepo & Infrastructure \\
	252 & BZR & 14,479 & 31,070 & 10 & NetRepo & Chemical \\
	253 & Actor & 7,600 & 30,019 & 5 & PyG & Wikipedia \\
	254 & SW-10000-6-0d3-L5 & 10,000 & 30,000 & 5 & NetRepo & Synthetic-Others \\
	255 & SW-10000-6-0d3-L2 & 10,000 & 30,000 & 2 & NetRepo & Synthetic-Others \\
	256 & soc-sign-bitcoinalpha & 3,783 & 28,248 & \textsc{n/a} & NetRepo & Social Networks \\
	257 & bio-HS-HT & 2,570 & 27,382 & \textsc{n/a} & NetRepo & Biological Networks \\
	258 & ca-GrQc & 4,158 & 26,844 & \textsc{n/a} & NetRepo & Collaboration \\
	259 & EmailEUCore & 1,005 & 25,571 & 42 & PyG & Email \\
	260 & bio-grid-fission-yeast & 2,026 & 25,274 & \textsc{n/a} & NetRepo & Biological Networks \\
	261 & RandomPartitionGraph-hr0.5-ad5 & 5,000 & 25,176 & 10 & PyG & Synthetic-RandPart \\
	262 & RandomPartitionGraph-hr0.7-ad5 & 5,000 & 25,056 & 10 & PyG & Synthetic-RandPart \\
	263 & RandomPartitionGraph-hr0.1-ad5 & 5,000 & 24,754 & 10 & PyG & Synthetic-RandPart \\
	264 & RandomPartitionGraph-hr0.3-ad5 & 5,000 & 24,704 & 10 & PyG & Synthetic-RandPart \\
	265 & ca-DBLP-kang & 2,879 & 22,652 & \textsc{n/a} & NetRepo & Collaboration \\
	266 & power-bcspwr10 & 5,300 & 21,842 & \textsc{n/a} & NetRepo & Power Networks \\
	267 & KPGM-log10-16-trial2 & 883 & 21,148 & \textsc{n/a} & NetRepo & Synthetic-KPGM \\
	268 & email-dnc-corecipient & 906 & 20,858 & \textsc{n/a} & NetRepo & Email \\
	269 & BA-1\_8\_10 & 1,040 & 20,690 & \textsc{n/a} & NetRepo & Synthetic-BA \\
	270 & rt\_lolgop & 9,765 & 20,150 & \textsc{n/a} & NetRepo & Retweet \\
	271 & scc\_enron-only & 146 & 19,656 & \textsc{n/a} & NetRepo & Temporal Reachability \\
	272 & rt\_barackobama & 9,631 & 19,547 & \textsc{n/a} & NetRepo & Retweet \\
	273 & rt\_justinbieber & 9,405 & 19,167 & \textsc{n/a} & NetRepo & Retweet \\
	274 & SFHH-conf-sensor & 403 & 19,130 & \textsc{n/a} & NetRepo & Proximity \\
	275 & PolBlogs & 1,490 & 19,025 & 2 & PyG & Web Graphs \\
	276 & power-eris1176 & 1,176 & 18,552 & \textsc{n/a} & NetRepo & Power Networks \\
	277 & AttributedGraph-Wiki & 2,405 & 17,981 & 17 & PyG & Wikipedia \\
	278 & bn-fly-drosophila-medulla1 & 1,781 & 17,927 & \textsc{n/a} & NetRepo & Brain Networks \\
	279 & web-EPA & 4,271 & 17,818 & \textsc{n/a} & NetRepo & Web Graphs \\
	280 & BA-1\_17\_10 & 895 & 17,790 & \textsc{n/a} & NetRepo & Synthetic-BA \\
	281 & rt\_gmanews & 8,373 & 17,438 & \textsc{n/a} & NetRepo & Retweet \\
	282 & BA-1\_1\_10 & 862 & 17,130 & \textsc{n/a} & NetRepo & Synthetic-BA \\
	283 & rt\_mittromney & 7,974 & 17,074 & \textsc{n/a} & NetRepo & Retweet \\
	284 & KPGM-log10-12-trial1 & 845 & 16,934 & \textsc{n/a} & NetRepo & Synthetic-KPGM \\
	285 & primary-school-proximity & 242 & 16,634 & \textsc{n/a} & NetRepo & Proximity \\
	286 & BA-1\_12\_10 & 827 & 16,430 & \textsc{n/a} & NetRepo & Synthetic-BA \\
	287 & CitationFull-CoraML & 2,995 & 16,316 & 7 & PyG & Citation \\
	288 & rt\_ksa & 7,302 & 16,216 & \textsc{n/a} & NetRepo & Retweet \\
	289 & rt\_onedirection & 7,987 & 16,203 & \textsc{n/a} & NetRepo & Retweet \\
	290 & rt\_saudi & 7,252 & 16,121 & \textsc{n/a} & NetRepo & Retweet \\
	291 & ia-reality & 6,809 & 15,360 & \textsc{n/a} & NetRepo & Interaction \\
	292 & econ-mahindas & 1,258 & 15,132 & \textsc{n/a} & NetRepo & Economic Networks \\
	293 & ca-Erdos992 & 5,094 & 15,030 & \textsc{n/a} & NetRepo & Collaboration \\
	294 & rt\_dash & 6,288 & 14,870 & \textsc{n/a} & NetRepo & Retweet \\
	295 & DD21 & 5,748 & 14,267 & 21 & NetRepo & Misc \\
	296 & rt\_alwefaq & 4,171 & 14,122 & \textsc{n/a} & NetRepo & Retweet \\
	297 & Airports-USA & 1,190 & 13,599 & 4 & PyG & Flight \\
	298 & tech-routers-rf & 2,113 & 13,264 & \textsc{n/a} & NetRepo & Technology \\
	299 & power-US-Grid & 4,941 & 13,188 & \textsc{n/a} & NetRepo & Power Networks \\
	300 & Peking-1 & 3,341 & 13,150 & 190 & NetRepo & Social Networks \\
	301 & web-edu & 3,031 & 12,948 & \textsc{n/a} & NetRepo & Web Graphs \\
	302 & rt\_uae & 5,248 & 12,772 & \textsc{n/a} & NetRepo & Retweet \\
	303 & rt\_oman & 4,904 & 12,456 & \textsc{n/a} & NetRepo & Retweet \\
	304 & scc\_infect-hyper & 113 & 12,444 & \textsc{n/a} & NetRepo & Temporal Reachability \\
	305 & econ-poli & 4,008 & 12,246 & \textsc{n/a} & NetRepo & Economic Networks \\
	306 & KPGM-log10-8-trial2 & 796 & 12,080 & \textsc{n/a} & NetRepo & Synthetic-KPGM \\
	307 & rt\_p2 & 4,902 & 12,034 & \textsc{n/a} & NetRepo & Retweet \\
	308 & contacts-prox-high-school-2013 & 327 & 11,636 & \textsc{n/a} & NetRepo & Proximity \\
	309 & rt\_gop & 4,687 & 11,058 & \textsc{n/a} & NetRepo & Retweet \\
	310 & rt\_tcot & 4,547 & 11,004 & \textsc{n/a} & NetRepo & Retweet \\
	311 & email-univ & 1,133 & 10,902 & \textsc{n/a} & NetRepo & Email \\
	312 & CitationFull-CiteSeer & 4,230 & 10,674 & 6 & PyG & Citation \\
	313 & ca-cora & 2,708 & 10,556 & \textsc{n/a} & NetRepo & Collaboration \\
	314 & PTC-FR & 5,110 & 10,532 & 19 & NetRepo & Chemical \\
	315 & DD6 & 4,152 & 10,320 & 20 & NetRepo & Misc \\
	316 & PTC-FM & 4,925 & 10,110 & 18 & NetRepo & Chemical \\
	317 & PTC-MR & 4,915 & 10,108 & 18 & NetRepo & Chemical \\
	318 & CL-1000-1d7-trial2 & 932 & 9,755 & \textsc{n/a} & NetRepo & Synthetic-CL \\
	319 & PTC-MM & 4,695 & 9,624 & 20 & NetRepo & Chemical \\
	320 & bio-DM-HT & 2,989 & 9,320 & \textsc{n/a} & NetRepo & Biological Networks \\
	321 & CL-1000-1d7-trial1 & 928 & 9,279 & \textsc{n/a} & NetRepo & Synthetic-CL \\
	322 & rt\_islam & 4,497 & 9,232 & \textsc{n/a} & NetRepo & Retweet \\
	323 & scc\_rt\_lolgop & 273 & 9,020 & \textsc{n/a} & NetRepo & Temporal Reachability \\
	324 & rt\_tlot & 3,665 & 8,949 & \textsc{n/a} & NetRepo & Retweet \\
	325 & rt\_lebanon & 3,961 & 8,871 & \textsc{n/a} & NetRepo & Retweet \\
	326 & email-dnc-leak & 1,891 & 8,849 & \textsc{n/a} & NetRepo & Email \\
	327 & TerroristRel & 881 & 8,592 & 2 & NetRepo & Collaboration \\
	328 & bio-SC-TS & 636 & 7,918 & \textsc{n/a} & NetRepo & Biological Networks \\
	329 & biogrid-human & 2,005 & 7,918 & \textsc{n/a} & NetRepo & Biological Networks \\
	330 & rt\_occupy & 3,225 & 7,883 & \textsc{n/a} & NetRepo & Retweet \\
	331 & copresence-InVS13 & 95 & 7,830 & \textsc{n/a} & NetRepo & Proximity \\
	332 & rt\_damascus & 3,052 & 7,738 & \textsc{n/a} & NetRepo & Retweet \\
	333 & rt\_occupywallstnyc & 3,609 & 7,663 & \textsc{n/a} & NetRepo & Retweet \\
	334 & biogrid-worm & 1,930 & 7,152 & \textsc{n/a} & NetRepo & Biological Networks \\
	335 & road-minnesota & 2,642 & 6,606 & \textsc{n/a} & NetRepo & Road \\
	336 & power-bcspwr09 & 1,723 & 6,511 & \textsc{n/a} & NetRepo & Power Networks \\
	337 & email-radoslaw & 167 & 6,501 & \textsc{n/a} & NetRepo & Email \\
	338 & bio-CE-GT & 924 & 6,478 & \textsc{n/a} & NetRepo & Biological Networks \\
	339 & bn-macaque-rhesus-brain1 & 242 & 6,108 & \textsc{n/a} & NetRepo & Brain Networks \\
	340 & CL-1000-2d0-trial3 & 916 & 6,004 & \textsc{n/a} & NetRepo & Synthetic-CL \\
	341 & Airports-Europe & 399 & 5,995 & 4 & PyG & Flight \\
	342 & bio-CE-HT & 2,617 & 5,970 & \textsc{n/a} & NetRepo & Biological Networks \\
	343 & CL-1000-2d0-trial2 & 899 & 5,861 & \textsc{n/a} & NetRepo & Synthetic-CL \\
	344 & soc-wiki-Vote & 889 & 5,828 & \textsc{n/a} & NetRepo & Social Networks \\
	345 & rt\_assad & 2,139 & 5,574 & \textsc{n/a} & NetRepo & Retweet \\
	346 & web-google & 1,299 & 5,546 & \textsc{n/a} & NetRepo & Web Graphs \\
	347 & infect-dublin & 410 & 5,530 & \textsc{n/a} & NetRepo & Proximity \\
	348 & CL-1000-2d1-trial2 & 911 & 5,457 & \textsc{n/a} & NetRepo & Synthetic-CL \\
	349 & AttributedGraph-Cora & 2,708 & 5,429 & 7 & PyG & Citation \\
	350 & econ-wm1 & 260 & 4,943 & \textsc{n/a} & NetRepo & Economic Networks \\
	351 & rt\_voteonedirection & 2,280 & 4,928 & \textsc{n/a} & NetRepo & Retweet \\
	352 & econ-wm3 & 259 & 4,918 & \textsc{n/a} & NetRepo & Economic Networks \\
	353 & econ-wm2 & 259 & 4,908 & \textsc{n/a} & NetRepo & Economic Networks \\
	354 & AttributedGraph-CiteSeer & 3,312 & 4,715 & 6 & PyG & Citation \\
	355 & web-polblogs & 643 & 4,560 & \textsc{n/a} & NetRepo & Web Graphs \\
	356 & bn-macaque-rhesus-interareal-cortical2 & 93 & 4,524 & \textsc{n/a} & NetRepo & Brain Networks \\
	357 & infect-hyper & 113 & 4,392 & \textsc{n/a} & NetRepo & Proximity \\
	358 & eco-foodweb-baydry & 128 & 4,212 & \textsc{n/a} & NetRepo & Ecology \\
	359 & eco-foodweb-baywet & 128 & 4,150 & \textsc{n/a} & NetRepo & Ecology \\
	360 & eco-florida & 128 & 4,150 & \textsc{n/a} & NetRepo & Ecology \\
	361 & power-1138-bus & 1,138 & 4,054 & \textsc{n/a} & NetRepo & Power Networks \\
	362 & KPGM-log8-12-trial3 & 230 & 3,522 & \textsc{n/a} & NetRepo & Synthetic-KPGM \\
	363 & ca-CSphd & 1,882 & 3,480 & \textsc{n/a} & NetRepo & Collaboration \\
	364 & DD242 & 1,284 & 3,303 & 20 & NetRepo & Misc \\
	365 & bio-CE-LC & 1,387 & 3,296 & \textsc{n/a} & NetRepo & Biological Networks \\
	366 & biogrid-mouse & 1,450 & 3,272 & \textsc{n/a} & NetRepo & Biological Networks \\
	367 & power-685-bus & 685 & 3,249 & \textsc{n/a} & NetRepo & Power Networks \\
	368 & bn-mouse-kasthuri-v4 & 1,029 & 3,118 & \textsc{n/a} & NetRepo & Brain Networks \\
	369 & KPGM-log8-10-trial3 & 224 & 3,040 & \textsc{n/a} & NetRepo & Synthetic-KPGM \\
	370 & ia-crime-moreno & 829 & 2,948 & \textsc{n/a} & NetRepo & Interaction \\
	371 & eco-mangwet & 97 & 2,892 & \textsc{n/a} & NetRepo & Ecology \\
	372 & inf-euroroad & 1,174 & 2,834 & \textsc{n/a} & NetRepo & Infrastructure \\
	373 & road-euroroad & 1,174 & 2,834 & \textsc{n/a} & NetRepo & Road \\
	374 & bn-macaque-rhesus-cerebral-cortex1 & 91 & 2,802 & \textsc{n/a} & NetRepo & Brain Networks \\
	375 & copresence-LH10 & 73 & 2,762 & \textsc{n/a} & NetRepo & Proximity \\
	376 & DD\_g106 & 574 & 2,710 & \textsc{n/a} & NetRepo & Protein \\
	377 & KPGM-log8-8-trial3 & 215 & 2,606 & \textsc{n/a} & NetRepo & Synthetic-KPGM \\
	378 & road-ChicagoRegional & 1,467 & 2,596 & \textsc{n/a} & NetRepo & Road \\
	379 & power-662-bus & 662 & 2,474 & \textsc{n/a} & NetRepo & Power Networks \\
	380 & DD\_g105 & 423 & 2,384 & \textsc{n/a} & NetRepo & Protein \\
	381 & hospital-ward-proximity & 75 & 2,278 & \textsc{n/a} & NetRepo & Proximity \\
	382 & DD\_g108 & 483 & 2,274 & \textsc{n/a} & NetRepo & Protein \\
	383 & bio-DM-LC & 658 & 2,258 & \textsc{n/a} & NetRepo & Biological Networks \\
	384 & scc\_rt\_gmanews & 135 & 2,156 & \textsc{n/a} & NetRepo & Temporal Reachability \\
	385 & biogrid-yeast & 836 & 2,098 & \textsc{n/a} & NetRepo & Biological Networks \\
	386 & rt-twitter-copen & 761 & 2,058 & \textsc{n/a} & NetRepo & Retweet \\
	387 & DD\_g100 & 349 & 2,010 & \textsc{n/a} & NetRepo & Protein \\
	388 & DD\_g104 & 372 & 1,998 & \textsc{n/a} & NetRepo & Protein \\
	389 & DD\_g115 & 336 & 1,892 & \textsc{n/a} & NetRepo & Protein \\
	390 & scc\_rt\_occupywallstnyc & 127 & 1,862 & \textsc{n/a} & NetRepo & Temporal Reachability \\
	391 & soc-physicians & 241 & 1,846 & \textsc{n/a} & NetRepo & Social Networks \\
	392 & ca-netscience & 379 & 1,828 & \textsc{n/a} & NetRepo & Collaboration \\
	393 & eco-everglades & 69 & 1,765 & \textsc{n/a} & NetRepo & Ecology \\
	394 & biogrid-plant & 523 & 1,676 & \textsc{n/a} & NetRepo & Biological Networks \\
	395 & power-494-bus & 494 & 1,666 & \textsc{n/a} & NetRepo & Power Networks \\
	396 & DD\_g1021 & 329 & 1,574 & \textsc{n/a} & NetRepo & Protein \\
	397 & DD\_g11 & 312 & 1,522 & \textsc{n/a} & NetRepo & Protein \\
	398 & ia-workplace-contacts & 92 & 1,510 & \textsc{n/a} & NetRepo & Interaction \\
	399 & bn-cat-mixed-species-brain1 & 65 & 1,460 & \textsc{n/a} & NetRepo & Brain Networks \\
	400 & DD\_g1022 & 294 & 1,460 & \textsc{n/a} & NetRepo & Protein \\
	401 & DD\_g101 & 306 & 1,456 & \textsc{n/a} & NetRepo & Protein \\
	402 & DD\_g103 & 265 & 1,294 & \textsc{n/a} & NetRepo & Protein \\
	403 & email-enron-only & 143 & 1,246 & \textsc{n/a} & NetRepo & Email \\
	404 & bn-macaque-rhesus-brain2 & 91 & 1,164 & \textsc{n/a} & NetRepo & Brain Networks \\
	405 & DD\_g1006 & 246 & 1,136 & \textsc{n/a} & NetRepo & Protein \\
	406 & Airports-Brazil & 131 & 1,074 & 4 & PyG & Flight \\
	407 & scc\_rt\_justinbieber & 62 & 884 & \textsc{n/a} & NetRepo & Temporal Reachability \\
	408 & DD\_g1000 & 183 & 816 & \textsc{n/a} & NetRepo & Protein \\
	409 & DD\_g1017 & 162 & 752 & \textsc{n/a} & NetRepo & Protein \\
	410 & scc\_rt\_alwefaq & 72 & 710 & \textsc{n/a} & NetRepo & Temporal Reachability \\
	411 & DD\_g1019 & 131 & 706 & \textsc{n/a} & NetRepo & Protein \\
	412 & eco-stmarks & 54 & 703 & \textsc{n/a} & NetRepo & Ecology \\
	413 & DD\_g1030 & 136 & 702 & \textsc{n/a} & NetRepo & Protein \\
	414 & DD\_g10 & 146 & 656 & \textsc{n/a} & NetRepo & Protein \\
	415 & internet-industry-partnerships & 219 & 631 & 3 & NetRepo & Collaboration \\
	416 & soc-student-coop & 185 & 622 & \textsc{n/a} & NetRepo & Social Networks \\
	417 & DD\_g1016 & 113 & 582 & \textsc{n/a} & NetRepo & Protein \\
	418 & soc-highschool-moreno & 70 & 548 & \textsc{n/a} & NetRepo & Social Networks \\
	419 & DD\_g1009 & 129 & 544 & \textsc{n/a} & NetRepo & Protein \\
	420 & webkb-wisc & 265 & 530 & 5 & NetRepo & Web Graphs \\
	421 & WebKB-Wisconsin & 251 & 515 & 5 & PyG & Web Graphs \\
	422 & DD\_g1015 & 102 & 488 & \textsc{n/a} & NetRepo & Protein \\
	423 & DD\_g1004 & 94 & 460 & \textsc{n/a} & NetRepo & Protein \\
	424 & scc\_rt\_barackobama & 80 & 452 & \textsc{n/a} & NetRepo & Temporal Reachability \\
	425 & DD\_g1027 & 108 & 446 & \textsc{n/a} & NetRepo & Protein \\
	426 & bn-mouse-visual-cortex2 & 193 & 428 & \textsc{n/a} & NetRepo & Brain Networks \\
	427 & DD\_g1025 & 88 & 410 & \textsc{n/a} & NetRepo & Protein \\
	428 & WebKB-Texas & 183 & 325 & 5 & PyG & Web Graphs \\
	429 & WebKB-Cornell & 183 & 298 & 5 & PyG & Web Graphs \\
	430 & ENZYMES\_g297 & 121 & 298 & \textsc{n/a} & NetRepo & Chemical \\
	431 & ENZYMES\_g296 & 125 & 282 & \textsc{n/a} & NetRepo & Chemical \\
	432 & DD\_g1028 & 72 & 274 & \textsc{n/a} & NetRepo & Protein \\
	433 & ENZYMES\_g118 & 95 & 242 & \textsc{n/a} & NetRepo & Chemical \\
	434 & ENZYMES\_g504 & 66 & 240 & \textsc{n/a} & NetRepo & Chemical \\
	435 & DD\_g1003 & 53 & 232 & \textsc{n/a} & NetRepo & Protein \\
	436 & ENZYMES\_g103 & 59 & 230 & \textsc{n/a} & NetRepo & Chemical \\
	437 & ENZYMES\_g594 & 52 & 228 & \textsc{n/a} & NetRepo & Chemical \\
	438 & ENZYMES\_g355 & 66 & 224 & \textsc{n/a} & NetRepo & Chemical \\
	439 & NCI1\_g3139 & 107 & 224 & \textsc{n/a} & NetRepo & Chemical \\
	440 & NCI1\_g1863 & 107 & 222 & \textsc{n/a} & NetRepo & Chemical \\
	441 & ENZYMES\_g575 & 51 & 220 & \textsc{n/a} & NetRepo & Chemical \\
	442 & ENZYMES\_g526 & 58 & 220 & \textsc{n/a} & NetRepo & Chemical \\
	443 & ENZYMES\_g199 & 62 & 216 & \textsc{n/a} & NetRepo & Chemical \\
	444 & ENZYMES\_g279 & 60 & 214 & \textsc{n/a} & NetRepo & Chemical \\
	445 & NCI1\_g3585 & 105 & 214 & \textsc{n/a} & NetRepo & Chemical \\
	446 & ENZYMES\_g527 & 57 & 214 & \textsc{n/a} & NetRepo & Chemical \\
	447 & NCI1\_g1677 & 102 & 212 & \textsc{n/a} & NetRepo & Chemical \\
	448 & NCI1\_g3711 & 89 & 212 & \textsc{n/a} & NetRepo & Chemical \\
	449 & ENZYMES\_g224 & 54 & 210 & \textsc{n/a} & NetRepo & Chemical \\
	450 & NCI1\_g3990 & 90 & 210 & \textsc{n/a} & NetRepo & Chemical \\
	451 & ENZYMES\_g291 & 62 & 208 & \textsc{n/a} & NetRepo & Chemical \\
	452 & NCI1\_g2079 & 88 & 206 & \textsc{n/a} & NetRepo & Chemical \\
	453 & NCI1\_g3444 & 93 & 204 & \textsc{n/a} & NetRepo & Chemical \\
	454 & NCI1\_g1893 & 96 & 204 & \textsc{n/a} & NetRepo & Chemical \\
	455 & NCI1\_g2082 & 86 & 202 & \textsc{n/a} & NetRepo & Chemical \\
	456 & ENZYMES\_g598 & 55 & 200 & \textsc{n/a} & NetRepo & Chemical \\
	457 & KarateClub & 34 & 156 & 4 & PyG & Friendship \\
	
\end{longtable}

\end{normalsize}

\section{Hosting, Licensing, and Maintenance Plan}\label{app:hlm}

\subsection{Hosting}\label{app:hlm:hosting}
The code and data of \benchmark are hosted at \benchmarkurl.

\subsection{Licensing}\label{app:hlm:licensing}

\textbf{Data License.} Among data included in \benchmark, 
performance records~\ref{app:data:overview:perf},
splitting of graph data~\ref{app:data:overview:graphsplits} and testbed data~\ref{app:data:overview:testbedsplits}, and
meta-graph features~\ref{app:data:overview:metafeats} were generated by \benchmark
by processing the graph data~(\Cref{app:data:graph}).
They are under the \href{https://creativecommons.org/licenses/by-nc/4.0/legalcode}{CC BY-NC 4.0} license.
Graph data~(\Cref{app:data:graph}) are from two public graph repositories, \ie, 
Network Repository~\cite{DBLP:conf/aaai/RossiA15} and PyTorch Geometric~\cite{DBLP:journals/corr/abs-1903-02428}.
Since all graph data used in this work are from these repositories, 
please refer to the corresponding repository for the license of graph datasets.

\textbf{Code License.} 
The \benchmark codebase is under the \href{https://creativecommons.org/licenses/by-nc/4.0/legalcode}{CC BY-NC 4.0} license.

\subsection{Maintenance}\label{app:hlm:maintenance}

We will maintain and continue to develop \benchmark for the long term.
Specifically, we aim to improve and expand \benchmark in two aspects, \ie, benchmark data and model selection algorithms.

\textbf{Benchmark Data.}
We will monitor newly available graphs from various domains, as well as new graph learning (GL) models, 
and expand \benchmark with those new graphs and GL models as follows.
\begin{itemize}[leftmargin=0.9em,topsep=0pt,itemsep=0pt]
	\item \textit{Performance Records:} We will evaluate new GL models on both new and existing graphs for applicable GL tasks, 
	and evaluate existing GL models on the new graphs as well.
	Those new results will enrich our collection of performance records.
	
	\item \textit{Graph Data Splits:} Node and edge splits to evaluate GL models on the new graphs will be added.
	
	\item \textit{Testbed Data Splits:} Testbed data splits (\eg, over the performance matrix) used to evaluate model selection algorithms
	will be shared.
	
	\item \textit{Meta-Graph Features:} We will generate different sets of meta-graph features for the new graphs.
\end{itemize}

\textbf{Model Selection Algorithms.}
As we discuss in the main text, there exist multiple directions for future work 
to improve the effectiveness and generalization capability of GL model selection algorithms.
We will continue to monitor the latest advancement of this area, and 
expand \benchmark with the state-of-the-art GL model selection algorithms.

\section{Author Statement}
We take all responsibilities of the benchmark data. 
In case of violation of any rights or data~licenses, we will take necessary actions, 
such as revising the problematic data, or removing it from the~benchmark.

\end{document}